\pdfoutput=1

\documentclass{article}[11pt]
\usepackage{fullpage}

\usepackage[utf8]{inputenc}
\usepackage{hyperref}
\usepackage{amsfonts}
\usepackage{amsmath}
\usepackage{amssymb}
\usepackage{amsthm}
\usepackage{qtree}
\usepackage{graphicx}
\usepackage{booktabs}
\usepackage{subcaption}
\usepackage[section]{placeins}
\graphicspath{ {./figures/} }
\usepackage{xcolor}
\usepackage[english]{babel}
\usepackage[ruled,vlined]{algorithm2e}
\usepackage{bbm}

\usepackage{url}
\usepackage{float}
\usepackage{enumitem}
\usepackage{bbm}
\usepackage{mathtools}
\usepackage{multirow}

\definecolor{DarkGreen}{rgb}{0.1,0.5,0.1}
\definecolor{DarkRed}{rgb}{0.5,0.1,0.1}
\definecolor{DarkBlue}{rgb}{0.1,0.1,0.5}
\hypersetup{
    unicode=false,          
    pdftoolbar=true,        
    pdfmenubar=true,        
    pdffitwindow=false,     
    pdfnewwindow=true,      
    colorlinks=true,       
    linkcolor=DarkGreen,    
    citecolor=DarkGreen,    
    filecolor=DarkGreen,    
    urlcolor=DarkBlue,
    linktocpage=true,
}

\usepackage{natbib}
\setcitestyle{authoryear,round,citesep={;},aysep={,},yysep={;}}

\definecolor{RoyalBlue}{RGB}{0,100,170}
\definecolor{peach}{rgb}{1, 0.56, 0.56}
\definecolor{midgray}{RGB}{150,150,150}
\definecolor{EasternBlue}{RGB}{37,150,190}
\definecolor{sand}{RGB}{250,150,120}
\definecolor{grass}{RGB}{120, 190, 50}
\definecolor{sky}{RGB}{50,150,250}
\definecolor{Orange}{RGB}{250,150,50}
\definecolor{Cerulean}{RGB}{80,150,220}
\definecolor{Emerald}{RGB}{62,156,94}
\definecolor{Rouge}{RGB}{250,95,95}

\definecolor{ColorDef}{RGB}{80, 180, 150}
\definecolor{RevisionRed}{RGB}{240,35,35}
\definecolor{Confirm}{RGB}{103, 162, 229}
\definecolor{TODO}{RGB}{250,150,50}
\definecolor{Future}{RGB}{198, 121, 235}

\definecolor{myGold}{RGB}{231,141,20}
\definecolor{myBlue}{rgb}{0.19,0.41,.65}
\definecolor{myPurple}{RGB}{175,0,124}
\definecolor{Gred}{RGB}{219, 50, 54}
\definecolor{Ggreen}{RGB}{60, 186, 84}
\definecolor{Gblue}{RGB}{72, 133, 237}
\definecolor{Gyellow}{RGB}{247, 178, 16}
\usepackage[colorinlistoftodos,prependcaption,textsize=footnotesize]{todonotes} 

\providecommand{\tdotoggle}{1}
\ifnum\tdotoggle=1
\paperwidth=\dimexpr \paperwidth + 3.2cm\relax
\oddsidemargin=\dimexpr\oddsidemargin + 1.4cm\relax
\evensidemargin=\dimexpr\evensidemargin + 1.4cm\relax
\marginparwidth=\dimexpr \marginparwidth + 3cm\relax
\fi
\newcommand{\mytodo}[1]{\ifnum\tdotoggle=1{#1}\fi}

\newcommand{\tableoftodos}{\ifnum\tdotoggle=1 \listoftodos[Comments/To Do's] \fi}

\usepackage{amsmath,amsfonts,bm}
\usepackage{cleveref}
\usepackage{thmtools}
\usepackage{dsfont}
\usepackage{stackrel}
\usepackage{tikz}

\newtheorem*{theorem*}{Theorem}

\newtheorem{lemma}{Lemma}

\newtheorem{proposition}{Proposition}
\newtheorem{corollary}{Corollary}

\newtheorem{assumption}{Assumption}
\newtheorem{definition}{Definition}
\newtheorem*{remark*}{Remark}
\newtheorem{remark}{Remark}

\makeatletter

\newcommand{\newreptheorem}[2]
{\newenvironment{rep#1}[1]
	{\def\rep@title{#2 \ref{##1}} \begin{rep@theorem}}%
		{\end{rep@theorem}}}
\makeatother
\newreptheorem{theorem}{Theorem}
\newreptheorem{lemma}{Lemma}
\newreptheorem{corollary}{Corollary}
\newreptheorem{proposition}{Proposition}

\crefname{hypothesis}{Hypothesis}{Hypothesis}
\crefname{condition}{Condition}{Conditions}
\crefname{distribution}{Distribution}{Distributions}

\crefname{thm}{Theorem}{Theorems}

\newcommand{\compl}[1]{-#1}

\newcommand{\tmle}{\hat \theta_{\mbox{MLE}}}

\newcommand{\hess}{\nabla^2}
\newcommand{\fisher}{\mathcal{I}}

\def\1{\bm{1}}

\def\vone{{\bm{1}}}

\def\vb{{\bm{b}}}

\def\ve{{\bm{e}}}

\def\vh{{\bm{h}}}

\def\vw{{\bm{w}}}
\def\vx{{\bm{x}}}
\def\vy{{\bm{y}}}
\def\vz{{\bm{z}}}

\def\mE{{\bm{E}}}

\def\mJ{{\bm{J}}}

\def\mP{{\bm{P}}}

\def\mW{{\bm{W}}}
\def\mX{{\bm{X}}}
\def\mY{{\bm{Y}}}

\DeclareMathAlphabet{\mathsfit}{\encodingdefault}{\sfdefault}{m}{sl}
\SetMathAlphabet{\mathsfit}{bold}{\encodingdefault}{\sfdefault}{bx}{n}

\newcommand{\N}{\mathbb{N}}
\newcommand{\R}{\mathbb{R}}

\newcommand{\KL}{D_{\mathrm{KL}}}
\newcommand{\KLp}[2]{
    D_{\mathrm{KL}} \left( #1, #2 \right)
}
\newcommand{\Var}{\mathrm{Var}}
\newcommand{\Varp}[1]{
    \mathrm{Var} \left( #1 \right)
}

\newcommand{\Cov}{\mathrm{Cov}}

\DeclareMathOperator*{\argmax}{arg\,max}
\DeclareMathOperator*{\argmin}{arg\,min}

\DeclareMathOperator{\Tr}{Tr}

\newcommand{\indp}[1]{{\mathds{1}_{#1}}}  

\newcommand{\norm}[1]{\left\|#1\right\|}
\def\abs#1{\left| #1 \right|}
\newcommand{\floor}[1]{\left\lfloor\, {#1}\,\right\rfloor}
\newcommand{\ceil}[1]{\left\lceil\, {#1}\,\right\rceil}
\newcommand{\inparen}[1]{\left(#1\right)}              

\newcommand*{\E}{\mathbb{E}}  
\newcommand*{\Ep}[2]{
    \mathbb{E}_{#1} \left[ #2 \right]
}

\newcommand*{\probp}[1]{\mathbb{P}\left\{#1\right\}}

\usepackage{multirow}

\newcommand*\circled[1]{\tikz[baseline=(char.base)]{
    \node[shape=circle,draw,inner sep=1pt] (char) {#1};}}
    
\usepackage[hang, bottom,flushmargin]{footmisc}

\newcommand{\margin}{\beta}

\newcommand{\TVp}[2]{
    D_{\mathrm{TV}} \left( #1, #2 \right)
}

\title{
Promises and Pitfalls of Generative Masked Language Modeling: Theoretical Framework and Practical Guidelines
}

\author{
    Yuchen Li$^{1,2}$
    \quad Alexandre Kirchmeyer$^1$\footnotemark[1]
    \quad Aashay Mehta$^1$\footnotemark[1]
    \quad Yilong Qin$^1$\footnotemark[1]  \\
    \quad Boris Dadachev$^2$\footnotemark[1] 
    \quad Kishore Papineni$^2$\footnote{
    Equal theoretical / empirical contribution.
    Correspond to: Yuchen Li \texttt{yuchenl4@cs.cmu.edu} and \\  
    Boris Dadachev \texttt{dadachev@google.com} and
    Andrej Risteski \texttt{aristesk@andrew.cmu.edu}
    }
    \quad Sanjiv Kumar$^2$ 
    \quad Andrej Risteski$^1$ \\
 \normalsize{
    $^1$Carnegie Mellon University
    \qquad $^2$Google Research}
}

\date{}

\begin{document}

\maketitle

\begin{abstract}
Autoregressive language models are the currently dominant paradigm for text generation, 
but they have some fundamental limitations that cannot be remedied by scale---for example inherently sequential and unidirectional generation.
While alternate classes of models have been explored, we have limited mathematical understanding of their fundamental power and limitations. 
In this paper we focus on \emph{Generative Masked Language Models (GMLMs)}, a non-autoregressive paradigm in which we train a model to fit conditional probabilities of the data distribution via masking, which are subsequently used as inputs to a Markov Chain to draw samples from the model. These models empirically strike a promising speed-quality trade-off as each step can be typically parallelized by decoding the entire sequence in parallel.  
We develop a mathematical framework for analyzing and improving such models which sheds light on questions of sample complexity and inference speed and quality.
Empirically, we adapt the T5 model for iteratively-refined parallel decoding, achieving 2-3x speedup in machine translation with minimal sacrifice in quality compared with autoregressive models.
We run careful ablation experiments to give recommendations on key design choices,
and make fine-grained observations on the common error modes in connection with our theory.
Our mathematical analyses and empirical observations characterize both potentials and limitations of this approach, 
and can be applied to future works on improving understanding and performance of GMLMs.~\footnote{
Our codes are released at 
\url{https://github.com/google-research/google-research/tree/master/padir}
} 
\end{abstract}
\vspace{-3mm}

\section{Introduction}
\label{sec:intro}
The current dominant approach to language modeling is \emph{autoregressive} (AR):
to generate a sequence of tokens, the language model starts by predicting the leftmost token, and then proceeds from left to right, each step predicting the next token based on everything on its left 
\citep{raffel2020exploring, brown2020language, touvron2023llama}.
AR models are not without limitations:
\begin{enumerate}
    \item \emph{Lack of parallelism:} To generate a sequence of $N$ tokens, AR language models need $N$ sequential decoding steps. Each step consists of a forward pass of the decoder component. When $N$ is large, $N$ sequential decoding steps incur high latency.
    \item \emph{Quality:} When predicting each token, the model cannot access its right hand side context, 
    and has no natural way to revise earlier predictions on the left. 
    This intuitive limitation was more formally explored in prior theoretical works \citep{li2021limitations, lin2021limitations, bachmann2024pitfalls}.
\end{enumerate}

One promising alternative is based on \emph{Generative Masked Language Models (GMLMs)}. 
They are trained to fit conditional probabilities for parts of the sequence (by applying a mask), conditioned on the rest. 
To produce samples, these conditionals are used as oracles for running Markov Chain, e.g. a Gibbs sampler \citep{wang2019bert, goyal2022exposing}. 
Alternatively, we can think of these steps as an iterative refinement process, typically starting with pure noise (i.e. all tokens are masked or randomized).
One can even fit conditional probabilities for noised versions of the input distribution, and use them as inputs to a denoiser to get certain types of discrete diffusion models 
\citep{austin2021structured}.       
In GMLMs, typically one step of the Markov Chain is operationalized by a Transformer that generates the sequence in parallel (i.e. \emph{parallel decoding} 
\citep{ghazvininejad2019mask, gu2021fully, savinov2022stepunrolled}
). Hence, if the total number of steps is small, the latency is low. 

However, none of these approaches robustly surpass autoregressive models in both speed and quality for a wider range of language generation tasks beyond machine translation. Thus, the following questions naturally arise: 
\begin{enumerate}
    \item[(Q1)] GMLMs are trained to learn conditional probabilities. 
    When does it also imply learning the \emph{joint} probability? 
    \item[(Q2)] What properties of the data distribution and training/inference algorithm govern the quality of the learned model and its generated samples? 
    \item[(Q3)] What are the best practices for training GMLMs, and can we use theory to elucidate the design space of losses, training and inference procedures? 
\end{enumerate}

\paragraph{Our contributions.}
Towards answering the questions above, we introduce a \emph{theoretical framework} to characterize the potentials and limitations of GMLMs, 
for both training and inference.
Precisely, 
\begin{itemize}
\setlength\itemsep{-0.1ex}
    \item The \textbf{asymptotic sample complexity} for estimating the parameters of a distribution via a broad class of masked-prediction losses can be related to the mixing time of a corresponding Markov Chain that can be used to sample from the distribution (\Cref{sec:theory:sample_complexity}). 
    In particular, we prove that training with larger masks always improves statistical efficiency (Theorem~\ref{thm:k_monotone}). 
    \item We show \textbf{finite-sample bounds} that relate how well the \emph{conditional} distributions of the data distribution are learned, to how well the \emph{joint} distribution is learned (\Cref{sec:theory:generalization}) if we have some capacity control over the distribution class being learned (e.g. covering number bounds).
    \item \textbf{Transformers} are only able to represent decoding steps that factorize over the coordinates---preventing them from \textbf{efficiently sampling} even simple distributions 
    with strong correlations between the coordinates (\Cref{sec:theory:sampling}).
    \vspace{-3mm}
\end{itemize}


We accompany these theoretical findings with an extensive set of empirical investigations detailing important components and common error modes. Precisely:  
\begin{itemize}
\item Our experiments (\Cref{sec:experiments}) suggest the \textbf{empirically critical components} include 
large masking ratio (c.f. theory in \Cref{sec:theory:sample_complexity}), 
custom vocabulary,
distillation from AR models,
and architecture improvements like positional attention.~\footnote{
The benefit of distillation was verified in \citet{kim2016sequence, gu2018nonautoregressive, zhou2020understanding, gu2021fully}.
Positional attention was tested in \citet{gu2018nonautoregressive, kreutzer2020inference}.
}
\item GMLMs with parallel-decoding work well on 
{\bf machine translation}: 
in fact, even \emph{one single} forward pass can often produce reasonable translations.
This aligns with our theoretical framework, as machine translation tasks typically involve lower-entropy and less multi-modal outputs, compared to other language generation tasks. 
\item Common {\bf error modes} (``stuttering") 
suggest limitations for parallel-decoding GMLMs for modeling strong dependencies
(c.f. theory in \Cref{sec:theory:sampling}),
which we empirically quantify (\Cref{sec:experiments:attention}). 
\end{itemize} 

Jointly, our theoretical and empirical findings suggest synergistically designing better Markov Chains that mix fast in the presence of strong correlations in the target, 
and corresponding losses that inherit good statistical behavior. 


\section{Theoretical framework}
\label{sec:theory}
We develop a mathematical framework for reasoning about the core ingredients for successfully training and using GMLMs: the \emph{statistical complexity} to learn the model, and the \emph{speed of inference}. 
We show that these two are surprisingly closely related: 
namely, we understand both the asymptotic and finite-sample statistical complexity through \emph{functional inequalities} (e.g. Poincar\'e, approximate tensorization of entropy) corresponding to the Markov Chains we would use at inference time---which in turn characterize the mixing time of these chains. 
This picture closely mirrors an emerging picture in the continuous case for score-based (diffusion) models \citep{koehler2023statistical, qin2023fit}---though with somewhat different proof techniques.

\subsection{Setup and notation}
\label{sec:theory:setup}

The most classical way of fitting distributions from data is maximum likelihood: 
that is, finding the choice of parameters that maximize the likelihood of the training data. There are well-understood statistical reasons to do so: in the asymptotic sense (as the number of sample grows), maximum likelihood is the most sample-efficient way to estimate the distribution \citep{hajek1972local}. 
However, many families of distributions are such that optimizing maximum likelihood is computationally challenging. Thus, many alternate strategies and losses to fit the parameters have been developed.


For continuous distributions, a common choice of loss is the \emph{score matching} loss, where instead of fitting the likelihood, we fit the gradient with respect to the input of the log-pdf, namely $\nabla_x \log p_{\theta}(x)$. In certain cases, this can provable computational benefits over maximum likelihood \citep{pabbaraju2023provable}. 
For discrete distributions, we cannot take gradients with respect to the input: though a closely related strategy is available --- trying to match the conditionals of subsets of variables. (This can be thought of as ``flipping'' the coordinates in the subsets, while keeping the remaining coordinates fixed.) Operationalizing this as a loss gives us the \emph{pseudolikelihood} loss \citep{besag1975statistical}. 

Variants of this strategy have been used in classical results for learning Ising models \citep{ravikumar2010high, vuffray2016interaction}. More recently, this strategy has been used in conjuction with neural models to both learn useful features in the guise of masked language modeling (MLM) \citep{devlin2019bert}, which can be also used to produce a generative model \citep{wang2019bert, goyal2022exposing}. The latter is done by using the learned conditionals inside a Gibbs sampler.
However, when the conditionals are not \emph{consistent}, i.e. there is not a joint distribution that satisfies these conditionals, Gibbs sampler may amplify errors. 
In general, mathematical understanding about sampling from masked language models is still lagging substantially behind.


\paragraph{Setup and notation:} 
Let $\Omega$ be a finite discrete set.
Let $p_{\mathcal{X}}$ denote a distribution over a sequence of $N$ variables $X = (X_1, X_2, \cdots, X_N) \in \Omega^N =: \mathcal{X} $.~\footnote{
In language models, $\Omega$ is the set of tokens in the vocabulary. 
}
Furthermore, for $K \subset [N]$, let $X_K$ denote the subsequence $\left( X_i \, | \, i \in K \right)$,
and $X_{-K}$ denote the subsequence $\left( X_i \, | \, i \notin K \right)$.

We consider learning parameters $\theta$ parametrizing some distribution $p_{\theta}$ over $\mathcal{X}$, for $\theta \in \Theta$. The classical way of fitting $\theta$ is to maximize the likelihood of the training data: 

\begin{definition}[MLE, \cite{van2000asymptotic}]
Given i.i.d.\ samples $x_1,\ldots,x_n \sim p_{\theta^*}$, the maximum likelihood estimator is $\tmle= \arg\max_{\theta \in \Theta} \hat \E \left[ \log p_{\theta}(X)\right]$, 
where $\hat \E$ denotes the expectation over the samples. 
As $n \to \infty$ and under suitable regularity conditions, we have 
$\sqrt{n}\left(\tmle - \theta^*\right) \to N\left(0, \Gamma_{MLE}\right) $
, where $\Gamma_{MLE} := \fisher^{-1}$
, $\fisher \coloneqq \Cov_{X \sim p_{\theta^*}}(\nabla_\theta p_{\theta}(X))_{\vert \theta = \theta^*}$ 
is the Fisher information matrix. 
 \label{def:mle}
\end{definition}
A classical result due to H\'ajek-Le Cam (for modern exposition see \cite{van2000asymptotic})
is that maximum likelihood is the asymptotically most sample-efficient estimator among all ``sufficiently regular'' estimators
(Section 8.5 in \citet{van2000asymptotic})
--- so we will treat it as the ``gold standard'' against which we will compare other estimators. The class of estimators we will be focusing most is the a broad generalization of the \emph{pseudo-likelihood estimator} \citep{besag1975statistical}.

\begin{definition}[Weighted pseudolikelihood]
\label{d:genmple}    
Consider a partition of $[N]$, namely a collection of sets $\mathcal{K} \coloneqq \{ K_1, \ldots, K_{\abs{\mathcal{K}}} \}$ such that 
$\bigcup_i K_i = [N]$, and a distribution $p_{\mathcal{K}}: \mathcal{K} \to \mathbb{R}^+$.

Given $n$ i.i.d samples of sequences and masks: $\{ \left(X^{(i)}, K^{(i)}\right) | X^{(i)} \sim p_{\mathcal{X}}, K^{(i)} \sim p_{\mathcal{K}}\}_{i \in [n]}$,
the \emph{weighted maximum pseudolikelihood estimator (MPLE)} is 
$
\hat{\theta}_{PL} \coloneqq \argmin_\theta 
\sum_{i = 1}^{n} -\log p_\theta (X^{(i)}_{K^{(i)}} | X^{(i)}_{\compl{K^{(i)}}})
.$
The population loss is~\footnote{
This is equivalent to minimizing the KL divergence of the ground-truth conditional distribution $p(X_K | X_{-K})$ from the predicted conditional distribution $p_\theta(X_K | X_{-K})$:
$
\Ep{X \sim p_{\mathcal{X}}, K \sim p_{\mathcal{K}}}{
        \KLp{p(\cdot | X_{-K})}{p_\theta(\cdot | X_{-K})}
}
$
}
$
L_{PL}(\theta) \coloneqq \Ep{X \sim p_{\mathcal{X}}, K \sim p_{\mathcal{K}}}{-\log p_\theta (X_K | X_{\compl{K}})}
$.
\end{definition}

As a special case, if $\mathcal{K}$ contains all subsets of a certain size $k$ and $p_{\mathcal{K}}$ is uniform over $\mathcal{K}$, 
we get the classical $k$-pseudolikelihood estimator:  
\begin{definition}[$k$-pseudolikelihood \citep{huang2002generalized}]
\label{def:mple}
    Same as \Cref{d:genmple} except that 
    $\mathcal{K} \coloneqq \{K \subseteq [N] \mid |K| = k\}$,
    $p_{\mathcal{K}} = \text{Unif}(\mathcal{K})$.
\end{definition}

\begin{remark}
    The distribution of $X$ and $K$ in the above loss is independent. In Section~\ref{sec:theory:sample_complexity:dependent} we will show that our results readily generalize to losses in which the distribution of the masks $K$ can depend on the current $X$. We present the independent case first for ease of presentation.
\end{remark}

Informally, we predict the variables in positions $K \in \mathcal{K}$, conditioned on the remaining variables. 
The benefit is that parametrizing conditionals over smaller subsets $K$ is often computationally cheaper. 
For instance, if $p_{\theta}(x)$ is an undirected graphical model, i.e. $p_{\theta}(x) \propto \exp(\sum_{C} \phi_{C,\theta}(x_C))$, 
where the sum is over all maximal cliques $C$ of the graph describing the distribution, 
the conditional distribution of $K$ only depends on its Markov blanket, which can be very small for sparse graphs and small sets $K$. 
Thus, computing the partition function corresponding to $p(x_K | x_{\compl{K}})$ takes time exponential in this Markov blanket. 
By contrast, computing the likelihood requires calculating the partition function of $p_{\theta}(x)$, which takes time exponential in the dimension of $X$. 
In fact, for Ising models, the corresponding loss is even \emph{convex} \footnote{This fact is well known, but for completeness included in \Cref{sec:theory:optimization}}. 
A similar tradeoff exists for masked language models: fitting the conditionals for larger masks would likely require a larger model, thus would be computationally more expensive.

\subsection{Asymptotic sample efficiency via functional inequalities} 
\label{sec:theory:sample_complexity}
In this section, we will provide a framework for bounding the asymptotic sample complexity of learning the parameters $\theta$ of a discrete probability distribution by minimizing a loss in a broad family of ``masked prediction'' objectives.
We will measure the quality of an estimator in terms of \emph{parameter recovery}. 
To make this formal, we first recall that under mild technical conditions, the estimator will be asymptotically normal:
\begin{lemma}[Asymptotic normality \citep{van2000asymptotic}] \label{lem:asymptotic_efficiency}
    Consider the weighted MPLE objective in \Cref{d:genmple},
    and let 
    $ \theta^* \in \argmin_\theta L_{PL}(\theta) $.
    Under mild regularity conditions (\Cref{lem:asymptotics} in \Cref{sec:appendix:proof:asymptotic}), 
    as $n \to \infty$,
    $\sqrt{n} (\hat{\theta}_{PL} - \theta^*) \xrightarrow{d}$
    $
    \mathcal{N} (0, (\nabla_\theta^2 L_{PL}(\theta^*))^{-1} \Cov(\nabla_\theta l_{PL}(\theta^*)) (\nabla_\theta^2 L_{PL}(\theta^*))^{-1})
    $
\end{lemma}
If we know $\sqrt{n}(\hat{\theta}_{PL} - \theta^*) \xrightarrow{d} \mathcal{N}(0, \Gamma_{PL})$, we can extract bounds on the expected $\ell^2_2$ distance between $\hat{\theta}_n$ and $\theta^*$. Namely, from Markov's inequality, 
(see e.g., Remark~4 in \citet{koehler2023statistical}), 
for sufficiently large $n$, with probability at least $0.99$ it holds that 
$
\| \hat{\theta}_{PL} - \theta^* \|_2^2 \leq \frac{\Tr(\Gamma_{PL})}{n}.
$

\subsubsection{Masking more is (statistically) better}

As a first application of our framework, we prove that increasing the number of variables $k$ we predict in $k$-pseudolikelihood (Definition~\ref{def:mple}) strictly improves the statistical efficiency of the resulting estimator. 
Note, for larger $k$, we expect the computational cost to optimize the corresponding loss to be larger, 
and when $k = N$ we just get max likelihood. 
Thus, this naturally formalizes a \emph{computational/statistical tradeoff} in choosing $k$.   

\begin{assumption}[Finite gradient and Hessian] 
\label{as:bounded}
$\forall \theta \in \Theta, \forall x \in \mathcal{X}, K \subset [N]$,
the norms of the gradient
$\| \nabla_\theta \log p_\theta (x_K | x_{\compl{K}} ) \|_2$ 
and the Hessian
$\| \nabla_\theta^2 \log p_\theta (x_K | x_{\compl{K}} ) \|_F$
exist and are finite
.
\end{assumption}

\begin{assumption}[Realizability] 
\label{as:realizability}
The data distribution $p_{\mathcal{X}}$ satisfies: $\exists \theta^* \in \Theta$
such that 
$p_{\theta^*} = p_{\mathcal{X}}$. 
\end{assumption}

\begin{restatable}[Masking more is (statistically) better]{thm}{thmCompStatTradeoffMask} 
\label{thm:k_monotone}
Let \Cref{as:bounded} and \Cref{as:realizability} be satisfied, and  
let $\Gamma_{PL}^k$ denote the asymptotic variance of the $k$-MPLE estimator (\Cref{def:mple}).
Then, we have:\footnote{
The notation $A \preceq B$ means $B - A$ is positive semidefinite.
}
$\Gamma_{PL}^{k+1} \preceq \Gamma_{PL}^k.$
\end{restatable}

\begin{remark}
    By monotonicity of trace, Thm~\ref{thm:k_monotone} implies
    $\Tr(\Gamma_{PL}^{k+1}) \leq \Tr(\Gamma_{PL}^k)$.
    By the remarks after Lemma~\ref{lem:asymptotic_efficiency},
    larger $k$ implies a better asymptotic $l_2$ bound for learning $\theta$
    since
    $
    \Ep{X_{1:n}, K_{1:n}}{\| \hat\theta_{PL}^k - \theta \|_2^2} \rightarrow \frac{\Tr(\Gamma^k_{PL})}{n}
    $.
\end{remark}


The main lemma needed for Theorem~\ref{thm:k_monotone} is that the two matrices in the asymptotic covariance of MPLE, 
$\hess_\theta L_{PL}(\theta^*)$ and  $\Cov_{X \sim p_{\mathcal{X}}, K \sim p_{\mathcal{K}}}(-\nabla_{\theta} \log p_\theta (X_K | X_{\compl{K}} ))_{\vert \theta = \theta^*}$ are actually equal. For MLE (namely, when $k=N$)
this is well-known and called the \emph{information matrix equality}.
Proofs of \Cref{l:informationmx} and \Cref{thm:k_monotone} are in \Cref{sec:appendix:proof:generalized_information_matrix} and \Cref{sec:appendix:proof:k_monotone}.
We empirically verify \Cref{thm:k_monotone} in \Cref{sec:experiments:ising}.
\begin{restatable}[Generalized information matrix equality]{lemma}{lemGeneralizedInformationMatrixEquality} 
\label{l:informationmx}
Under \Cref{as:bounded}
and \Cref{as:realizability},
the weighted pseudolikelihood loss (\Cref{d:genmple}) verifies: 
$
\nabla_\theta^2 L_{PL}(\theta^*) = \Cov_{X \sim p_{\mathcal{X}}, K \sim p_{\mathcal{K}}}(-\nabla_{\theta} \log p_\theta (X_K | X_{\compl{K}} ))_{\vert \theta = \theta^*}.$
\end{restatable}

\subsubsection{Statistical efficiency  bounds via mixing time bounds}
\label{s:asymptotic-independent}
We could in general conceive of masking strategies where certain subsets of variables get masked with different probabilities. For instance, in language, nearby words will tend to be more correlated; grammatical constraints will dictate the parts-of-speech that can occur in different positions.
We would then like to have theoretical guidance on what choices of masking distributions are better. 
Remarkably, it turns out that we can relate the statistical efficiency --- in the sense of $\mathbb{E}\|\hat{\theta} - \theta^*\|^2$ for the resulting estimator $\hat{\theta}$ --- and the mixing time of an appropriately chosen Markov Chain. In fact, this is the Markov Chain that would be typically chosen at inference time.
Towards making this formal, we will need several preliminary concepts and results for Markov chains. Recall, a Markov chain on a state space $\Omega$ is described by a (row-stochastic)
transition matrix $P$. Moreover, we can assign a natural bilinear form called the Dirichlet form: 

\begin{definition}[Dirichlet form] 
\label{def:dirichlet}
    Let $M$ be an ergodic, reversible Markov chain with transition matrix $P$ on state space $\Omega$. Let $\mu$ be its unique stationary distribution. 
    $\forall f, g: \Omega \rightarrow \mathbb{R}$
    the associated Dirichlet form is defined as:
    $$
    \mathcal{E}_P (f, g) := \langle f, (I - P) g \rangle_{\mu}
    = \frac{1}{2} \Sigma_{x, y \in \Omega} \mu(x) P(x, y) (f(x) - f(y))(g(x) - g(y)) 
    $$
\end{definition}

Mixing time of the Markov chain can be bounded in the $\chi^2$ sense by the gap between the 1st and 2nd eigenvalue of the Laplacian matrix $I-P$,
expressed as Poincar\'e inequality: 

\begin{definition}[Poincar\'e inequality]
    \label{def:poincare}
    We say that a Markov chain satisfies a \emph{Poincar\'e inequality} with constant $C$ if for all $f: \Omega \rightarrow \mathbb{R}$, we have $\mathcal{E}_P (f, f) \ge \frac{1}{C} \Var_{\mu}(f).$
\end{definition}

The Poincar\'e inequality implies exponential ergodicity of the Markov chain in $\chi^2$-divergence, precisely $\displaystyle   \chi^2(p_t, \mu) \le e^{-2t/C} \chi^2(p_0, \mu)$,
where $\mu$ is the stationary distribution of the chain and $p_t$ is the distribution after running the Markov process for time $t$, starting at $p_0$.
We will be particularly interested in several generalizations of Gibbs sampling:  

\begin{definition}[Weighted block dynamics]
    Let $\mathcal{K} \coloneqq \{ K_1, \ldots, K_{\abs{\mathcal{K}}} \}$ be a collection of sets (or blocks) such that $\bigcup_i K_i = [N]$. 
    A block dynamics with blocks $\mathcal{K}$ is a Markov chain that picks a block $K$ in each step according to some distribution $p_{\mathcal{K}}: \mathcal{K} \to \mathbb{R}^+$~\footnote{
    This is analogous to the training objective setting in \Cref{d:genmple}.
    } 
    and then updates the coordinates in $K$ according to the conditional distribution $p_{\mathcal{X}}(X_K | X_{-K})$. 
\label{d:blockdynamics}
\end{definition}

The stationary distribution for the above Markov Chain is $p_{\mathcal{X}}$. \citet{caputo2021block} also derived the Dirichlet form (\Cref{def:dirichlet}) corresponding to this Markov chain: 
\[
\mathcal{E}(f, g) := \mathbb{E}_{K \sim p_{\mathcal{K}}} \Ep{X_{\compl{K}}}{\text{Cov}_{X_K|X_{\compl{K}}} (f, g)}.
\]
The crucial result we show is that the statistical efficiency of the weighted MPLE (Definition~\ref{d:genmple}) as captured by the asymptotic variance can be related to the Poincar\'e constant of the corresponding weighted block dynamics (Definition~\ref{d:blockdynamics}).  
Proof of \Cref{thm:asymptotic_variance_alpha} is in \Cref{sec:appendix:proof:asymptotic_variance}.

\begin{restatable}[Asymptotic variance under a Poincar\'e Inequality]{thm}{thmAsymptoticVariance} 
\label{thm:asymptotic_variance_alpha}
Suppose the distribution $p_{\theta^*}$ satisfies a Poincar\'e inequality with constant $C$ with respect to the weighted block dynamics. Then under \Cref{as:bounded} and \Cref{as:realizability}
the asymptotic variance of the weighted MPLE can be bounded as:  
$
  \Gamma_{PL} \preceq C \fisher^{-1}
$
where $\fisher$ is the Fisher Information matrix
(\Cref{def:mle}).
\end{restatable}

\subsubsection{Adaptive masking: masked positions depend on the sequence}
\label{sec:theory:sample_complexity:dependent}

The machinery we developed in Section~\ref{s:asymptotic-independent} is in fact substantially more general --- it applies to even ``adaptive'' masking losses in which the conditional distribution of the mask can depend on the current $X$ (that is, for each $X$, there is a different conditional distribution $p_{\mathcal{K}}(K | X)$ which can be manually designed and is known to the model during training).

\begin{definition}[Adaptively weighted pseudolikelihood]
\label{d:genmple:dependent}    

Given $n$ i.i.d samples of sequences and masks: $\{ \left(X^{(i)}, K^{(i)}\right) | X^{(i)} \sim p_{\mathcal{X}}, K^{(i)} \sim p_{\mathcal{K}}(\cdot | X^{(i)})\}_{i \in [n]}$,
the \emph{weighted maximum pseudolikelihood estimator (MPLE)} is 
$
\hat{\theta}_{PL} \coloneqq \argmin_\theta 
\sum_{i = 1}^{n} -\log p_\theta (X_K | X_{\compl{K}},K)
$, where 
\begin{align}
        p_\theta (X_K | X_{\compl{K}},K) := \frac{p_\theta (X) p_{\mathcal{K}}(K | X)}{\sum_{X'_K} p_{\theta}\left(X'_K, X_{-K}\right) p_{\mathcal{K}}(K | \left(X'_K, X_{-K}\right))} \label{eq:defnconditional}
\end{align}
The population loss is correspondingly:
\[
L_{PL}(\theta) \coloneqq \mathbb{E}_{X \sim p_{\mathcal{X}}} \mathbb{E}_{K \sim p_{\mathcal{K}}(\cdot | X)}\left[-\log p_\theta (X_K | X_{\compl{K}},K)\right]
.\]
\end{definition}
\begin{remark} \label{rem:parameterization}
    Note, the distribution $p_{\mathcal{K}} (\cdot | X)$ doesn't depend on $\theta$, so $K^{(i)}$ can be generated readily by drawing samples from this distribution. Note also, the term $p_\theta (X_K | X_{\compl{K}},K)$ is expressible in terms of the joint distribution $p_{\theta}(X)$ and $p_{\mathcal{K}}(\cdot | X)$ and the expression in \eqref{eq:defnconditional} can be interpreted as a conditional distribution in the joint distribution $p_{\theta, \mathcal{K}}(X,K) := p_{\theta}(X) p_{\mathcal{K}}(K|X)$. Finally, note that conditioning on the set $K$ is subtle, but important --- see \Cref{lem:XS_implies_S} in \Cref{sec:appendix:proof:adaptive}.  
\end{remark}

We can analogously generalize the sampling process to the following Markov chain
in which $K$ is sampled dependent on $X$:

\begin{definition}[Adaptive weighted block dynamics]
    Let $\mathcal{K} \coloneqq \{ K_1, \ldots, K_{\abs{\mathcal{K}}} \}$ be a collection of sets (or blocks) such that $\bigcup_i K_i = [N]$. 
    A block dynamics with blocks $\mathcal{K}$ is a Markov chain that picks a block $K$ in each step according to some distribution~\footnote{
    This is analogous to the training objective setting in \Cref{d:genmple:dependent}.
    } 
    $p_{\mathcal{K}}(\cdot \mid X)$, 
    and then updates the coordinates in $K$ according to the conditional distribution $p_{\mathcal{X}}(X_K | X_{-K}, K)$.\footnote{Defined analogously as in Definition~\ref{eq:defnconditional}.}
\label{d:blockdynamics_adaptive}
\end{definition}

If we understand the domain of this Markov chain to be
$\{ (X, K) \mid X \in \mathcal{X}, K \in \mathcal{K} \}$, its stationary distribution is 
$$p_{\mathcal{X}, \mathcal{K}}(X,K) := p_{\mathcal{X}}(X) p_{\mathcal{K}}(K \mid X).$$ 
The Dirichlet form can also be explicitly written down (note, $f$ and $g$ are functions of both $X$ and $K$): 
\begin{restatable}[Dirichlet form for adaptive weighted block dynamics]{proposition}{propBlockDynamicsDirichlet} 
\label{prop:block_dynamics_dirichlet}
The Dirichlet form corresponding to the weighted block dynamics (\Cref{d:blockdynamics_adaptive}) is:
\begin{equation*}
    \mathcal{E}(f, g) 
    = \mathbb{E}_{(X_{\compl{K}}, K) \sim p_{\mathcal{X}, \mathcal{K}}} \left[ \text{Cov}_{X_K |  (X_{\compl{K}}, K)}(f, g) \right]
\end{equation*}
\end{restatable}

The proof of \Cref{prop:block_dynamics_dirichlet} is in \Cref{sec:appendix:proof:block_dynamics_dirichlet}.

Analogous to \Cref{thm:asymptotic_variance_alpha}, we again show that the statistical efficiency of the adaptively-weighted MPLE (Definition~\ref{d:genmple:dependent}), 
captured by the asymptotic variance, 
can be related to the Poincar\'e constant of the corresponding adaptively-weighted Block dynamics (Definition~\ref{d:blockdynamics_adaptive}):

\begin{restatable}[Asymptotic variance of adaptively-weighted MPLE under a Poincar\'e Inequality, generalization of \Cref{thm:asymptotic_variance_alpha}]{thm}{thmAsymptoticVarianceAdaptive} 
\label{thm:asymptotic_variance}
Suppose the distribution $p_{\theta^*}$ satisfies a Poincar\'e inequality with constant $C$ with respect to the adaptively-weighted block dynamics. Then 
under \Cref{as:bounded} and \Cref{as:realizability} where $p_\theta (x_K | x_{\compl{K}} )$ is replaced by $p_\theta (x_K | x_{\compl{K}}, K )$,
the asymptotic variance of the adaptively-weighted MPLE can be bounded as:  
$
  \Gamma_{PL} \preceq C \fisher^{-1}
$
where $\fisher$ is the Fisher Information matrix
(\Cref{def:mle}).
\end{restatable}

The proof of \Cref{thm:asymptotic_variance} is in \Cref{sec:appendix:proof:asymptotic_variance}.

\subsection{Finite sample bounds and distributional distance}
\label{sec:theory:generalization}
\vspace{-1mm}
The framework in Section~\ref{sec:theory:sample_complexity} was asymptotic in nature, and used parameter closeness as a notion of ``quality'' of the estimator. In this section, we remove both requirements, at the cost of the bounds depending on a notion of ``complexity'' of the parametric class we are fitting. It turns out that we can prove very similar results, with the notion of ``mixing'' --- as captured by the Poincar\'e constant --- being replaced by a different constant called the ``approximate tensorization constant''. These results mirror results in Section 5.1 in \cite{koehler2023statistical}, who focus on 1-MPLE and use a differrent notion of ``complexity'' based on Rademacher complexity. We first introduce several preliminary concepts. 



\begin{definition}[Block approximate tensorization of entropy \citep{caputo2021block}] \label{d:block_catconst}
Under fixed distribution $p_{\mathcal{K}}(\cdot \mid X)$ over binary masks $\mathcal{K}$ conditioned on $X$,
we say the distribution $q_{\mathcal{X}}$ over $\mathcal{X}$ satisfies \emph{block-generalized approximate tensorization of entropy}
with constant $\bar{C}_{AT}(q_{\mathcal{X}})$ if
for any distribution $r_{\mathcal{X}}$ over $\mathcal{X}$,
\begin{equation*}
\KL(r_{\mathcal{X}}, q_{\mathcal{X}}) \le \bar{C}_{AT}(q_{\mathcal{X}}) \cdot \E_{X \sim r_{\mathcal{X}}}[
    \E_{K \sim p_{\mathcal{K}}(\cdot \mid X)}[\KLp{r_{\mathcal{X}}(\cdot \mid X_{-K}, K)}{q_{\mathcal{X}}(\cdot \mid X_{-K}, K)}
    ]
]
\end{equation*}
\end{definition}

This inequality is closely related to the mixing time of weighted-block dynamics (Definition~\ref{d:blockdynamics}). Namely, the inequality is weaker than the standard discrete version of the log-Sobolev inequality \citep{diaconis1996logarithmic} and stronger than the Modified Log-Sobolev Inequality \citep{bobkov2006modified}, which implies exponential ergodicity of the weighted block-dynamics in KL divergence\footnote{This in turns, also implies a Poincar\'e inequality and exponential ergodicity in $\chi^2$ divergence.}, that is: $\displaystyle \mbox{KL}(p_t, q) \le e^{-2t/C_{AT}(q)} \mbox{KL}(p_0, q).$


To bound the distance between the population and empirical losses, as well as relate it to the distance between the estimated parameters and the ground truth, we first introduce a few useful pieces of notation. 

\paragraph{Notation:} For each sample $X^{(i)}, i \in [n]$, we assume we observe $m$ masks $\{K_j^{(i)} \mid j \in [m]\}$ sampled iid from $p_\mathcal{K}(\cdot \mid X^{(i)})$. 
We denote the corresponding empirical loss by $$\hat{L}_{PL}(\theta) \coloneqq \frac{1}{n m}\sum_{i = 1}^{n} \sum_{j = 1}^{m} -\log p_\theta (X^{(i)}_{K_j^{(i)}} | X^{(i)}_{\compl{K_j^{(i)}}}, K_j^{(i)}).$$
Furthermore, we will denote by $\tilde{p}_{\mathcal{X}}$ the uniform distribution over $\{X^{(i)}, i \in [n]\}$,
and denote
\begin{equation}
\tilde{L}_{PL}(\theta) \coloneqq 
\Ep{X \sim \tilde{p}_{\mathcal{X}}, K \sim p_\mathcal{K}(\cdot \mid X)}{
    -\log p_\theta (X_{K} | X_{\compl{K}}, K)
}
\label{eq:intermetiatepl}
\end{equation}
This is an intermediate quantity: it averages in the population sense over the masks, but it assumes a finite number of samples from $p_{\mathcal{X}}$. It will be a useful intermediate quantity for several concentration bounds.

We will also need a few mild assumptions on the distribution we are fitting. First, we assume that the learned conditional probabilities are uniformly lower-bounded by a constant:
\begin{assumption}[Support margin] \label{as:support_margin}
Exists constant $\margin \in (0, 1)$ s.t.
$\forall X \in \mathcal{X}, \forall K \subset [N]$,
$\forall \theta \in \Theta$,
if $p_\mathcal{X}(X_K | X_{-K}, K) > 0$,
then $p_\theta(X_K | X_{-K}, K) \ge \margin$.
\end{assumption}

We also assume that the log-probabilities (and hence the losses $\hat{L}_{PL}$ and $\tilde{L}_{PL}$) are Lipschitz with respect to $\theta$, and $\Theta$ has a finite covering bound. Namely: 
    
\begin{assumption}[Covering bound and Lipschitzness] \label{as:complexity}
$\forall \epsilon > 0$,
there exists a finite partition $\text{Par}_\epsilon(\Theta) = \{ \Theta_1, \cdots, \Theta_{\abs{\text{Par}(\Theta)}} \}$ of $\Theta$, such that
$\forall i, \forall \theta_1, \theta_2 \in \Theta_i$,
and $
\forall (X,K) \in \mathcal{X} \times \mathcal{K}$:  
\[
\left|
\log p_{\theta_1} (X_{K} | X_{\compl{K}}, K)
- 
\log p_{\theta_2} (X_{K} | X_{\compl{K}}, K)
\right| \le \frac{\epsilon}{2}.\]

Moreover, $C_\epsilon(\Theta)$ denote the smallest possible cardinality among such partitions $\text{Par}_\epsilon(\Theta)$.
\end{assumption}

With this setup, we can prove the following finite-sample bound on the closeness of the learned distribution, provided the weighted pseudolikelihood loss (Definition \ref{d:genmple}) is small:

\begin{restatable}[Generalization bound for learning the joint distribution]{thm}{thmGeneralizationBoundJoint} 
\label{thm:generalization_bound_joint}
Let $\hat\theta \coloneqq \argmin_{\theta} \hat{L}_{PL}(\theta)$.
Under \Cref{as:support_margin} and \Cref{as:complexity},
$\forall \epsilon > 0$,
$\forall \delta \in (0, 1)$,
with probability at least $1 - \delta$
we have 
$
\TVp{p_{\hat\theta}}{p_{\mathcal{X}}}
< \sqrt{\frac{1}{2} \bar{C}_{AT}(p_{\hat\theta})
\left( \hat{L}_{PL}(\hat\theta) + B \cdot \ln \frac{1}{\margin} + \epsilon \right) + C
}
$
where
$B = \sqrt{\frac{2^{3N} \abs{\Omega}^N C_\epsilon(\Theta)}{m \cdot \delta}} +  \sqrt{\frac{\ln\frac{8 C_\epsilon(\Theta)}{\delta}}{2 n}}$,
and
$C = \sqrt{\frac{\abs{\Omega}^{3N}}{8 \delta n}}$
.
\end{restatable}

Proof of \Cref{thm:generalization_bound_joint} is in \Cref{sec:appendix:proof:generalization_bound_joint}.
We can compare the statement to Theorem \ref{thm:asymptotic_variance}:
(1) On the LHS, rather than parameter distance, we have total variation distance between the learned distribution and $p$. 
(2) On the RHS, rather than a Poincar\'e inequality, we have the $\bar{C}_{AT}(p_{\hat\theta})$ constant. 
(3) On the RHS, instead of the Fisher information matrix, we have quantities capturing the generalization error, through a notion of complexity of the class ($C_{\epsilon}(\Theta)$). 

\subsection{Inference-time limitations due to parallelism}
\label{sec:theory:sampling}

In this section, we focus on limitations in the representational and computational efficiency that arise when using a parallel decoding approach to implement a step of the inference-time sampling algorithm. Precisely, at inference-time, using weighted block-dynamics with bigger blocks enables larger sets of coordinates to be re-randomized, facilitating a faster mixing time. A canonical example of this are $k$-Gibbs samplers:  

\begin{definition}[$k$-Gibbs sampler]
    The $k$-Gibbs sampler is a special case of the block dynamics (\Cref{d:blockdynamics}) when 
    $\mathcal{K} \coloneqq \{K \subseteq [N] \mid |K| = k\}$,
    and $p_{\mathcal{K}} = \text{Unif}(\mathcal{K})$.    \begin{equation} \label{eq:gibbs:k}
        X_K^{(t+1)} \sim p\left(\cdot \mid X_{-K}^{(t)}\right), \; X_{-K}^{(t+1)} = X_{-K}^{(t)}
    \end{equation}
    \label{d:kgibbs}
\end{definition}
Samplers with larger $k$ are well-known to mix faster (e.g., \cite{lee2023parallelising} shows the Poincar\'e inequality improves by a factor of at least $k$). However, taking a step of this Markov Chain requires being able to re-randomize the $k$ coordinates according to their conditional distribution --- which is intuitively harder for larger $k$ if we are trying to re-randomize the coordinates in parallel. 

In \Cref{sec:theory:sampling:transformers}, we show that the canonical GMLM-type parallel decoding language models can only implement Markov chains whose transitions are \emph{product distributions} over the sequence positions.~\footnote{
\Cref{rem:mechanistic} in \Cref{sec:appendix:proof:transformer_ddlm_expressive_power} connects our results to technical details of model architectures in prior works.
} 
We also consider a natural Markov Chain whose transitions are product distributions, and show it can be substantially slower to reach the modes of the distribution compared to $k$-Gibbs for a large $k$ (i.e. a Markov Chain with transitions that are far from a product distribution). Precisely, we consider: 

\begin{definition}[Independent parallel sampler]
\label{item:gibbs:parallel} The \emph {independent parallel sampler} performs \emph{coordinate-wise} updates for all $i$ in parallel~\footnote{
    The stationary distribution of this chain is unclear: 
    in fact, it is not even clear the chain is ergodic.
    }, namely:
    \begin{equation} \label{eq:gibbs:parallel}
        \forall i \in [N], \; X_i^{(t+1)} \sim p\left(\cdot \mid X_{-\{i\}}^{(t)}\right)
    \end{equation}
\end{definition}
In \Cref{sec:theory:sampling:gibbs}, we show that even if we do not care about mixing---just reaching the modes of the distribution---the independent parallel sampler can be much slower compared to $k$-Gibbs, for a large $k$. 

\subsubsection{Which Markov Chains are implementable via parallel decoding?}
\label{sec:theory:sampling:transformers}

In this section, we characterize the power and restrictions of Transformers at inference time when they are restricted to decoding the tokens of the sequence in parallel.
The inference algorithms for a model that has access to approximate conditional probabilities typically look like (potentially multiple) steps of block dynamics (Definition~\ref{d:blockdynamics}). We focus on understanding what kinds of transitions are implementable with a standard Transformer architecture. 


Note that while there are well-known prior results about the expressive power of Transformers as sequence-to-sequence modelers \citep{yun2020are}, 
representing steps of a Markov Chain with parallel decoding is more subtle, due to the fact that a step of a Markov Chain requires randomness. 
First, we state a result characterizing the power of Transformers to approximate ``deterministic'' Markov Chains: that is, Markov Chains whose transition distributions are delta functions. Unsurprisingly, standard universal approximation results can be adapted easily to this case. Namely: 

\begin{restatable}[informal]{proposition}{propTransformersDeterministicMC}
\label{prop:transformer_deterministic_mc}
Transformers (with sufficient depth and width) can implement any number of transitions of any deterministic Markov Chain over sequences in $\Omega^N$.
\end{restatable}

On the other hand, Transformers using parallel decoding cannot implement general Markov chains over $\Omega^N$. In fact, they can only implement Markov Chains for which the transition probabilities are product distributions:

\begin{restatable}[informal]{proposition}{propTransformersGeneralMC}
\label{prop:transformer_general_mc}
The class of Markov chains over sequences in $\Omega^N$ implementable by (sufficiently wide and deep) Transformers is
those whose next-state transition probability distributions are product distributions over the positions, conditioned on the current state.
\end{restatable}
Background information on the Transformer architecture,  
as well as proofs of 
formalized versions of \Cref{prop:transformer_deterministic_mc} and \Cref{prop:transformer_general_mc}
are relegated to  \Cref{sec:appendix:proof:transformer_ddlm_expressive_power}.
Note that this does \emph{not} mean one can only simulate Markov Chains whose \emph{stationary} distribution is a product distribution. 
In fact, the standard $1$-Gibbs sampler, by virtue of the fact that it only updates one coordinate at a time, encodes a product distribution for each transition. On the other hand, under fairly mild conditions on a joint $p$, the $1$-Gibbs sampler corresponding to $p$ is ergodic and has $p$ as a stationary distribution. On the other hand, a step of a $k$-Gibbs sampler for $k>1$ is in general \emph{not} a product distribution, and will not be implementable by a Transformer with parallel decoding.

\subsubsection{Markov Chains with dependent transitions can be (much) faster}
\label{sec:theory:sampling:gibbs}

In this section, we show that the $k$-Gibbs (Definition~\ref{d:kgibbs})---the prototypical example of a Markov Chain with dependent transitions---can reach modes of the distribution much faster than the independent parallel sampler (Definition~\ref{item:gibbs:parallel})---the prototypical example of a Markov Chain with independent transitions. Intuitively, in cases where there is a strong dependence between subsets of variables, jointly updating them will bring us much faster to their 
modes.

The toy probabilistic family in which we will illustrate this phenomenon is \emph{Ising models}, a canonical example of an undirected model in which ``dependence'' between the variables can be easily modulated. Precisely: 

\begin{definition}[Ising models]
An \emph{Ising model} is a distribution $p: \{\pm 1\}^{N} \to \mathbb{R}^+$, defined by a graph $G = (V,E)$ with $|V| = N$  and parameters $\left\{\mJ_{\{i,j\}}: \{i,j\} \in E\right\}$ and $\vh \in \R^N$
such that:  
\begin{equation} \label{eqn:ising}
    p_{G}(X = x) \propto \exp\left(\sum_{i \in [N]} \vh_i x_i + \sum_{\{i, j\} \in E} \mJ_{\{i,j\}} x_i x_j\right),
\end{equation}
\label{def:ising}
\end{definition}

For the results in this section, we will consider a graph $G$ that consists of a union of a clique $C_G$ (in which $\abs{C_G} \ge 2$, and the pairwise interactions among the variables are strong) and a set of $N-|C_G|$ independent vertices.
More formally, we consider: 
\begin{equation} \label{eqn:ising:construction}
    p_G(X=x) \propto \exp{\left( \sum_{i \in [N]} \vh_i x_i + \sum_{i \ne j, i, j \in C_G} J x_i x_j \right)}
\end{equation}
such that $\sum_{i \in C_G} \vh_i > 0$ and $J > 0$.
This is a \emph{ferromagnetic} Ising model (i.e. the pairwise interactions prefer the variables to have the same value). Moreover, when $J \gg \| \vh \|_1$,
the distribution $p_G$ has two ``modes'', in which all variables in $C_G$ have the same value: 
\begin{align}
    \mathcal{R}_1 &\coloneqq \{ x \in \{-1, 1\}^N | x_i = 1 \, \forall i \in C_G \} \label{eq:larger_mode}  \\
    \mathcal{R}_{-1} &\coloneqq \{ x \in \{-1, 1\}^N | x_i = -1 \, \forall i \in C_G \} \label{eq:smaller_mode}
\end{align}

The above distribution can be seen as a toy model of language tasks in which 
grammatical rules or semantic constraints create ``clusters'' of positions in which changing isolated words leads to very unlikely sentences.   
Next, we formalize the concentration around the ``modes'': 

\begin{assumption}[Strongly ferromagnetic Ising model] \label{as:strongly_ferromagnetic}
There exist constants $h_G > 0, J_0 > 0$ such that
$h_G \coloneqq \sum_{i \in C_G} \vh_i > \sum_{i \notin C_G} \abs{\vh_i}$, $J - \| \vh \|_1 \ge J_0$.
\end{assumption}

Informally, under \Cref{as:strongly_ferromagnetic},
sequences in $\mathcal{R}_{1}$ are much more likely under the groundtruth distribution than those in $\mathcal{R}_{-1}$,
which are further much more likely than all other sequences.
The formal statement and proof are in \Cref{sec:appendix:proof:strongly_ferromagnetic_mode}.
As a result, we can think of sampling from $\mathcal{R}_{1}$ as analogous to sampling a high-quality sentence,
and moreover, not reaching $\mathcal{R}_{1}$ implies the Markov chain sampling process has not mixed to the groundtruth distribution yet. Continuing the analogy to language tasks, in tasks like machine translation,
for each source sentence, 
sampling one high-quality target sentence is potentially good enough.
In some other tasks like creative writing,
producing well-calibrated samples might be desirable---so mixing would be needed.

First, we show that running the \textbf{$k$-Gibbs sampler} requires a small number of steps to reach $\mathcal{R}_{1}$.
This implies that if a model can efficiently approximate one step of \textbf{$k$-Gibbs sampler},
then it is fast to sample a high-probability sequence by iteratively applying the model.
Proof is in \Cref{sec:appendix:proof:mode_fast}.

\begin{restatable}[$k$-Gibbs sampler can reach the mode fast]{proposition}{propModeFast} 
\label{prop:mode_fast}
Consider the Ising model in \Cref{eqn:ising:construction} satisfying \Cref{as:strongly_ferromagnetic}. Let us denote 
$(0,1) \ni
c_{\mathcal{R}_{1}} \coloneqq 1 - \frac{{N-|C_G| \choose k-|C_G|}}{{N \choose k}} \frac{e^{2 (J_0 + h_G)}}{e^{2 (J_0 + h_G)} + e^{2 J_0} + 2^{\abs{C_G}} - 2}
.$

Then,  
for any initial $\mX^{(0)}$ and 
$\delta \in (0, 1)$,
with probability at least $1 - \delta$,
after $T \coloneqq \ceil{\log_{c_{\mathcal{R}_{1}}} \delta}$ steps of \textbf{$k$-Gibbs sampler} (\Cref{d:kgibbs}) with $k \ge |C_G|$,
we have
$\{ \mX^{(t)} | t \in [T] \} \cap \mathcal{R}_{1} \ne \emptyset.$
\end{restatable}

By contrast, we show that for nontrivial probability over the randomness in the initial sequence, 
running \textbf{independent parallel} requires a large number of steps to reach the largest mode $\mathcal{R}_{1}$ of the distribution, which implies that the sampling process may not quickly reach a high-probability sequence.

\begin{restatable}[Independent parallel sampling stuck in bad samples]{proposition}{propModeSlow} 
\label{prop:mode_slow}
Consider the Ising model in \Cref{eqn:ising:construction} satisfying \Cref{as:strongly_ferromagnetic}. Let us denote 
$
c_{\text{stuck}} \coloneqq \frac{2 \left( -1 + \frac{1 - \exp{\left( - 2 J_0 \right)}}{\exp{\left( - 2 J_0 \right)} + 1} \frac{\abs{C_G}}{2} \right)^2}{\abs{C_G}}
$.

For an initial $\mX^{(0)}$ such that $\sum_{i \in C_G} \mX_i^{(0)} \le -2$, for any
$\delta \in (0, 1)$,
with probability at least $1 - \delta$,
after $T \coloneqq \floor{\frac{\delta}{2} \exp{(c_{\text{stuck}})}}$ steps of \textbf{independent parallel} (\Cref{item:gibbs:parallel}),
we have
$\forall t \in [T], \sum_{i \in C_G} \mX_i^{(t)} \le -2$.
\end{restatable}

The proof is in \Cref{sec:appendix:proof:mode_slow}.
Combining \Cref{prop:mode_fast} and \Cref{prop:mode_slow} leads to a separation result between \textbf{$k$-Gibbs sampler} and \textbf{independent parallel},
in particular when the clique size in $G$ is large 
and dependency is strong within the clique:
with high probability,
while the former reaches $\mathcal{R}_{1}$ in 1 step,
the latter cannot do so in arbitrarily large number of steps:

\begin{assumption}[Strong interactions in Ising model] \label{as:strong_interactions}
On Ising model $G$ in \Cref{eqn:ising:construction}, for parameters $\delta \in (0, 1)$ and $M \in \N_+$, 
\begin{align*}
    \abs{C_G} &\ge 8 \left( 1 + \ln{\frac{4 M}{\delta}} \right) \\
    h_G &\ge \frac{1}{2} \ln{\frac{2 (4-\delta)}{\delta}} \\
    J_0 &\ge \frac{1}{2} \abs{C_G} \ln{2}
\end{align*}
\end{assumption}

\begin{assumption}[Large coordinate set per update] \label{as:large_k}
When running the $k$-Gibbs sampler (\Cref{d:kgibbs}) on Ising model $G$ in \Cref{eqn:ising:construction}, 
we assume $k$ is large wrt 
parameter $\delta \in (0, 1)$: 
\begin{align*}
    k \ge \max \{ \abs{C_G}, N - \delta \frac{N+1}{(4-\delta) \abs{C_G} + \delta} \}
\end{align*}
\end{assumption}

\begin{remark}
\Cref{as:large_k} requires $k$ to be not much smaller than $N$.
When $N$ is small and $\delta \approx 0$, \Cref{as:large_k} essentially requires $k = N$.
When $N \gg \abs{C_G}$, \Cref{as:large_k} allows a larger gap $N - k$.
\end{remark}

\begin{restatable}[Separation between $N$-Gibbs sampler and independent parallel sampling]{corollary}{propModeSeparation} 
\label{prop:mode_separation}
On Ising model $G$ in \Cref{eqn:ising:construction} under \Cref{as:strongly_ferromagnetic}, 
$\forall \delta \in (0, 1)$,
$\forall M \in \N_+$,
If $G$ additionally satisfies \Cref{as:strong_interactions}
and \Cref{as:large_k}
and the initial $\mX^{(0)}$ is such that $\sum_{i \in C_G} \mX_i^{(0)} \le -2$,
then with probability at least $1 - \delta$,
\begin{enumerate}
    \item Running the \textbf{$k$-Gibbs sampler}: $\mX_{\text{k-Gibbs}}^{(1)} \in \mathcal{R}_{1}$, and
    \item Running \textbf{independent parallel}: $\{ \mX_{indep}^{(t)} | t \in [M] \} \cap \mathcal{R}_{1} = \emptyset$
\end{enumerate}
\end{restatable}

Proof is in \Cref{sec:appendix:proof:mode_separation}.
We empirically verify our theory in \Cref{sec:appendix:experiments:synthetic_sampling}.


\section{Experiments}
\label{sec:experiments}

\subsection{Synthetic experiments on Ising model}
\label{sec:experiments:ising}

To empirically validate \Cref{thm:k_monotone} (Masking more is (statistically) better), 
we run controlled experiments with synthetic data generated by a ground-truth Ising model.
We train an Ising model using $k$-pseudolikelihood, and measure the squared error of parameter estimation. 
The results verify that with the same training data size, larger $k$ leads to lower error. 
We plot the results in \Cref{fig:ising_k_monotone_flat},
with several related experiments, in \Cref{sec:appendix:experiments:synthetic}.~\footnote{
Related simulations were also reported in \citet{huang2002generalized}.
}

\subsection{\underline{Pa}rallel \underline{D}ecoding by \underline{I}terative \underline{R}efinement (PaDIR)}
\label{sec:method}

We consider an encoder-decoder architecture, 
in which the decoder is modified to be \emph{non-autoregressive}:
instead of iteratively predicting the next token,
each of our decoder forward pass predicts an update to \emph{all} target positions \emph{in parallel}.
The encoder extracts features from the source sequence,
and based on these features, each decoder forward pass refines its current hypothesis of the target sequence.
The initial decoder hypothesis is a purely random sequence,
and more decoder forward passes correspond to more steps of refinement.
Note that we are \emph{not} the first in the literature to propose this language modeling paradigm.
~\footnote{
Representative prior works: 
\citet{ghazvininejad2020semiautoregressive, savinov2022stepunrolled}, inter alia. 
See \Cref{sec:related_works}.
}
Our focus in this paper is to provide theoretical and empirical analyses to characterize its potentials, limitations and document useful training practices.
Details of inference and training frameworks are in \Cref{sec:appendix:experiments:setup}.

\subsection{Evaluation}
\label{sec:experiments:eval}

We train models on machine translation datasets, 
provide practical recommendations based on our empirical observations, 
and discuss their connections to our theory.
Details of training recipe are provided in \Cref{sec:appendix:experiments:training}.


\paragraph{Benchmarking}
PaDIR models and AR models reach similar BLEU \citep{papineni2002bleu} and BLEURT \citep{sellam2020bleurt, pu2021metrics} scores.
Quantitative experimental results and common baselines are shown in \Cref{table:experiments}, \Cref{table:baseline_bleu}, and \Cref{table:experiments_bleurt}
in Appendix \ref{sec:appendix:experiments:results}.
We discuss several considerations for evaluation metrics in \Cref{sec:appendix:experiments:metrics}.

While bridging the gap between autoregressive and non-autoregressive model has so far focused on achieving
parity in terms of BLEU scores, we believe this is insufficient. 
Since BLEU relies on n-gram overlaps between groundtruths and model predictions,
it does not capture readability very well. Yet readability is paramount for most
practical applications, and it is indisputably something that current autoregressive LMs
excel at.
To provide additional perspectives, we introduce a 
word-level stutter metric, computing how often consecutive words are repeated in the model output but not in the reference.
For all datasets, we found that word-level stutter is 2 or more times more frequent for non-autoregressive models. 

\paragraph{Speed}


The average target length in all datasets ranges between 28 and 33 tokens, including the \texttt{EOS} token. As such a non-autogressive model using 4 decoding steps does 7 to 8 times fewer decoder passes. In practice we see an end-to-end speedup greater than $>$2x for the median and $>$5x for the 99th percentile latency on 
TPU v3
(with 4 decoding steps and batch size 1). The gap between expected and observed speedup is due to fixed costs (input tokenization, encoding, etc.) as well as a better optimization of AR decoding (e.g. through caching of intermediate results). For longer sequences, the constant number of decoding passes in GMLM is advantageous.
For completeness, it is worth noting that the number of decoder passes necessary to achieve good quality (and thus model speed) is application dependent, with some tasks like non-autoregressive text in-painting remaining slower than their autoregressive counterparts, as shown in \citet{savinov2022stepunrolled}.

\subsection{Connecting to theory: quantifying dependency via attention scores}
\label{sec:experiments:attention}

Our theory suggests that stronger dependency between target positions leads to worse generalization guarantee and sampling efficiency.
However, it is unclear how to measure such dependency
for Transformer-based language models trained on natural language data.
In this section, we empirically investigate: \emph{how to predict what target positions have strong dependency which may be challenging for Transformers?}
We test the following two hypotheses:
(1) Strongly dependent target positions have larger \textbf{decoder self-attention} between each other.
(2) Strongly dependent target positions have similar \textbf{cross-attention} distribution to source tokens.

For a pair of target positions,
to measure how well their dependency is modeled in the generated output, 
we focus on adjacent repetitive tokens, a.k.a. \emph{stutter}.
Stuttering is a common error mode among parallel decoding models, 
and we use it as one reasonable proxy for measuring failures in modeling target-side dependency.
~\footnote{
There are other error modes connected to the challenge of modeling target-side dependency,
but they are more ambiguous for measuring and exactly locating.
We do not aim to develop decoding algorithms tailored to just reducing stuttering rate.
(After all, stuttering can be easily removed by rule-based postprocessing.)
Instead, the above are general-purpose hypotheses which are potentially also predictive of other (more complex) failure modes related to target-side dependency.
}
We show:
\begin{itemize}
\item Hypothesis 1 is unlikely to hold:
even on average,
stuttering positions do not have larger \textbf{decoder self-attention} between each other, compared with non-stuttering adjacent positions.~\footnote{
Since all stuttering positions are by definition adjacent,
we think a fair comparison should only consider adjacent positions for non-stuttering position pairs.
}
\item By contrary, Hypothesis 2 is potentially promising:
with various of distribution distance measures,
stuttering positions in the generated output have more similar \textbf{cross-attention} distributions to source tokens, compared with non-stuttering adjacent positions.
\end{itemize}
Details are in \Cref{table:self_attn_main} and \Cref{table:cross_attn_main} 
in \Cref{sec:appendix:experiments:attention}.

\section{Related works}
\label{sec:related_works}

Our theory is inspired by recent progress in sampling:
the connections between pseudolikelihood and approximate tensorization of entropy are discussed in 
\citet{marton2013inequality, marton2015logarithmic, caputo2015approximate, caputo2021block, koehler2023statistical}. 
Benefits of $k$-Gibbs sampler are discussed in \citet{lee2023parallelising}.
Our experiments follow the framework that trains generative masked language models 
and generates samples using parallel decoding by iterative refinement:
\citep{lee2018deterministic, ghazvininejad2019mask, ghazvininejad2020semiautoregressive, kasai2020nonautoregressive, savinov2022stepunrolled},
which tend to be at least twice faster than autoregressive approaches with a small drop in quality
for tasks like machine translation.
The inference process, which converts complete noise to full samples, might resemble diffusion models
\citep{hoogeboom2021argmax, austin2021structured, li2022diffusionlm, gong2023diffuseq, zheng2023reparameterized, lou2023discrete},
but a key conceptual difference is that
diffusion models are trained to revert a small amount of noise at each step,
whereas the family of models that we study in this work are more similar to \emph{masked autoencoders}:
the training objective encourages reconstructing the whole target sequence in each step of decoding.
We discuss additional related works in Appendix~\ref{sec:appendix:related_works}.

\section{Conclusion}

We introduce a new theoretical framework for understanding the power and limitations of generative masked language models (GMLM).
In particular, our theory offers some guidance on the design spaces of learning and inference algorithms,
through the perspectives of 
asymptotic sample complexity for parameter learning,
finite-sample generalization bound for distribution learning,
and the efficiency of Gibbs-like sampling algorithms.
Empirically we adapt T5 to
parallel decoding by iterative refinement (an non-autoregressive GMLM-based language generation strategy which showed strong speed-quality trade-off in the literature for tasks like machine translation).
We recommend some rules of thumb for key design choices,
and discuss the connection between the the empirical findings and our theory.
For future works, we hope the theoretical framework and empirical observations can inspire new training objectives, inference algorithms, and neural network architectures better-suited for parallel decoding.

\subsubsection*{ACKNOWLEDGEMENTS}
We thank Bingbin Liu, Tanya Marwah, Ashwini Pokle, Runtian Zhai, Shanda Li, Chu-Cheng Lin, Nikunj Saunshi, Vaishnavh Nagarajan, Zonglin Li, Chong You, Srinadh Bhojanapalli, Daliang Li, Ziwei Ji, Seungyeon Kim for constructive feedback and insightful discussions.

\newpage
\bibliography{references}
\bibliographystyle{ref_style}

\newpage
\appendix

\thispagestyle{empty}

\def\toptitlebar{
\hrule height4pt
\vskip .25in}

\def\bottomtitlebar{
\vskip .25in
\hrule height1pt
\vskip .25in}

\newcommand{\makesupplementtitle}{\hsize\textwidth
    \linewidth\hsize \toptitlebar {\centering
        {\Large\bfseries Supplementary Material \par}}
    \bottomtitlebar}

\makesupplementtitle

\renewcommand{\theequation}{\thesection.\arabic{equation}}

\tableofcontents

\newpage



\section{Proof of \Cref{l:informationmx}: Generalized information matrix equality}
\label{sec:appendix:proof:generalized_information_matrix}



For convenience, we restate the generalized information matrix equality we are going to show: 

\lemGeneralizedInformationMatrixEquality*

\begin{proof}

All the expectations in the proof will be taken with respect to $(X,K) \sim p_{\mathcal{X}} \times p_{\mathcal{K}}$. To decrease the notational load, we will not explicitly write $p_{\mathcal{X}} \times p_{\mathcal{K}}$. The proof proceeds by first exchanging the order of expectations and derivatives, and using that to show the appropriate terms in the expression for $\nabla_{\theta}^2 L_{PL}(\theta^*)$ vanish. 
    
    \textbf{Step 1: Changing the order of expectations and derivatives}

    We will show that the following two equalities hold:  
    \begin{align}
    \nabla_\theta \E_{(X,K)} \log p_\theta (x_K | x_{\compl{K}})
    &= \E_{(X,K)} \nabla_\theta \log p_\theta (x_K | x_{\compl{K}}) \\ 
    \nabla_\theta^2 \E_{(X,K)} \log p_\theta (x_K | x_{\compl{K}})
    &= \E_{(X,K)} \nabla_\theta^2 \log p_\theta (x_K | x_{\compl{K}})
    \end{align}
    
    Since $\Omega$, $[N]$, and $K \subset [N]$ are both discrete finite,  
    the conditions for the Dominated Convergence Theorem holds under \Cref{as:bounded}: namely, 
    there exists a function $f: \Theta \times \Omega \times \mathcal{K} \mapsto \R$
    such that $\forall \theta \in \Theta$,
    $\Ep{(X,K)}{f(\theta, X, K)} < \infty$,
    $\| \nabla_\theta \log p_\theta (x_K | x_{\compl{K}}) \|_2 \le f(\theta, X, K)$,
    and
    $\| \nabla_\theta^2 \log p_\theta (x_K | x_{\compl{K}}) \|_F \le f(\theta, X, K)$.
    
    Denoting by $\mathbf{e}_i$ the $i$-th standard basis vector, we have:
    \begin{align}
        \frac{\partial}{\partial \theta_j} \Ep{(X,K)}{\log p_\theta (x_K | x_{\compl{K}})}
        &= \lim_{h \rightarrow 0} \frac{1}{h} \left( \Ep{(X,K)}{\log p_{\theta + \ve_j h} (x_K | x_{\compl{K}})} - \Ep{(X,K)}{\log p_\theta (x_K | x_{\compl{K}})} \right) \label{eq:changegrad}\\
        &= \lim_{h \rightarrow 0} \Ep{(X,K)}{ \frac{\log p_{\theta + \ve_j h} (x_K | x_{\compl{K}}) - \log p_\theta (x_K | x_{\compl{K}})}{h} }  \label{eq:changehess}
    \end{align}
    
    By the Mean Value Theorem,
    there exists $\xi(h) \in (0, h)$ such that
    \[
    \frac{\log p_{\theta + \ve_j h} (x_K | x_{\compl{K}}) - \log p_\theta (x_K | x_{\compl{K}}) }{h}
    = \frac{\partial}{\partial_{\theta_j}} \log p_{\theta + \ve_j \xi(h)} (x_K | x_{\compl{K}})
    \]

    So 
    \begin{align*}
        &\quad \frac{\partial}{\partial \theta_j} \Ep{(X,K)}{\log p_\theta (x_K | x_{\compl{K}})} \\
        &= \lim_{h \rightarrow 0} \left( \Ep{(X,K)}{\frac{\partial}{\partial_{\theta_j}} \log p_{\theta + \ve_j \xi(h)} (x_K | x_{\compl{K}})} \right) \\
        &= \Ep{(X,K)}{ \lim_{h \rightarrow 0} \left( \frac{\partial}{\partial_{\theta_j}} \log p_{\theta + \ve_j \xi(h)} (x_K | x_{\compl{K}}) \right) } \quad \text{(Dominated Convergence Thm and \Cref{as:bounded})} \\
        &= \Ep{(X,K)}{ \frac{\partial}{\partial_{\theta_j}} \log p_{\theta} (x_K | x_{\compl{K}}) }
    \end{align*}

    This implies that 
    \[
    \nabla_\theta \E_{(X,K)} \log p_\theta (x_K | x_{\compl{K}})
    = \E_{(X,K)} \nabla_\theta \log p_\theta (x_K | x_{\compl{K}})
    \]
    which proves \eqref{eq:changegrad}. The proof of \eqref{eq:changehess} follows analogously.

    \textbf{Step 2: Rewrite $\nabla_{\theta}^2  L_{PL}(\theta^*)$}
    \begin{align}
        \label{eq:hessian_pl_id}
        \nabla_\theta^2 L_{PL}(\theta)
        & =  - \nabla_\theta^2 \E_{(X,K)} \log p_\theta (x_K | x_{\compl{K}})_{| \theta=\theta^*} \nonumber \\
        &\stackrel{\mathclap{\circled{1}}}{=}  - \E_{(X,K)} \nabla_\theta^2 \log p_\theta (x_K | x_{\compl{K}})_{| \theta=\theta^*} \nonumber \\
        &\stackrel{\mathclap{\circled{2}}}{=} \E_{(X,K)} \nabla_\theta \log p_\theta (x_K | x_{\compl{K}}) \nabla_\theta \log p_\theta (x_K | x_{\compl{K}})^\top_{| \theta=\theta^*} - \frac{\nabla_\theta^2 p_\theta (x_K | x_{\compl{K}})}{p_\theta (x_K | x_{\compl{K}})}_{| \theta=\theta^*} \nonumber \\
        &\stackrel{\mathclap{\circled{3}}}{=} \E_{(X,K)} \nabla_\theta \log p_\theta (x_K | x_{\compl{K}}) \nabla_\theta \log p_\theta (x_K | x_{\compl{K}})^\top_{| \theta=\theta^*}
    \end{align}
    where $\circled{1}$ follows by exchanging the order of expectation and Hessian ($S \in \mathcal{S}_k$ and $x \in \Omega$ are finite),
    and this is valid by \textbf{Step 1} above
    , $\circled{2}$ by an application of chain rule. The last equality $\circled{3}$ follows by a similar calculation as the proof of the classical information matrix equality: 
    \begin{align*}
        &\quad \E_{(X,K)} \frac{\nabla_\theta^2 p_\theta (x_K | x_{\compl{K}})}{p_\theta (x_K | x_{\compl{K}})}_{|\theta=\theta^*} \\
        & = \E_K \E_{x_{\compl{K}}} \E_{x_K | x_{\compl{K}}} \frac{\nabla_\theta^2 p_\theta (x_K | x_{\compl{K}})}{p_\theta (x_K | x_{\compl{K}})}_{| \theta=\theta^*} \\
        & = \E_K \E_{x_{\compl{K}}} \int \frac{\nabla_\theta^2 p_{\theta^*} (x_K | x_{\compl{K}})}{p_{\theta^*} (x_K | x_{\compl{K}})} \cdot p_{\mathcal{X}} (x_K | x_{\compl{K}}) d x_K  \\
        & = \E_K \E_{x_{\compl{K}}} \int \nabla^2_\theta p_{\theta^*} (x_K | x_{\compl{K}}) d x_K \quad \text{, since $p_{\theta^*} = p_{\mathcal{X}}$ by \Cref{as:realizability})}  \\
        & = \E_K \E_{x_{\compl{K}}} \nabla^2_\theta \int p_{\theta^*} (x_K | x_{\compl{K}}) d x_K \quad \text{, by exchanging the order of expectation and Hessian} \\
        & = 0
    \end{align*}
    where the last equality follows since $\int p_{\theta^*} (x_K | x_{\compl{K}}) d x_K = 1$ (so doesn't depend on $\theta$). Similarly, we have:
    \begin{align*}
        &\quad \E_{(X,K)} \nabla_\theta \log p_\theta (x_K | x_{\compl{K}})_{| \theta=\theta^*} \\
        & = \E_K \E_{x_{\compl{K}}} \E_{x_K | x_{\compl{K}}} \frac{\nabla_\theta p_\theta (x_K | x_{\compl{K}})}{p_\theta (x_K | x_{\compl{K}})}_{| \theta=\theta^*} \\
        & = \E_K \E_{x_{\compl{K}}} \int \nabla_\theta p_\theta (x_K | x_{\compl{K}}) d x_K |_{\theta=\theta^*} \\
        & = \E_K \E_{x_{\compl{K}}} \nabla_\theta \int p_\theta (x_K | x_{\compl{K}}) d x_K |_{ \theta=\theta^*} \\
        & = 0
    \end{align*}
        where the last equality follows since $\int p_\theta (x_K | x_{\compl{K}}) d x_K = 1$ (so doesn't depend on $\theta$). Plugging this into the definition of covariance, we have: 
    \begin{align}
    \label{eq:cov_pl_id}
        &\Cov(\nabla_\theta -  \log p_\theta (X_K | X_{\compl{K}}) )_{| \theta=\theta^*} \nonumber \\
        & = \E_{(X,K)} \nabla_\theta \log p_\theta (x_K | x_{\compl{K}}) \nabla_\theta \log p_\theta (x_K | x_{\compl{K}})^\top \nonumber \\ 
        &\quad - \E_{(X,K)} \nabla_\theta \log p_\theta (x_K | x_{\compl{K}}) \cdot \E_{(X,K)} \nabla_\theta \log p_\theta (x_K | x_{\compl{K}} )^\top_{| \theta=\theta^*} \nonumber \\
        & = \E_{(X,K)} \nabla_\theta \log p_\theta (x_K | x_{\compl{K}}) \nabla_\theta \log p_\theta (x_K | x_{\compl{K}})^\top_{| \theta=\theta^*}
    \end{align}
    The proof of the lemma thus follows
    because the RHS of \Cref{eq:cov_pl_id} matches that of \Cref{eq:hessian_pl_id}. 

\end{proof}

\clearpage
\section{Proof of \Cref{thm:k_monotone}: Masking more is (statistically) better}
\label{sec:appendix:proof:k_monotone}
In this Section, we provide the proof for \Cref{thm:k_monotone}. 

\begin{proof}[Proof of \Cref{thm:k_monotone}]
All the expectations in the proof will be taken with respect to $(X,K) \sim p_{\mathcal{X}} \times p_{\mathcal{K}}$. To decrease the notational load, we will not explicitly write $p_{\mathcal{X}} \times p_{\mathcal{K}}$. By Lemma~\ref{l:informationmx}, we have:

\begin{equation}
\label{eq:hessian_pl_k}
\hess_\theta L_{PL}^{k} (\theta^*) = \E_{(X,K)} \nabla_\theta \log p_\theta (x_K | x_{\compl{K}} ) \nabla_\theta \log p_\theta (x_K | x_{\compl{K}} )^\top_  {| \theta=\theta^*}
\end{equation}


Let $\mathcal{S}_k$ denote the set 
$
\{ K \subset [N] \, | \, \abs{K} = k \}
$. For every $T \in \mathcal{S}_{k+1}$ and $a \in T$ we have:
\begin{align*}
\log p(x_T | x_{\compl{T}} ) &= \log p(x_S, x_a | x_{-\{S \cup a\}} ) \text{ where } S \coloneqq T \backslash \{a\} \\
&= \log \left( p(x_a | x_{-\{S \cup a\}} ) \cdot p(x_S | x_{-\{S \cup a\}}, x_a) \right)  \\
&= \log p(x_a | x_{-\{S \cup a\}}) + \log p(x_S | x_{\compl{S}} )
\end{align*}

Using this identity, we can write: 
\begin{align}
\hess_\theta L_{PL}^{k+1} (\theta^*)  
& = \E_{T \sim S_{k+1}} \E_{x_T, x_{\compl{T}} } \nabla_\theta \log p_\theta (x_T | x_{\compl{T}} ) \nabla_\theta \log p_\theta (x_T | x_{\compl{T}} )^\top_{| \theta=\theta^*} \nonumber \\
& = \E_{S \sim S_k} \E_{a \not \in S} \E_{x_S, x_a, x_{-\{S \cup a\}}  } \nabla_\theta \log p_\theta (x_T | x_{\compl{T}} ) \nabla_\theta \log p_\theta (x_T | x_{\compl{T}} )^\top_{|\theta=\theta^*} \nonumber\\
& = \E_{S \sim S_k} \E_{a \not \in S} \E_{x_S, x_a, x_{-\{S \cup a\}} } 
\left(
\nabla_\theta \log p_\theta(x_a | x_{-\{S \cup a\}} ) + \nabla_\theta \log p_\theta(x_S | x_{\compl{S}} )
\right) \nonumber \\
&\quad \cdot
\left(
\nabla_\theta \log p_\theta(x_a | x_{-\{S \cup a\}} ) + \nabla \log p_\theta(x_S | x_{\compl{S}} )
\right)
^\top_{| \theta=\theta^*}
\label{eq:mainsum}
\end{align}

Let us denote:
\begin{align*}
A &\coloneqq \E_{S \sim S_k} \E_{a \not \in S} \E_{x } \nabla_\theta \log p_\theta(x_S | x_{\compl{S}} ) \cdot \nabla_\theta \log p_\theta(x_S | x_{\compl{S}} )^\top_{| \theta=\theta^*} \\
B &\coloneqq \E_{S \sim S_k} \E_{a \not \in S} \E_{x } \nabla_\theta \log p_\theta(x_a | x_{-\{S \cup a\}} ) \cdot \nabla_\theta \log p_\theta(x_S | x_{\compl{S}} )^\top _{| \theta=\theta^*} \\
C &\coloneqq \E_{S \sim S_k} \E_{a \not \in S} \E_{x } \nabla_\theta \log p_\theta(x_a | x_{-\{S \cup a\}} ) \cdot \nabla_\theta \log p_\theta(x_a | x_{-\{S \cup a\}} )^\top _{| \theta=\theta^*}
\end{align*}

By expanding the previous expression, we have 
\begin{equation}
\label{eq:hessian_pl_k+1}
    \hess_\theta L_{PL}^{k+1} (\theta^*) = A + B + B^\top + C
\end{equation}

Consider $A$ first. Note that for a fixed $S \in S_k$, 
$\E_{x } \nabla_\theta \log p_\theta(x_S | x_{\compl{S}} ) \cdot \nabla_\theta \log p_\theta(x_S | x_{\compl{S}} )^\top$ 
is independent of $a \not \in S$ and therefore: 
\begin{align*}
A & = \E_{S \sim S_k} \E_{x } \nabla_\theta \log p_\theta(x_S | x_{\compl{S}} ) \cdot \nabla_\theta \log p_\theta(x_S | x_{\compl{S}} )^\top_{| \theta=\theta^*} \\
& = \hess_\theta L_{PL}^k (\theta^*) \quad \text{(by \Cref{eq:hessian_pl_k})}
\end{align*}

Proceeding to $B$, for a given $S \in S_k, x_{\compl{S}}$, we have 
\begin{align}
\label{eq:expected_grad_log_pl}
&\quad \Ep{x_S| x_{\compl S}}{\nabla_\theta \log p_\theta(x_S | x_{\compl{S}} )^\top}_{|\theta=\theta^*} \nonumber \\
&= \int \nabla_\theta \log p_\theta(x_S | x_{\compl{S}} )^\top \cdot p(x_S| x_{\compl S}) d x_S |_{\theta=\theta^*} \nonumber \\
&= \int \frac{\nabla_\theta p_\theta(x_S | x_{\compl{S}} )}{p_{\theta^*}(x_S | x_{\compl{S}} )} ^\top \cdot p(x_S| x_{\compl S}) d x_S  |_{\theta=\theta^*} \nonumber \\
&= \int \nabla_\theta p_\theta(x_S | x_{\compl{S}} ) ^\top d x_S |_{\theta=\theta^*} \nonumber \\
&= \nabla_\theta \int p_\theta(x_S | x_{\compl{S}} ) ^\top d x_S |_{\theta=\theta^*} \quad \text{(valid under \Cref{as:bounded}, see Step 1 in the proof of \Cref{l:informationmx})} \nonumber \\
&= \nabla_\theta 1 = 0
\end{align}

Therefore:
\begin{align*}
B & = \E_{S \sim S_k} \E_{a \not \in S} \E_{x_S, x_a, x_{-\{S \cup a\}}} \nabla_\theta \log p_\theta(x_a | x_{-\{S \cup a\}} ) \cdot \nabla_\theta \log p_\theta(x_S | x_{\compl{S}} )^\top |_{\theta=\theta^*} \\
&= \E_{S \sim S_k} \E_{a \not \in S} \E_{x_a, x_{-\{S \cup a\}}} \nabla_\theta \log p_\theta(x_a | x_{-\{S \cup a\}} ) \cdot \Ep{x_S}{ \nabla_\theta \log p_\theta(x_S | x_{\compl{S}} )^\top} |_{\theta=\theta^*} \\
&\quad \text{(valid under \Cref{as:bounded}, see Step 1 in the proof of \Cref{l:informationmx})} \\
&= \E_{S \sim S_k} \E_{a \not \in S} \E_{x_a, x_{-\{S \cup a\}}} \nabla_\theta \log p_\theta(x_a | x_{-\{S \cup a\}} ) |_{\theta=\theta^*} \cdot 0 \quad \text{(by \Cref{eq:expected_grad_log_pl})}  \\
& = 0
\end{align*}

Finally, each term 
$
\nabla_\theta \log p_\theta(x_a | x_{-\{S \cup a\}} ) \cdot \nabla_\theta \log p_\theta(x_a | x_{-\{S \cup a\}} )^\top 
\succeq 0$
therefore $C \succeq 0$.

Plugging this back in \eqref{eq:hessian_pl_k+1}, we have:  
$$\hess_\theta L_{PL}^{k+1} (\theta^*) 
= \hess_\theta L_{PL}^k (\theta^*) + C \succeq \hess_\theta L_{PL}^k (\theta^*)$$

Consequently, by monotonicity of the matrix inverse, we have 
$$\Gamma_{PL}^{k+1} 
= \left(\hess_\theta L_{PL}^{k+1} (\theta^*)\right)^{-1} 
\preceq \left(\hess_\theta L_{PL}^{k} (\theta^*) \right)^{-1} 
= \Gamma_{PL}^{k}$$ 
as we need. 
\end{proof}
\clearpage
\section{Generalizations for adaptive masking} 

In this section, we provide proofs for several of the claims in Section~\ref{sec:theory:sample_complexity:dependent}. 

\subsection{Conditioning on $K$} 
\label{sec:appendix:proof:adaptive}

First, we clarify a slightly subtle (and counterintuitive) point stressed in \Cref{rem:parameterization}: in general, $p_{\mathcal{X}} (x_{K} | x_{\compl{K}}, K) \ne p_{\mathcal{X}} (x_K | x_{\compl{K}})$.  

\begin{lemma} \label{lem:XS_implies_S}
    Consider $\mathcal{X} = \{(0,0), (0,1), (1,0), (1,1)\} $. There exists a distribution $p_{\mathcal{X}, \mathcal{K}}$ such that
    $p_{\mathcal{X}} (x_{K} | x_{\compl{K}}, K) \ne p_{\mathcal{X}} (x_K | x_{\compl{K}})$
    for some
    $x \in \mathcal{X}, \, K \in \mathcal{K}$.
\end{lemma}

\begin{proof}
To define $p_{\mathcal{X}, \mathcal{K}}$, it suffices to define $p_{\mathcal{X}}$ and $p_{\mathcal{K}}(\cdot | x), \forall x \in \mathcal{X}$. 
\[
p_{\mathcal{X}}(X) = \begin{cases}
        (0, 0), \quad &\text{with probability } \frac{1}{2} \\ 
        (0, 1), \quad &\text{with probability } \frac{1}{3} \\ 
        (1, 0), \quad &\text{with probability } \frac{1}{6} \\ 
        (0,0), \quad &\text{with probability } 0
        \end{cases}
\]
and let 
\begin{align*}
    p_{\mathcal{K}}(K \mid X = (0, 0)) &= \begin{cases}
        \{0\}, \quad &\text{with probability } \frac{1}{2} \\ 
        \{1\}, \quad &\text{with probability } \frac{1}{2} \\ 
    \end{cases} \\
    p_{\mathcal{K}}(K \mid X = (0, 1)) &= \begin{cases}
        \{0\}, \quad &\text{with probability } \frac{1}{3} \\ 
        \{1\}, \quad &\text{with probability } \frac{2}{3} \\ 
    \end{cases} \\
    p_{\mathcal{K}}(K \mid X = (1, 0)) &= \begin{cases}
        \{0\}, \quad &\text{with probability } \frac{1}{4} \\ 
        \{1\}, \quad &\text{with probability } \frac{3}{4} \\ 
    \end{cases} 
\end{align*}
By multiplying $p_{\mathcal{X}}(X)$ and $p_{\mathcal{K}}(K \mid X)$, we have
\begin{align*}
    p((0, 0), \{0\}) = \frac{1}{4}, \quad &p((0, 0), \{1\}) = \frac{1}{4} \\
    p((0, 1), \{0\}) = \frac{1}{9}, \quad &p((0, 1), \{1\}) = \frac{2}{9} \\
    p((1, 0), \{0\}) = \frac{1}{24},\quad  &p((1, 0), \{1\}) = \frac{1}{8} 
\end{align*}
Finally, we will see that
$p_{\mathcal{X}} (x_1=0 | x_0=0, \{0\}) \ne p_{\mathcal{X}} (x_1=0 | x_0=0)$:
\begin{align*}
    p_{\mathcal{X}} (x_1=0 | x_0=0, \{0\}) &= \frac{p((0, 0), \{0\})}{p((0, 0), \{0\}) + p((0, 1), \{0\})} = \frac{9}{13} \\
    p_{\mathcal{X}} (x_1=0 | x_0=0) &= \frac{p_{\mathcal{X}}((0,0))}{p_{\mathcal{X}}((0,0)) + p_{\mathcal{X}}((0,1))} = \frac{3}{5}
\end{align*}

\end{proof}

Instead, by correctly marginalizing, the following equality obtains: 
\begin{lemma} \label{lem:XS_expectation_over_S}
    For any distribution $p_{\mathcal{X}, \mathcal{K}}$, we have: 
\begin{equation}
    \forall x \in \mathcal{X}, \;
    \forall K \in \mathcal{K}, \;
    p_{\mathcal{X}} (x_K | x_{\compl{K}}) = \Ep{K' \sim p_{\mathcal{K}}(\cdot \mid x_{\compl{K}}) }{p_{\mathcal{X}, \mathcal{K}} (x_K | x_{\compl{K}}, K')}
\end{equation}
\end{lemma}

\begin{proof} 
The proof proceeds by a sequence of straightforward rewrites:
\begin{align*}
    p_{\mathcal{X}} (x_K | x_{\compl{K}}) &= \frac{p_{\mathcal{X}} (x_K, x_{\compl{K}})}{p_{\mathcal{X}} (x_{\compl{K}})} \\
    &= \sum_{K'} \frac{p_{\mathcal{X}, \mathcal{K}} (x_K, x_{\compl{K}}, K')}{p_{\mathcal{X}}(x_{\compl{K}})} \\
    &= \sum_{K'} \frac{p_{\mathcal{X}, \mathcal{K}}(K', x_{\compl{K}})}{p_{\mathcal{X}}(x_{\compl{K}})} \cdot \frac{p_{\mathcal{X}, \mathcal{K}} (x_K, x_{\compl{K}}, K')}{p_{\mathcal{X},\mathcal{K}} (x_{\compl{K}}, K')} \\
    &= \sum_{K'} p_{\mathcal{K}}(K' \mid x_{\compl{K}}) p_{\mathcal{X}, \mathcal{K}}(x_K | x_{\compl{K}}, K') \\
    &= \Ep{K' \sim p_{\mathcal{K}}(\cdot \mid x_{\compl{s}}) }{p_{\mathcal{X}, \mathcal{K}} (x_K | x_{\compl{K}}, K')} 
\end{align*}
\end{proof}

\subsection{Information matrix equality for adaptive masking}

We prove a more general version of \Cref{l:informationmx} when $p_\mathcal{K}$ is allowed to depend on $X$, 
which is needed for \Cref{thm:asymptotic_variance}. Recall from Section~\ref{sec:theory:sample_complexity:dependent} that the distribution $p_{\mathcal{X},\mathcal{K}}$ is defined such that: 
\[p_{\mathcal{X}, \mathcal{K}}(X, K) := p_{\mathcal{X}}(X) p_{\mathcal{K}}(K | X) \]

\begin{restatable}[Generalized information matrix equality, adaptive masking]{lemma}{lemGeneralizedInformationMatrixEqualityAdaptive} 
\label{l:informationmx_adaptive}
Under \Cref{as:bounded} and \Cref{as:realizability},
the weighted pseudolikelihood loss (\Cref{d:genmple:dependent}) verifies: 
$$
\nabla_\theta^2 L_{PL}(\theta^*) = \Cov_{(X,K) \sim p_{\mathcal{X}, \mathcal{K}}}(-\nabla_{\theta} \log p_\theta (X_K | X_{\compl{K}}, K))_{\vert \theta = \theta^*}.$$

\end{restatable}
\begin{proof}

All the expectations in the proof will be taken with respect to $(X,K) \sim p_{\mathcal{X}, \mathcal{K}}$. To decrease the notational load, we will not explicitly write $p_{\mathcal{X}, \mathcal{K}}$. Same as Lemma~\ref{l:informationmx}, the proof proceeds by first exchanging the order of expectations and derivatives, and using that to show the appropriate terms in the expression for $\nabla_{\theta}^2 L_{PL}(\theta^*)$ vanish.  

In fact, it's readily seen that the proof of Step 1 in Lemma~\ref{l:informationmx} 
(\Cref{sec:appendix:proof:generalized_information_matrix})
doesn't depend on $p_{\mathcal{X}, \mathcal{K}}$ being a product distribution, and the same proof applies to our setting, namely we have:

    \begin{align}
    \nabla_\theta \E_{(X,K)} \log p_\theta (x_K | x_{\compl{K}}, K)
    &= \E_{(X,K)} \nabla_\theta \log p_\theta (x_K | x_{\compl{K}}, K) \\ 
    \nabla_\theta^2 \E_{(X,K)} \log p_\theta (x_K | x_{\compl{K}}, K)
    &= \E_{(X,K)} \nabla_\theta^2 \log p_\theta (x_K | x_{\compl{K}}, K)
    \end{align}

    We can also rewrite the expression for $\nabla_{\theta}^2  L_{PL}(\theta^*)$ almost the same way we did in Step 2 in Lemma \ref{l:informationmx}: 
    \begin{align}
        \label{eq:hessian_pl}
        \nabla_\theta^2 L_{PL}(\theta)
        & =  - \nabla_\theta^2 \E_{(X,K)} \log p_\theta (x_K | x_{\compl{K}}, K)_{| \theta=\theta^*} \nonumber \\
        &\stackrel{\mathclap{\circled{1}}}{=}  - \E_{(X,K)} \nabla_\theta^2 \log p_\theta (x_K | x_{\compl{K}}, K)_{| \theta=\theta^*} \nonumber \\
        &\stackrel{\mathclap{\circled{2}}}{=} \E_{(X,K)} \nabla_\theta \log p_\theta (x_K | x_{\compl{K}}, K ) \nabla_\theta \log p_\theta (x_K | x_{\compl{K}}, K)^\top_{| \theta=\theta^*} - \frac{\nabla_\theta^2 p_\theta (x_K | x_{\compl{K}}, K)}{p_\theta (x_K | x_{\compl{K}}, K)}_{| \theta=\theta^*} \nonumber \\
        &\stackrel{\mathclap{\circled{3}}}{=} \E_{(X,K)} \nabla_\theta \log p_\theta (x_K | x_{\compl{K}}, K) \nabla_\theta \log p_\theta (x_K | x_{\compl{K}}, K)^\top_{| \theta=\theta^*}
    \end{align}
    where $\circled{1}$ follows by exchanging the order of expectation and Hessian ($S \in \mathcal{S}_k$ and $x \in \Omega$ are finite),
    and this is valid by \textbf{Step 1} above
    , $\circled{2}$ by an application of chain rule. The last equality $\circled{3}$ follows by a similar calculation as the proof of the classical information matrix equality (and again, analogously to the calculation in Lemma \ref{l:informationmx}): 
    \begin{align*}
        &\quad \E_{(X,K)} \frac{\nabla_\theta^2 p_\theta (x_K | x_{\compl{K}}, K)}{p_\theta (x_K | x_{\compl{K}}, K)}_{|\theta=\theta^*} \\
        & = \E_K \E_{x_{\compl{K}} \mid K } \E_{x_K | x_{\compl{K}}, K} \frac{\nabla_\theta^2 p_\theta (x_K | x_{\compl{K}}, K)}{p_\theta (x_K | x_{\compl{K}}, K)}_{| \theta=\theta^*} \\
        & = \E_K \E_{x_{\compl{K}} \mid K} \int \frac{\nabla_\theta^2 p_{\theta^*} (x_K | x_{\compl{K}}, K)}{p_{\theta^*} (x_K | x_{\compl{K}}, K)} \cdot p_{\mathcal{X}, \mathcal{K}} (x_K | x_{\compl{K}}, K) d x_K  \\
        & = \E_K \E_{x_{\compl{K}} \mid K} \int \nabla^2_\theta p_{\theta^*} (x_K | x_{\compl{K}}, K) d x_K \quad \text{, since $p_{\theta^*} = p_{\mathcal{X}}$ by \Cref{as:realizability} and Definition \eqref{eq:defnconditional}}  \\
        & = \E_K \E_{x_{\compl{K}} \mid K} \nabla^2_\theta \int p_{\theta^*} (x_K | x_{\compl{K}}, K) d x_K \quad \text{, by exchanging the order of expectation and Hessian} \\
        & = 0
    \end{align*}
    where the last equality follows since $\int p_{\theta^*} (x_K | x_{\compl{K}}, K) d x_K = 1$ (so doesn't depend on $\theta$). Similarly, we have:
    \begin{align*}
        &\quad \E_{(X,K)} \nabla_\theta \log p_\theta (x_K | x_{\compl{K}}, K)_{| \theta=\theta^*} \\
        & = \E_K \E_{x_{\compl{K}} \mid K} \E_{x_K | x_{\compl{K}}, K} \frac{\nabla_\theta p_\theta (x_K | x_{\compl{K}}, K)}{p_\theta (x_K | x_{\compl{K}}, K)}_{| \theta=\theta^*} \\
        & = \E_K \E_{x_{\compl{K}} \mid K} \int \nabla_\theta p_\theta (x_K | x_{\compl{K}}, K) d x_K |_{\theta=\theta^*} \\
        & = \E_K \E_{x_{\compl{K}} \mid K} \nabla_\theta \int p_\theta (x_K | x_{\compl{K}}, K) d x_K |_{ \theta=\theta^*} \\
        & = 0
    \end{align*}
        where the last equality follows since $\int p_\theta (x_K | x_{\compl{K}} , K) d x_K = 1$ (so doesn't depend on $\theta$). Plugging this into the definition of covariance, we have: 
    \begin{align}
    \label{eq:cov_pl}
        &\quad \Cov(\nabla_\theta l_{PL}(\theta^*))  \nonumber \\
        &= \Cov(\nabla_\theta -  \log p_\theta (X_K | X_{\compl{K}},K ) )_{| \theta=\theta^*} \nonumber \\
        & = \E_{(X,K)} \nabla_\theta \log p_\theta (x_K | x_{\compl{K}}, K) \nabla_\theta \log p_\theta (x_K | x_{\compl{K}}, K)^\top_{| \theta=\theta^*}
    \end{align}
   which finishes the proof of the Lemma. 
\end{proof}

\subsection{Proof of \Cref{prop:block_dynamics_dirichlet}: Dirichlet form for adaptive block dynamics}
\label{sec:appendix:proof:block_dynamics_dirichlet}

\propBlockDynamicsDirichlet*

\begin{proof}
Let us denote by $\Xi:= \mathcal{X} \times \mathcal{K}$, and note that $\Xi$ is the domain for both $f$ and $g$. According definition of block dynamics in \Cref{d:blockdynamics},
for each pair of states $(X, K_1), (Y, K_2) \in \Xi$,
the transition matrix $P$ is:
\begin{equation} \label{eq:transition_matrix_block_dynamics}
P\left((X, K_1), (Y, K_2)\right) = \indp{K_1 = K_2} \indp{X_{\compl{K_1}} = Y_{\compl{K_1}}} p(Y_{K_1} \mid X_{\compl{K_1}}, K_1)
\end{equation}
The rest of the proof is straightforward calculation, expanding the expression in the definition of the Dirichlet form (\Cref{def:dirichlet}). Namely, we have: 

\begin{align}
\label{eq:block_dynamics_dirichlet_expanded}
&\mathcal{E}_P (f, g) \nonumber\\
&= \frac{1}{2} \sum_{(X, K_1), (Y, K_2) \in \Xi} p(X, K_1) P\left((X, K_1), (Y, K_2)\right) (f(X, K_1) - f(Y, K_2))(g(X, K_1) - g(Y, K_2)) \nonumber \\
&= \frac{1}{2} \sum_{(X, K_1), (Y, K_2) \in \Xi} p(X, K_1) \cdot (f(X, K_1) - f(Y, K_2))(g(X, K_1) - g(Y, K_2)) \nonumber \\
&\quad \cdot \indp{K_1 = K_2} \indp{X_{\compl{K_1}} = Y_{\compl{K_1}}} p(Y_{K_1} \mid X_{\compl{K_1}}, X_1) \nonumber \\
&= \frac{1}{2} \sum_{X \in \mathcal{X}} p_{\mathcal{X}}(X) \sum_{K \in \mathcal{K}} p_{\mathcal{K}}(K \mid X) \sum_{Y \in \mathcal{X}, Y_{\compl{K}} = X_{\compl{K}} } p(Y_K \mid X_{\compl{K}}, K) \cdot (f(X, K) - f(Y, K))(g(X, K) - g(Y, K)) \nonumber  \\
&= \frac{1}{2} \mathbb{E}_{X} \mathbb{E}_{K \mid X} \mathbb{E}_{Y_K \mid X_{\compl{K}}, K}(f(X, K) - f(Y, K))(g(X, K) - g(Y, K)) \nonumber   \\
&=  \frac{1}{2} \cdot \big( 2 \cdot \mathbb{E}_{K} \mathbb{E}_{X_{\compl{K}} \mid K} \mathbb{E}_{X_K \mid X_{\compl{K}}, K}f(X, K) g(X, K) - 2 \cdot \mathbb{E}_{K}\mathbb{E}_{X_{\compl{K}} \mid K}\Ep{X_K \mid X_{\compl{K}}, K}{f(X, K)} \cdot \Ep{X_K \mid X_{\compl{K}}, K}{g(X, K)} \big) \nonumber \\
&\quad \text{(we can merge terms because the roles of $X$ and $Y$ are symmetric)} \nonumber \\
&= \E_{K \sim p} \big[ \E_{x_{\compl{K}} \sim p(x_{\compl{K}} \mid K)} \big[ \Ep{y_K \sim p(x_K \mid x_{\compl{K}}, K)}{f(y, K) g(y, K)}  - \Ep{y_K \sim p(x_K \mid x_{\compl{K}}, K)}{f(y, K)}  \nonumber  \\
&\quad \cdot \Ep{y_K \sim p(x_K \mid x_{\compl{K}}, K)}{ g(y, K) } \big] \big]  \nonumber \\
&= \mathbb{E}_{(X_{\compl{K}}, K) \sim p_{\mathcal{X}, \mathcal{K}}} \left[ \text{Cov}_{X_K |  (X_{\compl{K}}, K)}(f, g) \right]
\end{align}
which completes the proof.
\end{proof}

\subsection{Proof of \Cref{thm:asymptotic_variance}: Asymptotic sample complexity for adaptively-weighted MPLE}
\label{sec:appendix:proof:asymptotic_variance}

Note that \Cref{thm:asymptotic_variance} generalizes \Cref{thm:asymptotic_variance_alpha}.
To reduce proof duplication, we only write the proof for the more general \Cref{thm:asymptotic_variance} here.
Notational definition for $p_\theta (x_K | x_{\compl{K}}, K )$ and other background info are in \Cref{sec:theory:sample_complexity:dependent}.

\thmAsymptoticVarianceAdaptive*

\begin{proof}

By \Cref{l:informationmx_adaptive} and Lemma~\ref{lem:asymptotics}, for $n$ training samples as $n \to \infty$, we have: 

\begin{equation}
\label{eq:asymptotic_normal}
    \sqrt{n}(\hat{\theta}_{PL} - \theta^*)
    \rightarrow \mathcal{N}\left(0, (\Cov_{(X,K) \sim p_{\mathcal{X}, \mathcal{K}}}(-\nabla_{\theta} \log p_\theta (X_K | X_{\compl{K}}, K))_{\vert \theta = \theta^*})^{-1}\right)
\end{equation}

Now we relate the RHS of \eqref{eq:asymptotic_normal} to $\fisher$. Let $d_\Theta$ denote the dimensionality of $\theta$, i.e. $\theta \in \R^{d_\Theta}$. Then, for any test vector $v \in \R^{d_\Theta}$ we have: 

\begin{align}
    &v^\top \E_{(X,K)} \nabla_\theta \log p_\theta (x_K | x_{\compl{S}}, K) \nabla_\theta \log p_\theta (x_K | x_{\compl{S}}, K)^\top v_{|\theta=\theta^*} \nonumber \\
    & = \E_{(X,K)} (\nabla_\theta \log p_\theta (x_K | x_{\compl{K}}, K)^\top v)^2_{|\theta=\theta^*} \nonumber \\
    & = \E_K \E_{x_{\compl{K}} \mid K } \Var_{x_K | x_{\compl{K}, K }} (\nabla_\theta \log p_\theta (x_K | x_{\compl{K}}, K)^\top v)_{| \theta=\theta^*} + (\E_{x_K | x_{\compl{K}}, K} \nabla_\theta \log p_\theta (x_K | x_{\compl{K}}, K)^\top v)^2_{|\theta=\theta^*} \label{eq:cov_grad_pl}
\end{align}  

Denote 
$f(X, K) \coloneqq \nabla_\theta \log p_\theta (x_K | x_{\compl{K}}, K)^\top v$
Consider the two parts in \Cref{eq:cov_grad_pl} separately:
the first term is simply 
$\E_K \E_{x_{\compl{K}}\mid K} \Var_{x_S | x_{\compl{K}, K }} (f(X, K))$,
which, by \Cref{prop:block_dynamics_dirichlet},
is equal to
$\mathcal{E}_P (f, f)$.
Moreover, by Poincar\'e inequality (\Cref{def:poincare}),
this is $\ge \frac{1}{C} \Var_{p}(f)$.

The second term simplifies to
\begin{align*}
    &\quad \E_K \E_{x_{\compl{K}} \mid K} \left(\Ep{x_K | x_{\compl{K}}, K}{\nabla_\theta \log p_\theta (x_K | x_{\compl{K}}, K)^\top v}\right)^2 _{|\theta=\theta^*} \\
    &= \E_K \E_{x_{\compl{K}} \mid K} \left(\Ep{x_K | x_{\compl{K}}, K}{\left( \frac{\nabla_\theta p_\theta (x_K | x_{\compl{K}}, K)}{p_\theta (x_K | x_{\compl{K}}, K)} \right)^\top v}\right)^2 _{|\theta=\theta^*} \\
&= \E_K \Ep{x_{\compl{K}} \mid K}{\int \left( \frac{\nabla_\theta p_\theta (x_K | x_{\compl{K}}, K)}{p_\theta (x_K | x_{\compl{K}}, K)} \right)^\top v \cdot p_{\mathcal{X},\mathcal{K}}(x_K | x_{\compl{K}}, K) \, d x_K}^2 _{|\theta=\theta^*} \\
&= \E_K \Ep{x_{\compl{K}} \mid K}{\int \left( \nabla_\theta p_\theta (x_K | x_{\compl{K}}, K) \right)^\top v \, d x_K}^2_{| \theta=\theta^*} \quad \text{, since $p_{\theta^*} = p_{\mathcal{X}}$ by \Cref{as:realizability} and Definition \eqref{eq:defnconditional}} \\
&= \E_K \Ep{x_{\compl{K}} \mid K}{\nabla_\theta \left( \int  p_\theta (x_K | x_{\compl{K}}, K) \, d x_K \right)^\top v }^2_{|\theta=\theta^*} \quad \text{(since $p_{\theta}$ is differentiable wrt $\theta$ by \Cref{as:bounded})} \\
&= 0
\end{align*}


Therefore, we have  $\displaystyle  \Cov_{(X,K) \sim p_{\mathcal{X}, \mathcal{K}}}(-\nabla_{\theta} \log p_\theta (X_K | X_{\compl{K}}, K))_{\vert \theta = \theta^*} \succeq \frac{1}{C} \fisher$. 

Plugging into \Cref{eq:asymptotic_normal}, and using the monotonicity of the matrix inverse \citep{toda2011operator}, we obtain the upper bound on the asymptotic variance of our estimator we want: 
\begin{align*}
    \Gamma_{PL} \preceq C \fisher^{-1}
\end{align*}
\end{proof}

\clearpage
\section{Proof of Theorem~\ref{thm:generalization_bound_joint}: Generalization bound for learning the joint distribution}
\label{sec:appendix:proof:generalization_bound_joint}

We first state our overall structure of the proof of \Cref{thm:generalization_bound_joint},
and then state and prove the key lemmas mentioned therein.

\thmGeneralizationBoundJoint*

\begin{proof}

We first introduce a few pieces of notation. We will denote the data samples as
$\mathcal{S_X} \coloneqq \{ X^{(i)} | X^{(i)} \sim p(X) \}$,
$\abs{\mathcal{S_X}} = n$, 
and for each $X^{(i)}$ we sample $m$ masks $\mathcal{S_K}^{(i)} \coloneqq \{K_1^{(i)}, \ldots ,K_{m}^{(i)} \}$ in which $K_j^{(i)}$ is sampled iid from $\mathcal{K}$ according to probabilities 
$p_\mathcal{K}(\cdot \mid X)$.~\footnote{
Note that the $\{ \cdot \}$ notation does not mean sets: 
duplicate entries are allowed in the training data $\mathcal{S_X}$ and $\mathcal{S_K}^{(i)}$.
}
\Cref{thm:generalization_bound_joint} follows by combining the following steps:

\paragraph{Step 1: relating closeness of the \emph{conditional} distributions (i.e. the loss) to closeness of the \emph{joint} distribution.}
The connection is established through the definition of the block-generalized approximate tensorization of entropy in \Cref{d:block_catconst}, by which we get\footnote{Recall, $\tilde{L}_{PL}$ is defined in \eqref{eq:intermetiatepl}}: 
\[ \KL \left( \tilde{p}_{\mathcal{X}}, p_{\hat\theta} \right) 
\le \bar{C}_{AT}(p_{\hat\theta}) \tilde{L}_{PL}(\hat\theta) \]
The details are in \Cref{prop:conditional_to_joint}.
By Pinsker’s inequality, this implies
\begin{equation} \label{eq:conditional_to_joint}
\TVp{\tilde{p}_{\mathcal{X}}}{p_{\hat\theta}} \le \sqrt{\frac{1}{2} \KLp{\tilde{p}_{\mathcal{X}}}{p_{\hat\theta}}}
\le \sqrt{\frac{1}{2} \bar{C}_{AT}(p_{\hat\theta}) \tilde{L}_{PL}(\hat\theta)}    
\end{equation}

\paragraph{Step 2: generalization bound for learning the \emph{conditional} distributions.}
We show that \Cref{as:support_margin} and \Cref{as:complexity} 
imply a generalization guarantee for learning the \emph{conditional} distributions from a finite sample of sequences and masked positions. We show that with probability at least $1 - \frac{\delta}{2}$,
we have
\begin{equation} \label{eq:generalization_bound_conditional}
\abs{L_{PL}(\hat\theta) - \tilde{L}_{PL}(\hat\theta)}
< \left( \sqrt{\frac{2^{3N} \abs{\Omega}^N C_\epsilon(\Theta)}{m \cdot \delta}} +  \sqrt{\frac{\ln\frac{8 C_\epsilon(\Theta)}{\delta}}{2 n}} \right) \cdot \ln \frac{1}{\margin} + \epsilon
\end{equation}
Proof details of this step are in \Cref{lem:generalization_bound_conditional}.

\paragraph{Step 3: \emph{empirical} joint distribution converges to \emph{population} joint distribution.}
With probability at least $1 - \frac{\delta}{2}$,
we have: 
\begin{equation} \label{eq:empirical_pmf}
\TVp{\tilde{p}_{\mathcal{X}}}{p_{\mathcal{X}}}
< \sqrt{\frac{\abs{\Omega}^{3N}}{8 \delta n}}
\end{equation}
The proof of this is standard and details are in \Cref{lem:empirical_pmf}.

\paragraph{Step 4: union bound and triangle inequality}
By union bound, with probability at least $1 - \delta$, both \Cref{eq:generalization_bound_conditional} and \Cref{eq:empirical_pmf} hold.
Therefore, putting together the previous steps, we get: 
\begin{align*}
    &\quad \TVp{p_{\hat\theta}}{p_{\mathcal{X}}} \le \TVp{\tilde{p}_{\mathcal{X}}}{p_{\hat\theta}} + \TVp{\tilde{p}_{\mathcal{X}}}{p_{\mathcal{X}}}  \quad \text{(by triangle inequality)}  \\
    &\le \sqrt{\frac{1}{2} \bar{C}_{AT}(p_{\hat\theta}) \tilde{L}_{\text{PL}}(\hat\theta)} + \sqrt{\frac{\abs{\Omega}^{3N}}{8 \delta n}} \quad \text{(by \Cref{eq:conditional_to_joint} and \Cref{eq:empirical_pmf})}  \\
    &< \sqrt{\frac{1}{2} \bar{C}_{AT}(p_{\hat\theta}) \left( \hat{L}_{PL}(\hat\theta) + \left( \sqrt{\frac{2^{3N} \abs{\Omega}^N C_\epsilon(\Theta)}{m \cdot \delta}} +  \sqrt{\frac{\ln\frac{8 C_\epsilon(\Theta)}{\delta}}{2 n}} \right) \cdot \ln \frac{1}{\margin} + \epsilon \right)} + \sqrt{\frac{\abs{\Omega}^{3N}}{8 \delta n}} \\ 
    &\quad \text{(by \Cref{eq:generalization_bound_conditional})}  \\
\end{align*}

This completes the proof of the Theorem. 
\end{proof}

We proceed to Step 1 first. We show: 
\begin{proposition} \label{prop:conditional_to_joint}
$\KL \left( p_{\mathcal{X}}, p_\theta \right) 
\le \bar{C}_{AT}(p_\theta) L_{\text{PL}}(\theta)$
and 
$\KL \left( \tilde{p}_{\mathcal{X}}, p_\theta \right) 
\le \bar{C}_{AT}(p_\theta) \tilde{L}_{\text{PL}}(\theta)$
\end{proposition}

\begin{proof}

By definition of block-generalized approximate tensorization of entropy in \Cref{d:block_catconst}
\begin{align*}
    \KL(p_{\mathcal{X}}, p_\theta) &\le \bar{C}_{AT}(p_{{\theta}}) \cdot \Ep{X \sim p_{\mathcal{X}}}{
    \Ep{K \sim p_{\mathcal{K}}(\cdot \mid X)}{
        \KLp{p_{\mathcal{X}}(\cdot \mid X_{-K}, K)}{p_\theta(\cdot \mid X_{-K}, K}
    }
} \\
    &= \bar{C}_{AT}(p_{{\theta}}) \cdot L_{\text{PL}}({\theta}) 
\end{align*}

Likewise the latter holds when we replace $p$ with $\tilde{p}$.

\end{proof}



We will need the following simple observation in several concentration bounds we prove: 

\begin{proposition}[Bound on KL] \label{prop:KL_bounded}
Under \Cref{as:support_margin},
\[ \KLp{\tilde{p}_{\mathcal{X}}(\cdot | X_{-K}, K)}{p_\theta(\cdot | X_{-K}, K)} \in \left[0, \ln \frac{1}{\margin}\right] \]
\end{proposition}

\begin{proof}

By definition of $\KL$, 
\begin{align*}
    0 \le \KLp{\tilde{p}_{\mathcal{X}}(\cdot | X_{-K}, K)}{p_\theta(\cdot | X_{-K}, K)} &= \sum_{X_K \in \Omega^{\abs{K}}} \tilde{p}_{\mathcal{X}}(X_K | X_{-K}, K) \ln \frac{\tilde{p}_{\mathcal{X}}(X_K | X_{-K}, K)}{p_\theta(X_K | X_{-K}, K)} \\
    &\le \sum_{X_K \in \Omega^{\abs{K}}} \tilde{p}_{\mathcal{X}}(X_K | X_{-K}, K) \ln \frac{1}{p_\theta(X_K | X_{-K}, K)} \\
    &\le \sum_{X_K \in \Omega^{\abs{K}}} \tilde{p}_{\mathcal{X}}(X_K | X_{-K}, K) \ln \frac{1}{\margin} \quad \text{(by \Cref{as:support_margin})} \\
    &= \ln \frac{1}{\margin}
\end{align*}

\end{proof}

we also recall a standard version of Hoeffding's inequality we'll use repeatedly: 

\begin{lemma}[Hoeffding's inequality] \label{lem:hoeffding}
Let $Y_1, \cdots, Y_n$ be independent random variables such that 
$a \leq Y_{i} \leq b$ almost surely. 
Consider the sum of these random variables,
$S_n = Y_1 + \cdots + Y_n$ whose expectation is $\Ep{}{S_n}$.
Then, $\forall t > 0$,
with probability at least $1 - 2 e^{-\frac{2 t^2}{n (b-a)^2}}$,
we have $\abs{S_n - \Ep{}{S_n}} < t.$
\end{lemma}

Most of the generalization bounds we need for Step 2 (in particular, \Cref{lem:generalization_bound_conditional}) will be derived from the following Lemma:

\begin{lemma}[Point-wise generalization bound for learning conditional distributions] \label{lem:point_wise_generalization_bound_conditional}
Fix a $\theta \in \Theta$ satisfying \Cref{as:support_margin}. 
$\forall \epsilon, t > 0$,
with probability at least $1 - \frac{2^{N-2} \abs{\Omega}^N}{\epsilon^2 m} - 2 e^{-\frac{2 t^2}{n \cdot \left(\ln \frac{1}{\margin}\right)^2}}$,
we have
\[
\abs{\hat{L}_{PL}(\theta) - \tilde{L}_{\text{PL}}(\theta)}
< 2^N \epsilon \cdot \ln \frac{1}{\margin} + \frac{t}{n}
\]
\end{lemma}

\begin{proof}

For notational convenience, let us denote by $\mathcal{S}_{\mathcal{X}}$ the training data points $\{X^{(i)}\}_{i \in [n]}$, and let us denote by $\mathcal{S}_{\mathcal{K}}(X)$ the set of masks corresponding to the training data point $X$. 

\textbf{Step 1: concentration over masked configurations}

We first prove that $\hat{L}_{PL}(\theta)$ (\Cref{d:genmple}) concentrates to the expectation over masked positions $K$ as $m$ increases.~\footnote{
Note that the terms $\KLp{\tilde{p}_{\mathcal{X}}(\cdot | X_{-K}, K)}{p_\theta(\cdot | X_{-K}, K)}$ are (generally) not independent for different $K$.
Besides, the terms $\sum_{K \in \mathcal{S_K}} \KLp{\tilde{p}_{\mathcal{X}}(\cdot | X_{-K}, K)}{p_\theta(\cdot | X_{-K}, K)}$ are (generally) not independent for different $\mathcal{S_K}$.
}

Denote 
\begin{equation} \label{eq:instance_X_expectation_K}
    f(X) \coloneqq \Ep{K \sim p_{\mathcal{K}}(\cdot \mid X)}{\KLp{\tilde{p}_{\mathcal{X}}(\cdot | X_{-K}, K)}{p_\theta(\cdot | X_{-K}, K)}}
\end{equation}

Then the expectation of $\hat{L}_{PL}(\theta)$ over the randomness of $\mathcal{S_K}$ is:
\begin{align} \label{eq:sum_X_expectation_K}
    \Ep{\{\mathcal{S_K}(j) \, | \, j \in [n]\}}{\hat{L}_{PL}(\theta)} &= \frac{1}{n} \sum_{X \in \mathcal{S_X}} \frac{1}{m} \Ep{\mathcal{S_K}(j)}{\sum_{K \in \mathcal{S_K}(X)} \KLp{\tilde{p}(\cdot | X_{-K}, K)}{p_\theta(\cdot | X_{-K}, K)}}  \nonumber \\
    &= \frac{1}{n} \sum_{X \in \mathcal{S_X}} \Ep{K \sim p_{\mathcal{K}}(\cdot \mid X)}{\KLp{\tilde{p}_{\mathcal{X}}(\cdot | X_{-K}, K)}{p_\theta(\cdot | X_{-K}, K)}} \nonumber \\
    &= \frac{1}{n} \sum_{X \in \mathcal{S_X}} f(X)
\end{align}


Moreover, for each $K$, the (observed) empirical probability $p_S(K \mid X)$ 
converges to the true probability $p_{\mathcal{K}}(K \mid X)$ as $m$ increases,
because the count, $p_S(K \mid X) \cdot m$, follows the binomial distribution $\text{Binomial}(m, p_{\mathcal{K}}(K \mid X)).$
More specifically, by Chebyshev's inequality, $\forall \epsilon > 0$, and a fixed $X$ we have:
\begin{align*}
    \probp{\abs{p_S(K \mid X) - p_{\mathcal{K}}(K \mid X)} \ge \epsilon} &= \probp{\abs{p_S(K \mid X) m - p_{\mathcal{K}}(K \mid X) m} \ge \epsilon m}  \\
    &\le \frac{\Varp{p_S(K \mid X) m}}{\epsilon^2 m^2}  \quad \text{(Chebyshev's inequality)}  \\
    &= \frac{m p_{\mathcal{K}}(K \mid X) (1-p_{\mathcal{K}}(K \mid X))}{\epsilon^2 m^2}  \quad \text{(since $p_S(K) m \sim \text{Binomial}(m, p(K))$)}  \\
    &= \frac{ p_{\mathcal{K}}(K \mid X) (1-p_{\mathcal{K}}(K \mid X))}{\epsilon^2 m}  \\
    &\le \frac{1}{4 \epsilon^2 m}
\end{align*}
Applying union bound over $K \in \{0, 1\}^N, X \in \mathcal{X}$,
\begin{equation} \label{eq:concentration_of_mask_prob}
    \probp{\abs{p_S(K \mid X) - p_{\mathcal{K}}(K \mid X)} < \epsilon, \, \forall K \in \{0, 1\}^N, \forall X \in \mathcal{S}_{\mathcal{X}}} 
    \ge 1 - \frac{2^N \abs{\Omega}^N}{4 \epsilon^2 m}
    = 1 - \frac{2^{N-2} \abs{\Omega}^N}{\epsilon^2 m} 
\end{equation}

Plugging into \Cref{eq:instance_X_expectation_K} and \Cref{eq:sum_X_expectation_K},
we get with probability at least $1 - \frac{2^{N-2} \abs{\Omega}^N}{\epsilon^2 m}$,
\begin{align} \label{eq:concentration_over_K}
    &\quad \abs{\hat{L}_{PL}(\theta) - \frac{1}{n} \sum_{X \in \mathcal{S_X}} f(X)}  \nonumber  \\
    &= \large| \frac{1}{n} \sum_{X \in \mathcal{S_X}} \frac{1}{|\mathcal{S_K}(X)|} \sum_{K \in \mathcal{S_K}(X)} \KLp{\tilde{p}_{\mathcal{X}}(\cdot | X_{-K}, K)}{p_\theta(\cdot | X_{-K}, K)}  \nonumber \\
    &\quad - \frac{1}{n} \sum_{X \in \mathcal{S_X}} \Ep{K}{\KLp{\tilde{p}_{\mathcal{X}}(\cdot | X_{-K}, K)}{p_\theta(\cdot | X_{-K}, K)}} \large|  \nonumber \\
    &\le \frac{1}{n} \sum_{X \in \mathcal{S_X}} \sum_{K \in \{0, 1\}^N} \abs{p_S(K \mid X) - p_{\mathcal{K}}(K \mid X)} \cdot \KLp{\tilde{p}_{\mathcal{X}}(\cdot | X_{-K}, K)}{p_\theta(\cdot | X_{-K}, K)}  \quad \text{(triangle inequality)}  \nonumber \\
    &< \frac{1}{n} \sum_{X \in \mathcal{S_X}} \sum_{K \in \{0, 1\}^N} \epsilon \cdot \KLp{\tilde{p}_{\mathcal{X}}(\cdot | X_{-K}, K)}{p_\theta(\cdot | X_{-K}, K)}  \quad \text{(by \Cref{eq:concentration_of_mask_prob})}  \nonumber \\
    &\le \frac{1}{n} \sum_{X \in \mathcal{S_X}} \sum_{K \in \{0, 1\}^N} \epsilon \cdot \ln \frac{1}{\margin}  \quad \text{(by \Cref{prop:KL_bounded})}  \nonumber \\
    &= \frac{1}{n} \sum_{X \in \mathcal{S_X}} 2^N \epsilon \cdot \ln \frac{1}{\margin} 
    \quad = 2^N \epsilon \cdot \ln \frac{1}{\margin} 
\end{align}



\textbf{Step 2: concentration over sequences $X$ in training data.}

Recall $f(X)$ defined in \Cref{eq:instance_X_expectation_K}. We have:
\begin{align*}
    \Ep{}{f(X)} &= \Ep{X \sim \tilde{p}_{\mathcal{X}}}{\Ep{K \sim p_{\mathcal{K}}(\cdot \mid X)}{ \KLp{\tilde{p}(\cdot | X_{-K}, K)}{p_\theta(\cdot | X_{-K}, K)}}} \\
    &= \tilde{L}_{\text{PL}}(\theta)
\end{align*}

Note that $f(X) \in [0, \ln \frac{1}{\margin}]$ by \Cref{prop:KL_bounded}. Thus, applying Hoeffding's inequality (\Cref{lem:hoeffding}),
$\forall t > 0$,
with probability at least $1 - 2 e^{-\frac{2 t^2}{n \cdot \left(\ln \frac{1}{\margin}\right)^2}}$,
we have
\begin{equation} \label{eq:concentration_over_X}
    \abs{\frac{1}{n} \sum_{X \in \mathcal{S_X}} f(X) - \Ep{}{f(X)}} < \frac{t}{n}
\end{equation}

\textbf{Step 3: combining results: concentration over both masks $K$ and sequences $X$.}

By union bound, 
with probability at least 
\[ 
1 - \frac{2^{N-2} \abs{\Omega}^N}{\epsilon^2 m} - 2 e^{-\frac{2 t^2}{n \cdot \left(\ln \frac{1}{\margin}\right)^2}}
\]
both \Cref{eq:concentration_over_K} and \Cref{eq:concentration_over_X} hold, 
giving us: 
\begin{align*}
    &\quad \abs{\hat{L}_{PL}(\theta) - \tilde{L}_{\text{PL}}(\theta)} \\
    &\le \abs{\hat{L}_{PL}(\theta) - \frac{1}{n} \sum_{X \in \mathcal{S_X}} f(X)} 
    + \abs{\frac{1}{n} \sum_{X \in \mathcal{S_X}} f(X) - \tilde{L}_{\text{PL}}(\theta)}  \quad \text{(triangle inequality)} \\
    &< 2^N \epsilon \cdot \ln \frac{1}{\margin} + \frac{t}{n}
\end{align*}

\end{proof}

\begin{remark}
The two terms in the bound given by \Cref{lem:point_wise_generalization_bound_conditional}, i.e. 
$2^N \epsilon \cdot \ln \frac{1}{\margin}$ and $\frac{t}{n}$,
can be controlled by setting appropriate $\epsilon$ and $t$ based on $m$ and $n$, respectively.
These two terms can reduce by increasing $m$ and $n$, respectively, as we will show in the subsequent corollary. 
This is intuitive: we expect a smaller generalization gap when the model is trained on more mask configurations for each sequence, and when more sequences are included in the data.
The first term grows with $N$ --- this is also intuitive:
when the sequences are longer, it is natural to require observing more mask configurations.
\end{remark}

\begin{corollary}[Point-wise generalization bound for learning conditional distributions, special case] \label{cor:point_wise_generalization_bound_conditional_special}
Fix a $\theta \in \Theta$ satisfying \Cref{as:support_margin}. 
with probability at least $1 - \delta$,
we have
\[
\abs{\hat{L}_{PL}(\theta) - \tilde{L}_{\text{PL}}(\theta)}
< \left( \sqrt{\frac{2^{3N-1} \abs{\Omega}^N}{m \cdot \delta}} +  \sqrt{\frac{\ln\frac{4}{\delta}}{2 n}} \right) \cdot \ln \frac{1}{\margin}
\]
\end{corollary}

\begin{proof}
Apply \Cref{lem:point_wise_generalization_bound_conditional} with $\epsilon$ and $t$ satisfying
\begin{align*}
    \epsilon &= \sqrt{\frac{2^{N-1} \abs{\Omega}^N}{m \cdot \delta}} \\
    t &= \sqrt{\frac{\ln\frac{4}{\delta} \cdot n}{2}} \ln\frac{1}{\margin}
\end{align*}
we have that with probability at least $1 - \delta$, it holds:
\begin{align*}
&\quad \abs{\hat{L}_{PL}(\theta) - \tilde{L}_{\text{PL}}(\theta)} \\
&< 2^N \epsilon \cdot \ln \frac{1}{\margin} + \frac{t}{n} \\
&= \left( \sqrt{\frac{2^{3N-1} \abs{\Omega}^N}{m \cdot \delta}} +  \sqrt{\frac{\ln\frac{4}{\delta}}{2 n}} \right) \cdot \ln \frac{1}{\margin}
\end{align*}
\end{proof}

\begin{corollary}[Uniform convergence generalization bound for learning conditional distributions] \label{lem:generalization_bound_conditional}
Under \Cref{as:support_margin} and \Cref{as:complexity}, 
$\forall \delta \in (0, 1)$,
$\forall \epsilon > 0$,
with probability at least $1 - \delta$,
we have
\[
\abs{\hat{L}_{PL}(\theta) - \tilde{L}_{\text{PL}}(\theta)}
< \left( \sqrt{\frac{2^{3N-1} \abs{\Omega}^N C_\epsilon(\Theta)}{m \cdot \delta}} +  \sqrt{\frac{\ln\frac{4 C_\epsilon(\Theta)}{\delta}}{2 n}} \right) \cdot \ln \frac{1}{\margin} + \epsilon
\]
\end{corollary}

\begin{proof}
By \Cref{as:complexity}, let $C_\epsilon(\Theta)$ denote the complexity of parameter space $\Theta$,
with the corresponding partition $\text{Par}_\epsilon(\Theta) = \{ \Theta_1, \cdots, \Theta_{C_\epsilon(\Theta)} \}$.
As a corollary of \Cref{as:complexity},
$\forall i, \forall \theta_1, \theta_2 \in \Theta_i$,
$
\abs{\tilde{L}_{PL}(\theta_1) - \tilde{L}_{PL}(\theta_2)} \le \frac{\epsilon}{2}
$,
$
\abs{\hat{L}_{PL}(\theta_1) - \hat{L}_{PL}(\theta_2)} \le \frac{\epsilon}{2}
$.

Moreover, for each $i \in [C_\epsilon(\Theta)]$,
arbitrarily select any point $\theta_i^* \in \Theta_i$ (as a ``representative" of that region of the parameter space).
Let the set of ``representative points" be $\Theta^* = \{ \theta_i^* \, | \, i \in [C_\epsilon(\Theta)] \}$. By \Cref{cor:point_wise_generalization_bound_conditional_special}, fixing any $\theta \in \Theta$ satisfying \Cref{as:support_margin}, then
with probability at least $1 - \frac{\delta}{C_\epsilon(\Theta)}$,
we have
\[
\abs{\hat{L}_{PL}(\theta) - \tilde{L}_{\text{PL}}(\theta)}
< \left( \sqrt{\frac{2^{3N-1} \abs{\Omega}^N C_\epsilon(\Theta)}{m \cdot \delta}} +  \sqrt{\frac{\ln\frac{4 C_\epsilon(\Theta)}{\delta}}{2 n}} \right) \cdot \ln \frac{1}{\margin}
\]

Applying union bound over $\theta_i^* \in \Theta^*$,
since $\abs{\Theta^*} = C_\epsilon(\Theta)$,
with probability at least $1 - \delta$,
\begin{equation} \label{eq:union_bound}
    \forall i \in [C_\epsilon(\Theta)], \quad 
    \abs{L_{\text{PL}}(\theta_i^*) - \tilde{L}_{\text{PL}}(\theta_i^*)}
    < \left( \sqrt{\frac{2^{3N-1} \abs{\Omega}^N C_\epsilon(\Theta)}{m \cdot \delta}} +  \sqrt{\frac{\ln\frac{4 C_\epsilon(\Theta)}{\delta}}{2 n}} \right) \cdot \ln \frac{1}{\margin}
\end{equation}

Finally, by \Cref{as:complexity}, $\forall \theta \in \Theta$,
there exists $i \in [C_\epsilon(\Theta)]$ such that $\theta \in \Theta_i$ (i.e. $\theta$ falls into that partition), and 
\begin{align} \label{eq:representative}
    \abs{\tilde{L}_{\text{PL}}(\theta) - \tilde{L}_{\text{PL}}(\theta_i^*)} &\le \frac{\epsilon}{2}  \nonumber  \\
    \abs{\hat{L}_{PL}(\theta) - \hat{L}_{PL}(\theta_i^*)} &\le \frac{\epsilon}{2}
\end{align}

Combining \Cref{eq:union_bound} and \Cref{eq:representative} gives
\begin{align*}
    &\quad \abs{\hat{L}_{PL}(\theta) - \tilde{L}_{\text{PL}}(\theta)} \\
    &\le \abs{\hat{L}_{PL}(\theta) - \hat{L}_{PL}(\theta_i^*)} + \abs{\hat{L}_{PL}(\theta_i^*) - \tilde{L}_{\text{PL}}(\theta_i^*)}  \\
    &\quad + \abs{\tilde{L}_{\text{PL}}(\theta_i^*) - \tilde{L}_{\text{PL}}(\theta)} \quad \text{(by triangle inequality)} \\
    &< \frac{\epsilon}{2} + \left( \sqrt{\frac{2^{3N-1} \abs{\Omega}^N C_\epsilon(\Theta)}{m \cdot \delta}} +  \sqrt{\frac{\ln\frac{4 C_\epsilon(\Theta)}{\delta}}{2 n}} \right) \cdot \ln \frac{1}{\margin} + \frac{\epsilon}{2}  \quad \text{(by \Cref{eq:union_bound} and \Cref{eq:representative})} \\
    &= \left( \sqrt{\frac{2^{3N-1} \abs{\Omega}^N C_\epsilon(\Theta)}{m \cdot \delta}} +  \sqrt{\frac{\ln\frac{4 C_\epsilon(\Theta)}{\delta}}{2 n}} \right) \cdot \ln \frac{1}{\margin} + \epsilon
\end{align*}

\end{proof}

Finally, we complete Step 3 (proving \Cref{eq:empirical_pmf}): 

\begin{lemma}[Empirical PMF converges to population PMF] \label{lem:empirical_pmf}
For any $\delta > 0$,
with probability at least $1 - \delta$,
we have:
\[
\TVp{\tilde{p}_{\mathcal{X}}}{p_{\mathcal{X}}}
< \sqrt{\frac{\abs{\Omega}^{3N}}{16 \delta n}}
\]
\end{lemma}

\begin{proof}
$\forall X \in \Omega^N$, the number of times that $X$ appears in the training data $\mathcal{S_X}$ follows the binomial distribution 
\[ 
\tilde{p}_{\mathcal{X}}(X) n \sim \text{Binomial}(n, p_{\mathcal{X}}(X))
\]
with mean $n p_{\mathcal{X}}(X)$ and variance $n p_{\mathcal{X}}(X) (1 - p_{\mathcal{X}}(X))$.
Hence, by Chebyshev's inequality, $\forall \epsilon > 0$
\begin{align*}
    \probp{\abs{\tilde{p}_{\mathcal{X}}(X) - p_{\mathcal{X}}(X)} \ge \epsilon} &= \probp{\tilde{p}_{\mathcal{X}}(X) n - p_{\mathcal{X}}(X) n \ge \epsilon n}  \\
    &\le \frac{\Varp{\tilde{p}_{\mathcal{X}}(X) n}}{\epsilon^2 n^2}  \quad \text{(Chebyshev's inequality)}  \\
    &= \frac{n p_{\mathcal{X}}(X) (1-p_{\mathcal{X}}(X))}{\epsilon^2 n^2}  \quad \text{(since $\tilde{p}_{\mathcal{X}}(X) n \sim \text{Binomial}(n, p_{\mathcal{X}}(X))$)}  \\
    &= \frac{ p_{\mathcal{X}}(X) (1-p_{\mathcal{X}}(X))}{\epsilon^2 n}  \\
    &\le \frac{1}{4 \epsilon^2 m}
\end{align*}

Applying union bound over $X \in \Omega^N$,
\begin{equation} \label{eq:concentration_of_empirical_pmf}
    \probp{\abs{\tilde{p}_{\mathcal{X}}(X) - p_{\mathcal{X}}(X)} < \epsilon, \, \forall X \in \Omega^N} 
    \ge 1 - \frac{\abs{\Omega}^N}{4 \epsilon^2 n}
\end{equation}

Hence,
we get with probability at least $1 - \frac{\abs{\Omega}^N}{4 \epsilon^2 n}$,
\begin{equation} \label{eq:tv_tilde_p_vs_p}
    \TVp{\tilde{p}_{\mathcal{X}}}{p_{\mathcal{X}}} = \frac{1}{2} \sum_{X \in \Omega^N} \abs{\tilde{p}_{\mathcal{X}}(X) - p_{\mathcal{X}}(X)}
    < \frac{1}{2} \sum_{X \in \Omega^N} \epsilon
    = \frac{1}{2} \abs{\Omega}^N \epsilon 
\end{equation}

Solving for $\delta = \frac{\abs{\Omega}^N}{4 \epsilon^2 n}$ 
gives $\epsilon = \sqrt{\frac{\abs{\Omega}^N}{4 \delta n}}.$
Therefore, by \Cref{eq:concentration_of_empirical_pmf}, with probability at least $1 - \delta$,
we have
\[
\TVp{\tilde{p}_{\mathcal{X}}}{p_{\mathcal{X}}}
< \frac{1}{2} \abs{\Omega}^N \epsilon
= \sqrt{\frac{\abs{\Omega}^{3N}}{16 \delta n}}
\]

\end{proof}



\clearpage
\section{Proof of \Cref{prop:strongly_ferromagnetic_mode}: Modes of the strongly ferromagnetic Ising model}
\label{sec:appendix:proof:strongly_ferromagnetic_mode}

This section provides formal proofs for the discussion under \Cref{as:strongly_ferromagnetic} in \Cref{sec:theory:sampling:gibbs}.

\begin{restatable}[Modes of the strongly ferromagnetic Ising model]{proposition}{propStronglyFerromagneticModes} 
\label{prop:strongly_ferromagnetic_mode}
On Ising model $G$ in \Cref{eqn:ising:construction} under \Cref{as:strongly_ferromagnetic},
the regions $\mathcal{R}_{1}$ and $\mathcal{R}_{-1}$ defined in \Cref{eq:larger_mode} and \Cref{eq:smaller_mode} satisfy: 
\begin{enumerate}
    \item $\forall \vx \in \mathcal{R}_{1}, \, \forall \vy \in \mathcal{R}_{-1}, \, \forall \vz \in \{-1, 1\}^N \backslash \left(\mathcal{R}_{1} \cup \mathcal{R}_{-1}\right)$:
    $p_G(\vx) > p_G(\vy) > e^{2 J_0} p_G(\vz)$
    \item There exists a bijection $f: \mathcal{R}_{1} \mapsto \mathcal{R}_{-1}$ such that $\forall \vx \in \mathcal{R}_{1}, \, p_G(\vx) = e^{2 h_G} p_G(f(\vx))$
\end{enumerate}
\end{restatable}


\begin{proof}
Consider $\vx \in \mathcal{R}_{1}, \vy \in \mathcal{R}_{-1}$. We have: 
\begin{align*}
    \frac{p_G(\vx)}{p_G(\vy)} &= \frac{\exp{\left( \sum_{i \in [N]} \vh_i \mathbf{x}_i + \sum_{i \ne j \in C_G \subset [N]} J \mathbf{x}_i \mathbf{x}_j \right)}}{\exp{\left( \sum_{i \in [N]} \vh_i \mathbf{y}_i + \sum_{i \ne j \in C_G \subset [N]} J \mathbf{y}_i \mathbf{y}_j \right)}} \\
    &= \frac{\exp{\left( \sum_{i \in [N]} \vh_i \mathbf{x}_i \right)}}{\exp{\left( \sum_{i \in [N]} \vh_i \mathbf{y}_i \right)}} \quad \text{(since $\mathbf{x}_i \mathbf{x}_j = \mathbf{y}_i \mathbf{y}_j = 1, \, \forall \vx \in \mathcal{R}_{1}, \forall \vy \in \mathcal{R}_{-1}$)} \\
    &= \frac{\exp{\left( \sum_{i \in C_G} \vh_i \mathbf{x}_i + \sum_{i \notin C_G} \vh_i \mathbf{x}_i \right)}}{\exp{\left( \sum_{i \in C_G} \vh_i \mathbf{y}_i + \sum_{i \notin C_G} \vh_i \mathbf{y}_i \right)}}  \\
    &= \frac{\exp{\left( \sum_{i \in C_G} \vh_i + \sum_{i \notin C_G} \vh_i \mathbf{x}_i \right)}}{\exp{\left( - \sum_{i \in C_G} \vh_i + \sum_{i \notin C_G} \vh_i \mathbf{y}_i \right)}} \quad \text{(since $\vx \in \mathcal{R}_{1}, \vy \in \mathcal{R}_{-1}$)}  \\
    &\ge \frac{\exp{\left( \sum_{i \in C_G} \vh_i - \sum_{i \notin C_G} \abs{\vh_i} \right)}}{\exp{\left( - \sum_{i \in C_G} \vh_i + \sum_{i \notin C_G} \abs{\vh_i} \right)}} \quad \text{(since $\mathbf{x}_i, \mathbf{y}_i \in \pm 1$)} \\
    &= \exp{\left( 2 \sum_{i \in C_G} \vh_i - 2 \sum_{i \notin C_G} \abs{\vh_i} \right)} \\
    &> \exp{(0)} \quad \text{(by \Cref{as:strongly_ferromagnetic})} \\
    & = 1
\end{align*}

On the other hand, if we consider $\vy \in \mathcal{R}_{-1}, \vz \in \{-1, 1\}^N \backslash \left(\mathcal{R}_{1} \cup \mathcal{R}_{-1}\right)$, we have: 

\begin{align*}
\frac{p_G(\vy)}{p_G(\vz)} &= \frac{\exp{\left( \sum_{i \in [N]} \vh_i \mathbf{y}_i + \sum_{i \ne j \in C_G \subset [N]} J \mathbf{y}_i \mathbf{y}_j \right)}}{\exp{\left( \sum_{i \in [N]} \vh_i \mathbf{z}_i + \sum_{i \ne j \in C_G \subset [N]} J \mathbf{z}_i \mathbf{z}_j \right)}} \\
&= \frac{\exp{\left( \sum_{i \in [N]} \vh_i \mathbf{y}_i + \sum_{i \ne j \in C_G \subset [N]} J \right)}}{\exp{\left( \sum_{i \in [N]} \vh_i \mathbf{z}_i + \sum_{i \ne j \in C_G \subset [N]} J \mathbf{z}_i \mathbf{z}_j \right)}} \quad \text{(since $\mathbf{y}_i \mathbf{y}_j = 1, \, \forall \vy \in \mathcal{R}_{-1}$)} \\
&\ge \frac{\exp{\left( \sum_{i \in [N]} \vh_i \mathbf{y}_i + \sum_{i \ne j \in C_G \subset [N]} J \right)}}{\exp{\left( \sum_{i \in [N]} \vh_i \mathbf{z}_i + \sum_{i \ne j \in C_G \subset [N]} J - 2 (\abs{C_G} - 1) J \right)}} \\
&\quad \text{the above is because, since $\vz \in \{-1, 1\}^N \backslash \left(\mathcal{R}_{1} \cup \mathcal{R}_{-1}\right)$, obviously the denominator} \\
&\quad \text{
$
\sum_{i \ne j \in C_G \subset [N]} J \mathbf{z}_i \mathbf{z}_j
\le \sum_{i \ne j \in C_G \subset [N]} J - 2 (\abs{C_G} - 1) J
$
} \\
&= \frac{\exp{\left( \sum_{i \in [N]} \vh_i \mathbf{y}_i \right)}}{\exp{\left( \sum_{i \in [N]} \vh_i \mathbf{z}_i - 2 (\abs{C_G} - 1) J \right)}} 
\ge \frac{\exp{\left( \sum_{i \in [N]} \vh_i \mathbf{y}_i \right)}}{\exp{\left( \sum_{i \in [N]} \vh_i \mathbf{z}_i - 2 J \right)}}
\ge \frac{\exp{\left( - \| \vh \|_1 \right)}}{\exp{\left( \| \vh \|_1 - 2 J \right)}} \\
&= \exp{\left( 2 (J - \| \vh \|_1) \right)} 
\ge \exp{\left( 2 J_0 \right)}  \quad \text{(by \Cref{as:strongly_ferromagnetic})} 
\end{align*}

Proceeding to Part 2, 
let's define $f: \mathcal{R}_{1} \mapsto \mathcal{R}_{-1}$ as:
\[
\forall \vx \in \mathcal{R}_{1}, \quad
f(\vx)_i = \begin{cases}
        -1, \quad &\text{if } i \in C_G \\ 
        \vx_i, \quad &\text{if } i \notin C_G \\ 
        \end{cases}
\]
Let $\vw \coloneqq f(\vx)$. Then, 
\begin{align*}
    \frac{p_G(\vx)}{p_G(\vw)} &= \frac{\exp{\left( \sum_{i \in [N]} \vh_i \mathbf{x}_i + \sum_{i \ne j \in C_G \subset [N]} J \mathbf{x}_i \mathbf{x}_j \right)}}{\exp{\left( \sum_{i \in [N]} \vh_i \mathbf{w}_i + \sum_{i \ne j \in C_G \subset [N]} J \mathbf{w}_i \mathbf{w}_j \right)}} \\
    &= \frac{\exp{\left( \sum_{i \in [N]} \vh_i \mathbf{x}_i \right)}}{\exp{\left( \sum_{i \in [N]} \vh_i \mathbf{w}_i \right)}} \quad \text{(since $\mathbf{x}_i \mathbf{x}_j = \mathbf{w}_i \mathbf{w}_j = 1, \, \forall \vx \in \mathcal{R}_{1}, \forall \vw \in \mathcal{R}_{-1}$)} \\
    &= \frac{\exp{\left( \sum_{i \in C_G} \vh_i \mathbf{x}_i + \sum_{i \notin C_G} \vh_i \mathbf{x}_i \right)}}{\exp{\left( \sum_{i \in C_G} \vh_i \mathbf{w}_i + \sum_{i \notin C_G} \vh_i \mathbf{w}_i \right)}}  \\
    &= \frac{\exp{\left( \sum_{i \in C_G} \vh_i + \sum_{i \notin C_G} \vh_i \mathbf{x}_i \right)}}{\exp{\left( - \sum_{i \in C_G} \vh_i + \sum_{i \notin C_G} \vh_i \mathbf{w}_i \right)}} \quad \text{(since $\vx \in \mathcal{R}_{1}, \vw \in \mathcal{R}_{-1}$)}   \\
    &= \frac{\exp{\left( \sum_{i \in C_G} \vh_i \right)}}{\exp{\left( - \sum_{i \in C_G} \vh_i \right)}} \quad \text{(since $\vw_i = \vx_i, \, \forall i \notin C_G$)}   \\
    &= \exp{\left( 2 \sum_{i \in C_G} \vh_i \right)}    \\
    &= \exp{\left( 2 h_G \right)} \quad \text{(by \Cref{as:strongly_ferromagnetic})}  
\end{align*}
\end{proof}

\clearpage
\section{Proof of \Cref{prop:mode_fast}: $k$-Gibbs sampler can reach the mode fast}
\label{sec:appendix:proof:mode_fast}

\propModeFast*

\begin{proof}
At any step, let $K$ (with $\abs{K} = k$) denote the set of coordinates to re-sample.
We first consider the probability of $C_G \subset K$,
which allows the whole $C_G$ to be updated jointly: 
The total number of ways to select $K$ is $N \choose k$.
Now count the number of $K$'s which satisfy $C_G \subset K$. 
This is the same as selecting the remaining $k - |C_G|$ coordinates from the other $N - |C_G|$ coordinates which are not in $C_G$.
So there are $N-|C_G| \choose k-|C_G|$ distinct $K$'s which satisfy $C_G \subset K$.

\begin{equation} \label{eq:include_clique_prob}
    \probp{C_G \subset K} = \frac{{N-|C_G| \choose k-|C_G|}}{{N \choose k}}
\end{equation}

$\forall t \in \N, \forall \mX^{(t)} \in \{-1, 1\}^N$,
and $K \in [N]$ such that $|K| = k$ and $C_G \subset K$,
consider $X_K^{(t+1)} \sim p_G(\cdot \mid X_{-K}^{(t)})$.
There are three cases (whhich exhaust all possibilities):
\begin{enumerate}
    \item $\mX^{(t+1)} \in \mathcal{R}_{1}$
    \item $\mX^{(t+1)} \in \mathcal{R}_{-1}$
    \item $\mX^{(t+1)} \in \{-1, 1\}^N \backslash (\mathcal{R}_{1} \cup \mathcal{R}_{-1})$
\end{enumerate}

We will use \Cref{prop:strongly_ferromagnetic_mode} to show that Case 1 occurs with probability at least a constant. We have:
\begin{align*}
    \probp{\mX^{(t+1)} \in \mathcal{R}_{1}} &= e^{2 h_G} \probp{\mX^{(t+1)} \in \mathcal{R}_{-1}}  \\
    \frac{\probp{\mX^{(t+1)} \in \mathcal{R}_{-1}}}{\probp{\mX^{(t+1)} \in \{-1, 1\}^N \backslash \left(\mathcal{R}_{1} \cup \mathcal{R}_{-1}\right)}} &\ge e^{2 J_0} \frac{\abs{\mathcal{R}_{-1}}}{\abs{\{-1, 1\}^N \backslash \left(\mathcal{R}_{1} \cup \mathcal{R}_{-1}\right)}} 
    = \frac{e^{2 J_0}}{2^{\abs{C_G}} - 2} 
\end{align*}
Since the probabilities of the three cases sum up to 1,
\[
\probp{\mX^{(t+1)} \in \mathcal{R}_{1}} 
\ge \frac{e^{2 (J_0 + h_G)}}{e^{2 (J_0 + h_G)} + e^{2 J_0} + 2^{\abs{C_G}} - 2}
\]

Combining with \Cref{eq:include_clique_prob}, we have $\forall t \in \N$ and $\forall \mX^{(t)} \in \{-1, 1\}^N$, it holds that:
\[
    \probp{\mX^{(t+1)} \in \mathcal{R}_{1}} 
    \ge \probp{C_G \subset K, \mX^{(t+1)} \in \mathcal{R}_{1}} 
    \ge \frac{{N-|C_G| \choose k-|C_G|}}{{N \choose k}} \frac{e^{2 (J_0 + h_G)}}{e^{2 (J_0 + h_G)} + e^{2 J_0} + 2^{\abs{C_G}} - 2}
    \coloneqq 1 - c_{\mathcal{R}_{1}}
\]
From this it follows that
\[
\probp{\{ \mX^{(t)} | t \in [T] \} \cap \mathcal{R}_{1} = \emptyset}
\le c_{\mathcal{R}_{1}}^T
\]
Therefore, when $T \ge \log_{c_{\mathcal{R}_{1}}} \delta$,
\[
\probp{\{ \mX^{(t)} | t \in [T] \} \cap \mathcal{R}_{1} = \emptyset}
\le \delta
\]

\end{proof}

\clearpage
\section{Proof of \Cref{prop:mode_slow} independent parallel sampling stuck in bad samples}
\label{sec:appendix:proof:mode_slow}

\propModeSlow*

\begin{proof}
Suppose at step $t$, 
$\mX^{(t)}$ is such that $\sum_{i \in C_G} \mX_i^{(t)} \le -2$
(satisfied at $t = 0$),
then
\begin{equation} \label{eq:others_in_clique}
    \forall j \in C_G, \; 
    \sum_{i \in C_G, i \ne j} \mX_i^{(t)} \le -1
\end{equation}

Hence its next-step distribution 
$X_j^{(t+1)} \sim p(\cdot \mid X_{-\{j\}}^{(t)})$
satisfies

\begin{align*}
    \frac{\probp{X_j^{(t+1)} = 1}}{\probp{X_j^{(t+1)} = -1}}
    &= \frac{\exp{\left( \sum_{i \in [N]} \vh_i \mathbf{x}_i + \sum_{i \ne j \in C_G \subset [N]} J \mathbf{x}_i \mathbf{x}_j \right)} |_{\mathbf{x}_j = 1}}{\exp{\left( \sum_{i \in [N]} \vh_i \mathbf{x}_i + \sum_{i \ne j \in C_G \subset [N]} J \mathbf{x}_i \mathbf{x}_j \right)} |_{\mathbf{x}_j = -1}} \quad \text{(by definition in \Cref{eqn:ising:construction})} \\
    &= \frac{\exp{\left( \vh_j + \sum_{i \in C_G, i \ne j} J \mathbf{x}_i \right)}}{\exp{\left( -\vh_j - \sum_{i \in C_G, i \ne j} J \mathbf{x}_i \right)}} \quad \text{(canceling the same terms)} \\
    &= \exp{\left( 2 \vh_j + 2 J \sum_{i \in C_G, i \ne j} \mathbf{x}_i \right)} \\
    &\le \exp{\left( 2 \vh_j - 2 J \right)} \quad \text{(by \Cref{eq:others_in_clique})} \\
    &\le \exp{\left( - 2 J_0 \right)} \quad \text{(by \Cref{as:strongly_ferromagnetic})} 
\end{align*}

Therefore
\begin{equation} \label{eq:single_coordinate_update}
    X_j^{(t+1)} = \begin{cases}
    1, \quad &\text{with prob } \le \frac{\exp{\left( - 2 J_0 \right)}}{\exp{\left( - 2 J_0 \right)} + 1} \\ 
    -1, \quad &\text{with prob } \ge \frac{1}{\exp{\left( - 2 J_0 \right)} + 1} 
\end{cases}
\end{equation}

Denote
\begin{equation} \label{eq:single_coordinate_updates_affine}
    Y_j \coloneqq \frac{X_j^{(t+1)} + 1}{2}
\end{equation}

Note that $\{Y_j \, | \, j \in [N] \}$ are independent Bernoulli random variables.

By \Cref{lem:hoeffding},
$\forall r > 0$,
with probability at least $1 - 2 e^{-\frac{2 r^2}{\abs{C_G}}}$,
\begin{align*}
    \frac{1}{\abs{C_G}} \sum_{j \in C_G} Y_j &< \Ep{j \in C_G}{Y_j} + \frac{r}{\abs{C_G}} \quad \text{(by Hoeffding's inequality \Cref{lem:hoeffding})} \\
    &\le \frac{\exp{\left( - 2 J_0 \right)}}{\exp{\left( - 2 J_0 \right)} + 1} + \frac{r}{\abs{C_G}}  \quad \text{(by \Cref{eq:single_coordinate_update} and definition of $Y_j$ in \Cref{eq:single_coordinate_updates_affine})}
\end{align*}
implying that 
with probability at least $1 - 2 e^{-\frac{2 r^2}{\abs{C_G}}}$,
\[
\frac{1}{\abs{C_G}} \sum_{j \in C_G} X_j^{(t+1)} 
= 2 \frac{1}{\abs{C_G}} \sum_{j \in C_G} Y_j - 1
< 2 \left( \frac{\exp{\left( - 2 J_0 \right)}}{\exp{\left( - 2 J_0 \right)} + 1} + \frac{r}{\abs{C_G}} \right) - 1
\]
i.e.
\[
\sum_{j \in C_G} X_j^{(t+1)} < \frac{\exp{\left( - 2 J_0 \right)} - 1}{\exp{\left( - 2 J_0 \right)} + 1} \abs{C_G} + 2 r
\]

Setting RHS to -2 solves to 
\[
r = -1 + \frac{1 - \exp{\left( - 2 J_0 \right)}}{\exp{\left( - 2 J_0 \right)} + 1} \frac{\abs{C_G}}{2}
\]

Hence
\begin{equation} \label{eq:sum_coordinate_update}
    \text{with probability at least } 
    1 - 2 e^{-\frac{2 \left( -1 + \frac{1 - \exp{\left( - 2 J_0 \right)}}{\exp{\left( - 2 J_0 \right)} + 1} \frac{\abs{C_G}}{2} \right)^2}{\abs{C_G}}}, \quad
    \sum_{j \in C_G} X_j^{(t+1)} < -2
\end{equation}

By union bound, $\forall T \in \N_+$,
\begin{equation} \label{eq:sum_coordinate_update_union}
    \text{with probability at least } 
    1 - 2 T e^{-\frac{2 \left( -1 + \frac{1 - \exp{\left( - 2 J_0 \right)}}{\exp{\left( - 2 J_0 \right)} + 1} \frac{\abs{C_G}}{2} \right)^2}{\abs{C_G}}}, \quad
    \forall t \in [T], \quad
    \sum_{j \in C_G} X_j^{(t)} < -2
\end{equation}

Note that when $\sum_{j \in C_G} X_j^{(t)} < -2$,
$\mX^{(t)} \notin \mathcal{R}_{1}$.

Finally, aligning the probabilities:
setting
\[
2 T e^{-\frac{2 \left( -1 + \frac{1 - \exp{\left( - 2 J_0 \right)}}{\exp{\left( - 2 J_0 \right)} + 1} \frac{\abs{C_G}}{2} \right)^2}{\abs{C_G}}} 
= \delta
\]
solves to
\[
T = \frac{\delta}{2} e^{\frac{2 \left( -1 + \frac{1 - \exp{\left( - 2 J_0 \right)}}{\exp{\left( - 2 J_0 \right)} + 1} \frac{\abs{C_G}}{2} \right)^2}{\abs{C_G}}} 
\]

\end{proof}

\clearpage
\section{Proof of \Cref{prop:mode_separation}: Separation between $N$-Gibbs sampler and independent parallel sampling}
\label{sec:appendix:proof:mode_separation}

This section provides additional information for the discussion at the end of \Cref{sec:theory:sampling:gibbs}.

\propModeSeparation*

\begin{proof}
Under the given conditions, 
with \textbf{$k$-Gibbs sampler},
by \Cref{prop:mode_fast}, 
\begin{equation} \label{eq:mode_fast_N}
    \text{with probability at least }
    1 - \frac{\delta}{2}, \quad
    \{ \mX_{\text{k-Gibbs}}^{(t)} | t \in [\ceil{\log_{c_{\mathcal{R}_{1}}} \frac{\delta}{2}}] \} \cap \mathcal{R}_{1} \ne \emptyset
\end{equation}
in which the constant 
\begin{equation} \label{eq:enter_mode_prob_N}
c_{\mathcal{R}_{1}} \coloneqq 1 - \frac{{N-|C_G| \choose k-|C_G|}}{{N \choose k}} \frac{e^{2 (J_0 + h_G)}}{e^{2 (J_0 + h_G)} + e^{2 J_0} + 2^{\abs{C_G}} - 2}
\end{equation}

Applying \Cref{as:strong_interactions} to bound parts of the RHS of \Cref{eq:enter_mode_prob_N}:
\[
\frac{e^{2 J_0}}{e^{2 (J_0 + h_G)}}
= e^{-2 h_G}
\]
\[
\frac{2^{\abs{C_G}} - 2}{e^{2 (J_0 + h_G)}}
\le \frac{2^{\abs{C_G}}}{e^{2 (J_0 + h_G)}}
\le \frac{2^{\abs{C_G}}}{e^{\abs{C_G} \ln{2} + 2 h_G}}
= \frac{2^{\abs{C_G}}}{2^{\abs{C_G}}  e^{2 h_G}}
= e^{-2 h_G}
\]
Taking the sum:
\[
\frac{e^{2 J_0} + 2^{\abs{C_G}} - 2}{e^{2 (J_0 + h_G)}}
\le 2 e^{-2 h_G} 
\]
Adding 1 to both sides:
\[
\frac{e^{2 (J_0 + h_G)} + e^{2 J_0} + 2^{\abs{C_G}} - 2}{e^{2 (J_0 + h_G)}}
\le 1 + 2 e^{-2 h_G}  
\]
Taking the inverse:
\begin{equation}
\label{eq:enter_mode_prob_N_part1}
\frac{e^{2 (J_0 + h_G)}}{e^{2 (J_0 + h_G)} + e^{2 J_0} + 2^{\abs{C_G}} - 2}
\ge \frac{1}{1 + 2 e^{-2 h_G}} 
\ge \frac{1}{1 + 2 e^{-2 \frac{1}{2} \ln{\frac{2 (4-\delta)}{\delta}}}}
= \frac{1}{1 + 2 \frac{\delta}{2 (4-\delta)} }
= \frac{1}{1 + \frac{\delta}{4-\delta} }
= \frac{4-\delta}{ 4 }
\end{equation}

Similarly, applying \Cref{as:large_k} to bound the other parts of the RHS of \Cref{eq:enter_mode_prob_N}:
\begin{align*}
\frac{{N-|C_G| \choose k-|C_G|}}{{N \choose k}}
&= \frac{ \frac{\inparen{N-|C_G|}!}{\inparen{k-|C_G|}! \inparen{N-k}!} }{ \frac{N!}{k! \inparen{N-k}!} }
= \frac{ \frac{\inparen{N-|C_G|}!}{\inparen{k-|C_G|}!} }{ \frac{N!}{k!} }
= \frac{ \inparen{k+1-|C_G|} \cdots \inparen{N-|C_G|} }{ (k+1) \cdots N } \\
&\ge \inparen{ \frac{ k+1-|C_G| }{ k+1 } }^{(N-k)}
= \inparen{ 1 - \frac{ |C_G| }{ k+1 } }^{(N-k)} \\
&\ge 1 - (N-k) \frac{ |C_G| }{ k+1 } \quad \text{(since $\frac{ |C_G| }{ k+1 } \in (0, 1)$ )} \\
&\ge \frac{4 - 2 \delta}{4 - \delta} \quad \text{(by \Cref{as:large_k} and straightforward calculation)}
\end{align*}

Plugging in the above and \Cref{eq:enter_mode_prob_N_part1} to \Cref{eq:enter_mode_prob_N}:
\[
c_{\mathcal{R}_{1}} 
\le 1 - \frac{4 - 2 \delta}{4 - \delta} \cdot \frac{ 4-\delta }{4}
= \frac{ \delta }{2}
\]
Plugging into \Cref{eq:mode_fast_N}:
\begin{equation} \label{eq:mode_fast_N_simplified}
    \text{with probability at least }
    1 - \frac{\delta}{2}, \quad
    \{ \mX_{\text{k-Gibbs}}^{(t)} | t \in [1] \} \cap \mathcal{R}_{1} \ne \emptyset, \quad
    \text{namely $\mX_{\text{k-Gibbs}}^{(1)} \in \mathcal{R}_{1}$}
\end{equation}

On the other hand, 
with \textbf{independent parallel},
by \Cref{prop:mode_slow}, 
\begin{equation} \label{eq:mode_slow_strong}
    \text{with probability at least }
    1 - \frac{\delta}{2}, \quad
    \{ \mX_{indep}^{(t)} | t \in [\floor{\frac{\delta}{4} \exp{(c_{\text{stuck}})}}] \} \cap \mathcal{R}_{1} = \emptyset
\end{equation}
in which the constant 
\begin{equation} \label{eq:stuck_prob_strong}
c_{\text{stuck}} \coloneqq \frac{2 \left( -1 + \frac{1 - \exp{\left( - 2 J_0 \right)}}{\exp{\left( - 2 J_0 \right)} + 1} \frac{\abs{C_G}}{2} \right)^2}{\abs{C_G}}
\end{equation}

Applying \Cref{as:strong_interactions} to bound parts of the RHS:
\[
\frac{1 - \exp{\left( - 2 J_0 \right)}}{\exp{\left( - 2 J_0 \right)} + 1}
\ge \frac{1}{2}
\]
Plugging into \Cref{eq:stuck_prob_strong}:
\begin{align*}
    c_{\text{stuck}} &\ge \frac{2 \left( -1 + \frac{1}{2} \frac{\abs{C_G}}{2} \right)^2}{\abs{C_G}} \\
    &= \frac{2 \left( 1 - \frac{\abs{C_G}}{2} + \frac{\abs{C_G}^2}{4} \right) }{\abs{C_G}} \\
    &\ge -1 + \frac{\abs{C_G}}{8} \\
    &\ge -1 + \left( 1 + \ln{\frac{4 M}{\delta}} \right) \quad \text{(by \Cref{as:strong_interactions})} \\
    &= \ln{\frac{4 M}{\delta}}
\end{align*}
Plugging into \Cref{eq:mode_slow_strong}:
\begin{equation} \label{eq:mode_slow_strong_simplified}
    \text{with prob }
    \ge 1 - \frac{\delta}{2}, \quad
    \{ \mX_{indep}^{(t)} | t \in [\floor{\frac{\delta}{4} \cdot \frac{4 M}{\delta}}] = [M] \} \cap \mathcal{R}_{1} = \emptyset, \quad
    \text{namely $\mX_{indep}^{(t)} \notin \mathcal{R}_{1} \forall t \in [M]$}
\end{equation}

By union bound,
with probability at least $1 - \delta$,
both \Cref{eq:mode_fast_N_simplified} and \Cref{eq:mode_slow_strong_simplified} hold.

\end{proof}

\clearpage
\section{Background and proofs of \Cref{prop:transformer_deterministic_mc} and \Cref{prop:transformer_general_mc}: on the expressive power of Transformers for implementing sequence-to-sequence Markov chains in parallel}
\label{sec:appendix:proof:transformer_ddlm_expressive_power}

\subsection{Technical setup and proofs}

\paragraph{Background: Transformer network architecture.}
The transformer architecture \citep{vaswani2017attention} is a critical building block of many leading approaches to language modeling \citep{devlin2019bert, brown2020language}.
We refer the readers to these works for more details on the empirical promise that Transformer-based models have demonstrated.
For theoretical understanding of Transformers,
we refer the readers to prior works on their
representational power 
\citep{yun2020are, yao2021self, liu2023Transformers, zhao2023Transformers}, 
statistical sample complexity \citep{wei2021statistically, edelman2022inductive},
optimization process 
\citep{lu2021on,jelassi2022vision,li2023Transformers},
and interpretability
\citep{wen2023uninterpretability},
and references cited therein.

\paragraph{Mathematical setup.}
In the following we adapt and use the mathematical notations for the Transformer network architecture in \citet{yun2020are} and \citet{li2023Transformers}.

For each position of an input sequence ($N$ tokens)
$\tau_1 \cdots \tau_N$,
use a $d$-dimensional \emph{positional embedding} to represent that \emph{position},
and use a $d$-dimensional \emph{token embedding} for the \emph{content} at that position.
Hence, for the input sequence, both the token embeddings $\mE$ and the positional embeddings $\mP$ are matrices in $\R^{d \times N}$.
Following empirical convention, 
the encoder function $g_{\text{enc}}: \Omega^N \mapsto \R^{d \times N}$ is
\begin{equation}
\label{eq:encoder}
    \mX \coloneqq g_{\text{enc}}(\tau_1 \cdots \tau_N) = \mE + \mP
\end{equation}
which form the input to the Transformer blocks:

A \emph{Transformer block} $t^{h,m,r}$ 
(with $h$ heads, head size $m$, and feed-forward hidden layer size $r$)
is defined as
\begin{equation} \label{eq:transformer_block}
    t^{h,m,r}(\mX) \coloneqq \text{Attn}(\mX) + \mW_2 \cdot \text{ReLU} (\mW_1 \cdot \text{Attn}(\mX) + \vb_1 \vone_n^T) + \vb_2 \vone_n^T
\end{equation}
where 
\begin{equation} \label{eq:attn}
    \text{Attn}(\mX) \coloneqq \mX + \sum\nolimits_{i=1}^h \mW_O^i \mW_V^i \mX \cdot \sigma [(\mW_K^i \mX)^T \mW_Q^i \mX]
\end{equation}
where the weight parameters $\mW_O^i \in \R^{d \times m}$, $\mW_V^i, \mW_K^i, \mW_Q^i \in \R^{m \times d}$, $\mW_2 \in \R^{d \times r}, \mW_1 \in \R^{r \times d}, \vb_2 \in \R^{d}, \vb_1 \in \R^{r}$,
and
\[
\sigma: \R^{N_1 \times N_2} \mapsto (0, 1)^{N_1 \times N_2}
\]
is the column-wise softmax operation, such that 
\begin{equation} \label{eq:softmax}
    \sigma(A)_{ij} = \frac{\exp{(A_{ij})}}{\sum_{l=1}^N \exp{(A_{lj})}}
\end{equation}

An $L$-layer Transformer is a composition of Transformer blocks:
\begin{equation} \label{eq:transformer}
    \mathcal{T}
    \coloneqq
    \{ g_{\text{transformer}}: \R^{d \times N} \to \R^{d \times N} \mid 
    g_{\text{transformer}} = t_1 \circ \cdots \circ t_L
    \text{where $t_i$ is a Transformer block} 
    \}
\end{equation}

In the class of parallel decoding Transformers (denoted as $\mathcal{T}_{\text{PD}}$), 
for any Transformer $g_{\text{transformer}} \in \mathcal{T}$,
its output $g_{\text{transformer}}(\mX) \in \R^{d \times N}$ goes through a final affine transform and softmax (\Cref{eq:softmax}) to predict a distribution over tokens, for all positions
\begin{equation} \label{eq:transformer_pred}
    g_{\text{pred}}(\mX)
    \coloneqq \sigma \left( \mW^{\text{pred}} g_{\text{transformer}}(\mX) + \vb^{\text{pred}} \right)
    \in (0, 1)^{\abs{\Omega} \times N}
\end{equation}
where $\mW^{\text{pred}} \in \R^{\abs{\Omega} \times d}$ and $\vb^{\text{pred}} \in \R^{\abs{\Omega}}$ are the prediction head weights and biases. 
$\Omega$ is the vocabulary of tokens. 

For each position $j$, the predicted token $\tau_j$ is sampled from the predicted distribution $g_{\text{pred}}(\mX)_{:, j}$
\emph{independently} with other positions
\begin{equation} \label{eq:transformer_sample}
    \tau_j \sim \texttt{sample}(g_{\text{pred}}(\mX)_{:, j})
    \quad j \in [N]
\end{equation}
where $\texttt{sample}$ can be the standard sampling algorithm for multinomial distributions, 
or truncating the low-probability tail \citep{holtzman2020the},
or more conservatively, argmax sampling.

In the following definition, the sampling step can be denoted as a postprocessing function $s: (0, 1)^{\abs{\Omega}} \mapsto [0, 1]^{\abs{\Omega}}$ applied to each column of $g_{\text{pred}}(\mX)$.
For example, argmax sampling $s_{\text{argmax}}$ can be written as~\footnote{
In \Cref{eq:argmax}, the ``min" stipulates that when multiple coordinates of $p$ are tied for being argmax, 
i.e. $\abs{\argmax_{l} p_l} > 1$,
then $s_{\text{argmax}}$ would choose the smallest index in $s_{\text{argmax}}$.
Other reasonable tie-breaking behaviors also work.
}:
\begin{equation}
    \label{eq:argmax}
    \forall p \in (0, 1)^{\abs{\Omega}}, \quad
    s_{\text{argmax}}(p)_i = \begin{cases}
        1, \quad &\text{if } i = \min \argmax_{l} p_l \\ 
        0, \quad &\text{otherwise}
        \end{cases}
\end{equation}

Thus, 
combining \Cref{eq:encoder} and \Cref{eq:transformer_pred},
a parallel decoding Transformer (denoted as $\mathcal{T}_{\text{PD}}$) 
and a sampling function $s$
together define a Markov chain over sequences in $\Omega^N$:
\begin{equation} \label{eq:parallel_decoding_transformer}
    \mathcal{T}_{\text{PD}}
    \coloneqq
    \{ g \coloneqq s \circ_c g_{\text{pred}} \circ g_{\text{enc}} : \Omega^N \to [0, 1]^{\abs{\Omega} \times N} \}.
\end{equation}
where, again, $s$ is understood as applying to each column of its input matrix separately.
Specifically, in this Markov chain,
$\forall g \in \mathcal{T}_{\text{PD}}$ the transition probabilities are
\begin{equation}
\label{eq:transformer_mc}
\probp{\tau_1 \cdots \tau_N, \tau_1' \cdots \tau_N'} 
= \prod_{j \in [N]} g(\tau_1 \cdots \tau_N)_{\tau_j', j}
\end{equation}
where the last $g(\tau_1 \cdots \tau_N)_{\tau_j', j}$ denotes the $\tau_j'$-th row, $j$-th column of the matrix $g(\tau_1 \cdots \tau_N)$.

\citet{yun2020are} proved the following result on the expressivity of the Transformer network architecture:

\begin{lemma}[Universal approximation by Transformers, informal \citep{yun2020are}] \label{lem:universal_approximation}
Let $1 \leq p < \infty$ and $\epsilon > 0$, then 
for any compact set $\mathcal{D} \subset \R^{d \times n}$,
for any given function $f: \mathcal{D} \mapsto \R^{d \times n}$, there exists a Transformer network $g \in \mathcal{T}^{2,1,4}$ of $O(N \left( \frac{1}{\delta} \right)^{d N})$ layers such that
\[
\Big (\int \norm{f(\mX) - g(\mX)}_p^p d\mX \Big )^{1/p} \leq \epsilon
\]
in which $\delta$ is the smallest real number such that
$\forall \mX, \mY \in \R^{d \times n}$,
if $\| \mX - \mY \|_\infty < \delta$,
then $\norm{f(\mX) - f(\mY)}_p < \epsilon$.
Moreover, 
the bound on the size of the constructed Transformer is asymptotically tight.
\end{lemma}

\begin{lemma}[Transformers can simulate parallel solution to automata, Theorem 1 in \citep{liu2023Transformers}] \label{lem:shortcut_automata}
Transformers can simulate the length-$T$ output of all semiautomata with states $Q$, input alphabet $\Sigma$, and transition function $\delta: Q \times \Sigma \mapsto Q$.
Moreover, the size of the simulating Transformer has depth $O(\log T)$, embedding dimension $O(\abs{Q})$, attention width $O(\abs{Q})$, and MLP width $O(\abs{Q}^2)$.
\end{lemma}

\begin{remark}
\Cref{lem:shortcut_automata} gives a more compact construction than a direct implication of more general universal approximation results \Cref{lem:universal_approximation} for Transformers.
\end{remark}

A direct corollary is \Cref{prop:transformer_deterministic_mc}:


\begin{proposition}[\Cref{prop:transformer_deterministic_mc} formalized]
\label{prop:transformer_deterministic_mc_formalized}
For any function $f: \Omega^N \mapsto \Omega^N$,~\footnote{
This is equivalent to Markov chains over sequences in $\Omega^N$
whose transition probabilities are delta distributions
$\probp{\tau_1 \cdots \tau_N, \tau_1' \cdots \tau_N'} \in \{0, 1\}$.
The correspondence is: for any Markov chain state $\tau_1 \cdots \tau_N$, 
$f(\tau_1 \cdots \tau_N)$ specifies the unique deterministic next state.
}
for any $T \in \N_+$,
there exists a parallel decoding Transformer $g \in \mathcal{T}_{\text{PD}}$ (\Cref{eq:parallel_decoding_transformer})
whose sampling function is $s = s_{\text{argmax}}$ (\Cref{eq:argmax}),
such that 
\[
\forall \tau_1 \cdots \tau_N \in \Omega^N,
\forall t \in [T],
g^{(t)}(\tau_1 \cdots \tau_N) = f^{(t)}(\tau_1 \cdots \tau_N)
\]
where the $f^{(t)}$ denotes composing the function $f$ for $t$ times.
\end{proposition}

\begin{proof}
When each transition of a Markov chain is deterministic, 
i.e. if the next state distribution from any state is always a delta function, 
then the Markov chain reduces to a deterministic finite state automata, with states $\Omega^N$, length $N$.

Applying \Cref{lem:shortcut_automata},
we get Transformers can simulate length-$T$ output of this automata 
with depth $O(\log T)$, embedding dimension $O(\abs{\Omega}^N)$, attention width $O(\abs{\Omega}^N)$, and MLP width $O(\abs{\Omega}^{2N})$.
\end{proof}

\begin{proposition}[\Cref{prop:transformer_general_mc} formalized]
\label{prop:transformer_general_mc_formalized}
Let $Prod$ denote the set of Markov chains over sequences in $\Omega^N$
whose transition probabilities are product distributions over the positions, conditioned on the current state, i.e.
$\probp{\tau_1 \cdots \tau_N, \tau_1' \cdots \tau_N'} = \prod_{i=1}^N \probp{\tau_i' \mid \tau_1 \cdots \tau_N}$.
Then,
$\mathcal{T}_{\text{PD}} = Prod$.
\end{proposition}

\begin{proof} 
The statement involves both a positive result and a negative result on the expressivity of parallel decoding Transformers.

\textbf{Positive:} 
if the transition probability distribution is a product distribution conditioned on the current state,
then the task of representing a Markov chain can be reduced to universally approximating a continuous function which maps all sequences to the correct logits $\mW^{\text{pred}} \mathcal{T}(\mX) + \vb^{\text{pred}}$ in \Cref{eq:transformer_pred},
such that after softmax (\Cref{eq:softmax}) these logits produce the correct marginal distribution at each position.
This is achievable by the construction in \Cref{lem:universal_approximation}.

\textbf{Negative:} 
if the transition probability distribution is \emph{not} a product distribution conditioned on the current state,
then note that the sampling operations (\Cref{eq:transformer_sample}) at positions $j_1$ and $j_2$ are \emph{independent},
so Transformers cannot implement such Markov chains.
\end{proof}

\begin{remark}
\Cref{prop:transformer_general_mc} implies that parallel decoding Transformers cannot \emph{exactly represent} non-product distributions.
However, it does not prevent parallel decoding Transformers from \emph{approximating} some of non-product distributions.
This is because certain non-product distributions can be approximated by product distributions.
\end{remark}

\subsection{Connection to prior works in GMLM}
\label{sec:appendix:proof:transformer_ddlm_expressive_power:prior_works}

Among existing language generation approaches via iterative refinement, 
\citet{wang2019bert} uses $1$-Gibbs sampler.
The approaches in \citet{ghazvininejad2019mask, savinov2022stepunrolled} and our experiments do not closely fall into either of \textit{independent parallel} (\Cref{eq:gibbs:parallel}) or the $k$-Gibbs sampler (\Cref{eq:gibbs:k}) in \Cref{sec:theory:sampling}.
See \Cref{rem:mechanistic} for technical details.

Moreover, these approaches train models to learn the parameterized conditional distributions,
which empirically may not admit a consistent joint distribution \citep{young2022inconsistencies, hennigen2023deriving}.

To formally reason about the iterative refinement process in GMLMs, in \Cref{sec:theory:sampling} we relax some of these limitations to focus on several underlying theoretical obstacles that these methods face.

\begin{remark}[Technical details in theoretically formalizing GMLM architectures] \label{rem:mechanistic}
By \Cref{prop:transformer_general_mc},
the sampling process in \citet{ghazvininejad2019mask, savinov2022stepunrolled} and our experiments are different from $N$-Gibbs sampler.
Moreover, 
the sampling process is also different from  \textbf{independent parallel} (Gibbs sampler~\ref{item:gibbs:parallel}):
note that \textbf{independent parallel} strictly freezes all $X_{-\{i\}}^{(t)}$ when sampling
\[
X_i^{(t+1)} \sim p(\cdot \mid X_{-\{i\}}^{(t)})
\]
whereas in \citet{savinov2022stepunrolled} and our experiments, the model is trained to update all positions in parallel,
which implies a different groundtruth next-iteration token distribution compared with $p(\cdot \mid X_{-\{i\}}^{(t)})$.
In other words, although the updates are conditionally independent given the current state, the update probabilities are not trained to model $p(\cdot \mid X_{-\{i\}}^{(t)})$.

Mechanistically, \citet{savinov2022stepunrolled} and our models in principle can take certain inter-position dependency into consideration (which \textbf{independent parallel} cannot):
for example, in layer $L$, position $i$ can attend to~\footnote{
via Transformer attention \Cref{eq:attn}
} 
other positions e.g. $j$ in the layer-$(L-1)$ representations.
This enables the layer-$L$ computation at position $i$ to be conditioned upon the intermediate representations at position $j$, 
which are \emph{not independent} from the final prediction at position $j$.

\citet{ghazvininejad2019mask} can be understood as predicting the subset of masked indices $K$ in each update.
The extent to which each update incorporates dependency between masked positions
depends on implementation details:
for example, whether attention masks are added to prevent any masked position from receiving attention.

\end{remark}

\clearpage\section{Regularity conditions for asymptotic behavior of M-estimators}
\label{sec:appendix:proof:asymptotic}

In the limite of infinite samples, 
M-estimators (in particular, maximum likelihood and the estimators in Definitions \ref{d:genmple} and \ref{d:genmple:dependent}) converge in distribution to a normal distribution,
under mild regularity conditions:

\begin{lemma}[\cite{van2000asymptotic}, Theorem 5.23; statement adapted from \citet{qin2023fit}] \label{lem:asymptotics}
Consider a loss $L: \Theta \mapsto \mathbb{R}$, such that $L(\theta) = \E_p [\ell_{\theta}(x)]$ for $l_{\theta}:\mathcal{X} \mapsto \mathbb{R}$. Let $\Theta^*$ be the set of global minima of $L$, that is 
\[\Theta^* = \{\theta^*: L(\theta^*) = \min_{\theta \in \Theta} L(\theta)\}\]
Suppose the following conditions are met:
\begin{itemize}
\item (Gradient bounds on $l_{\theta}$) The map $\theta \mapsto l_{\theta}(x)$ is measurable and differentiable at every $\theta^* \in \Theta^*$ for $p$-almost every $x$. Furthermore, there exists a function $B(x)$, s.t. $\Ep{}{ B(x)^2 } < \infty$ and for every $\theta_1, \theta_2$ near $\theta^*$, we have: 
$$ |l_{\theta_1}(x) - l_{\theta_2}(x)| < B(x) \|\theta_1-\theta_2\|_2$$
\item (Twice-differentiability of $L$) $L(\theta)$ is twice-differentiable at every $\theta^* \in \Theta^*$ \\
with Hessian $\nabla_{\theta}^2 L(\theta^*)$, and furthermore $\nabla_{\theta}^2 L(\theta^*) \succ 0$. 
\item (Uniform law of large numbers) The loss $L$ satisfies a uniform law of large numbers, that is 
\[\sup_{\theta \in \Theta} \left|\hat{\E}[l_{\theta}(x)] - L(\theta) \right| \xrightarrow{p} 0 \] 
\item (Realizability)
The data distribution $p$ satisfies: $\exists \theta^* \in \Theta$
such that 
$p_{\theta^*} = p$. 
\end{itemize}
Then, for every $\theta^* \in \Theta^*$, and every sufficiently small neighborhood $S$ of $\theta^*$, there exists a sufficiently large $n$, such that there is a unique minimizer $\hat{\theta}_n$ of $\hat{\E}[l_{\theta}(x)] $ in $S$. Furthermore, $\hat{\theta}_n$ satisfies: 
\begin{align*}
    \sqrt{n} (\hat{\theta}_n - \theta^*) \xrightarrow{d} \mathcal{N} \left(0, (\nabla_\theta^2 L(\theta^*))^{-1} \mbox{Cov}(\nabla_\theta \ell(\theta^*; x)) (\nabla_\theta^2 L(\theta^*))^{-1} \right)
\end{align*}
\end{lemma}

\clearpage
\section{Convexity of pseudolikelihood for Ising models}
\label{sec:theory:optimization}

Here, we expand on a comment in \Cref{sec:theory}. We show that for a classic parameteric class of distributions (namely, Ising models --- which appear also in Section~\ref{sec:theory:sampling}) the k-MPLE loss is in fact \emph{convex}. 
This is a known fact which has been used to design (provably) efficient algorithms for learning bounded-degree Ising models \citep{ravikumar2010high, vuffray2016interaction}, and is just included for completeness. Recall the definition of Ising models
from \Cref{eqn:ising} in \Cref{def:ising}.
Let $Z$ be the partition function.

\begin{restatable}[Fitting an Ising model over the conditional distributions is convex]{proposition}{thmIsingConvexConditional}
\label{thm:ising_convex_conditional}
When $p_\theta$ is an Ising model (\Cref{eqn:ising}),
i.e. $\theta = \left( \mJ, \vh \right)$, the weighted pseudolikelihood objective (Definition~\ref{d:genmple}) is convex.
\end{restatable}
\begin{proof}
When $p_\theta$ is an Ising model (\Cref{eqn:ising}), we have:
\begin{align*}
    - \ln {p_\theta(\mathbf{x}_K | \mathbf{x}_{-K})} &= - \ln \frac{\exp{(\sum_{i \in [N]} \vh_i \mathbf{x}_i + \sum_{i \ne j \in [N]} \mJ_{ij} \mathbf{x}_i \mathbf{x}_j)}}{Z(\mathbf{x}_{-K})} \\
    &= - {\left(\sum_{i \in [N]} \vh_i \mathbf{x}_i + \sum_{i \ne j \in [N]} \mJ_{ij} \mathbf{x}_i \mathbf{x}_j\right)} + \ln {Z(\mathbf{x}_{-K})} 
\end{align*}
in which the denominator
\begin{align*}
Z(\mathbf{x}_{-K}) 
&= \sum_{X_K \in \{-1,1\}^{\abs{K}}}  \exp \bigg( \sum_{i \in K} \vh_i X_K 
+  \sum_{i \in [N] \backslash K} \vh_i \mathbf{x}_i \\
&+ \sum_{i \ne j \in [K]} \mJ_{ij} X_i X_j + \sum_{i \in K, j \in [N] \backslash K} \mJ_{ij} X_i \mathbf{x}_j  
+ \sum_{i \ne j \in [N] \backslash K} \mJ_{ij} \mathbf{x}_i \mathbf{x}_j \bigg) 
\end{align*}

Note that $- {\left(\sum_{i \in [N]} \vh_i \mathbf{x}_i + \sum_{i \ne j \in [N]} \mJ_{ij} \mathbf{x}_i \mathbf{x}_j\right)}$ is linear in $(\mathbf{h}, \mathbf{J})$
and 
$\ln {Z(\mathbf{x}_{-K})}$ is convex in $(h, J)$,
so $- \ln {p_\theta(X_K = \mathbf{x}_K | X_{-K} = \mathbf{x}_{-K})}$ is convex in $(\mathbf{h}, \mathbf{J})$,
which completes the last piece of the proof.

\end{proof}

\newpage
\clearpage
\section{Additional experimental details}
\label{sec:appendix:experiments}

\subsection{Training and inference approach for PaDIR}
\label{sec:appendix:experiments:setup}

We provide a formal description of the training and inference strategy outlined in \Cref{sec:method}.

\subsubsection{Inference}
\label{sec:method:inference}

An input sequence $X^{\text{source}}$ first goes through the encoder $f_{\theta_e}^{\text{enc}}$ (parameterized by $\theta_e$) to produce the hidden representation $h$:
\[ h = f_{\theta_e}^{\text{enc}}(X^{\text{source}}) \]

A length predictor $f_{\theta_l}^{\text{len}}$ (parameterized by $\theta_l$) takes $h$ and predicts $B_l$ most likely target lengths,
where $B_l \in \mathbb{N}_+$ (beam size for length prediction) is an inference-time hyperparameter.

For each predicted length $N$, 
an initial hypothesis target sequence $X^{(0)} = X_1^{(0)} \cdots X_N^{(0)}$ in which each $X_i^{(0)}$
can be a \texttt{[MASK]} token, or
chosen uniformly randomly from the vocabulary of tokens.

For each decoder step $t \in 1 \cdots T$,
the decoder $f_{\theta_d}^{\text{dec}}$ (parameterized by $\theta_d$) takes two inputs: $h$ and $X_{1 \cdots N}^{(t)}$,
and refines the hypothesis target sequence to $X_{1 \cdots N}^{(t+1)}$, using one forward pass:
\begin{equation} \label{eq:decoder_forward_pass}
    X_{1 \cdots N}^{(t+1)} = f_{\theta_d}^{\text{dec}}(X_{1 \cdots N}^{(t)}, h)
\end{equation}
where $T \in \mathbb{N}_+$ (number of refinement steps) is an inference-time hyperparameter,
and we can stop early if $X^{(t+1)} = X^{(t)}$.

\subsubsection{Training}
\label{sec:method:training}

\paragraph{One-stage training}
Given source sequence $X^{\text{source}}$ and target sequence $X^{\text{target}}$ in the supervised training data $\mathcal{D}_{\text{train}}$,
we use a preprocessing rule to create the initial hypothesis target sequence $X^{(0)}$.~\footnote{
Each position in $X^{(0)}$ may contain a \texttt{[MASK]} token, a random token, or the correct token in $X^{\text{source}}$, dependning on the preprocessing rule.
}
The training objective is
\begin{equation} \label{eq:objective_one_stage}
    L^{(1)} = \sum_{X^{\text{source}}, X^{\text{target}} \in \mathcal{D}_{\text{train}} } l(f_{\theta_d}^{\text{dec}}(X^{(0)}, f_{\theta_e}^{\text{enc}}(X^{\text{source}}))
\end{equation}
where $l$ is the cross-entropy loss applied to each position.

\paragraph{Multi-stage training}
One limitation of the one-stage training is that the inference situation is \emph{out-of-distribution}:
when decoder step $t > 1$, the model needs to refine its own predictions in step $t-1$, which is not reflected in the training objective.
Therefore, we use the multi-stage training objective \citep{ghazvininejad2020semiautoregressive, savinov2022stepunrolled}:
$
L^{(S)} = \frac{1}{S} \sum_{s \in [S]} L^{(s)}
$
where $S$ is the number of training stages,
and 
$
L^{(s)} = \sum_{X^{\text{source}}, X^{\text{target}} \in \mathcal{D}_{\text{train}} } l(f_{\theta_d}^{\text{dec}}(X^{(s-1)}, f_{\theta_e}^{\text{enc}}(X^{\text{source}}))
$

\subsection{Details of training recipe}
\label{sec:appendix:experiments:training}

We provide more details to \Cref{sec:experiments:eval}.

\paragraph{Model training} 
We use Transformer encoder-decoder with size similar to Transformer-Base \citep{vaswani2017attention} and T5-Small-1.0 \citep{raffel2020exploring}: 6 encoder and decoder layers, 8 attention heads, 512 embedding dimensions and 2048 FFN hidden dim. 
We add a positional attention mechanism \citep{gu2018nonautoregressive, kreutzer2020inference} in each Transformer layer and use learnt positional embeddings.
The total number of parameters is 67M. We initialize model parameters randomly and train using a batch size of 2048 for 500k iterations, with a $10\%$ dropout rate, $15\%$ unmasking rate~\footnote{
This means, in \Cref{eq:objective_one_stage}, $15\%$ of the tokens in $X^{(0)}$ are the correct tokens in $X^{\text{target}}$,
and the remaining $85\%$ are random tokens in the vocabulary.
} 
and 2 training stages. The optimizer is AdaFactor \citep{shazeer2018adafactor}, with default T5X hyperparameters \citep{roberts2022t5x}. The learning rate peaks at $0.003$ with a linear rampup for 10k steps followed by cosine decay, from and to a minimum value of $1e-5$. Unlike most prior work, we do not use a remasking schedule;~\footnote{
We experimented with various remasking schedules but the results were not visibly affected.
} 
we simply remask token-level stutter (i.e., consecutive repeated tokens) across iterations and drop repeated tokens after the final iteration.
As commonly done, we distill our models by training on the output of an autoregressive model. For simplicity, we use 
the Google Cloud Translation API
to generate this distillation data.

\paragraph{Datasets} We evaluate our models on machine translation benchmarks commonly used in the non-autoregressive modeling literature. We conduct experiments on both directions of three WMT datasets: WMT14 DE$\leftrightarrow$EN ($4.5M$ examples) \citep{wmt14}, WMT16 RO$\leftrightarrow$EN ($610k$ examples) \citep{wmt16} and WMT17 ZH$\leftrightarrow$EN ($20M$ examples) \citep{wmt17}. 
We load the data from the \texttt{tensorflow\_datasets} library and do not apply any preprocessing other than sentence piece tokenization (\cite{kudo2018sp}). Bilingual vocabularies of 32k tokens are created using the training sets of each language pair.

\subsection{Discussion on modeling and metrics}
\label{sec:appendix:experiments:metrics}

This section provides additional information about \Cref{sec:experiments:eval}.

\begin{remark}
In principle, following a similar paradigm, a non-autoregressive \emph{decoder-only} architecture is also possible. 
In this work we use encoder-decoder for two reasons:
(1) Efficiency: in the iterative refinement process of the hypothesis target sequence, each forward pass only involves the decoder, but not the encoder.
(2) Benchmarking: the encoder-decoder design is closer to a series of prior works,
allowing for more informative comparison on benchmarks.
\end{remark}

We measure BLEU \citep{papineni2002bleu} using the SacreBLEU
implementation \citep{post2018call} with language appropriate tokenizers~\footnote{
For public reproducibility: SacreBLEU signatures: BLEU+c.mixed+\#.1+s.exp+tok.zh+v.1.3.0
for Chinese and BLEU+c.mixed+\#.1+s.exp+tok.13a+v.1.3.0 for other languages.
}. 
For the same model, SacreBLEU on average reports a lower score than BLEU (e.g. see \cite{savinov2022stepunrolled}).
Unfortunately, this does not allow a direct comparison with most of the existing literature. This is a deliberate choice since it has been shown that subtle differences in preprocessing can significantly impact metrics \citep{schmidt2022clarity}, making comparisons error prone,
and SacreBLEU is the recommended metric in \citet{post2018call}.
Furthermore, common preprocessing steps (lowercasing, separating punctuation, stripping diacritics, etc.) may artificially inflate scores while not being fully reversible, as such preventing real-world uses for such models.

For our experiments in \Cref{sec:experiments:attention},
there are other error modes connected to the challenge of modeling target-side dependency,
but they are more ambiguous for measuring and exactly locating.
We do not aim to develop decoding algorithms tailored to just reducing stuttering rate.
(After all, stuttering can be easily removed by rule-based postprocessing.)
Instead, the above are general-purpose hypotheses which are potentially also predictive of other (more complex) failure modes related to target-side dependency.

\subsection{Quantitative experimental results on machine translation task}
\label{sec:appendix:experiments:results}

This section provides quantitative evaluation results for \Cref{sec:experiments:eval}.

\begin{table*}[h]
\caption{Test SacreBLEU scores on three WMT datasets. We report scores without any preprocessing. Our AR baselines are trained on the distilled dataset for a fair comparison. The `Steps' column indicates the number of decoding iterations. The `\# Hyp.' column denotes the number of hypotheses decoded in parallel (beam size for AR models and top$\_$k predicted lengths for NAR models).}
\label{table:experiments}
\vskip 0.15in
\begin{center}
\begin{small}
\begin{tabular}{ lcccccccc }
\toprule
& & & \multicolumn{2}{c}{WMT14} & \multicolumn{2}{c}{WMT16} & \multicolumn{2}{c}{WMT17} \\ 
Model & \# Hyp. & Steps & DE$\rightarrow$EN & EN$\rightarrow$DE & RO$\rightarrow$EN & EN$\rightarrow$RO & ZH$\rightarrow$EN & EN$\rightarrow$ZH \\
\hline
AR Baselines & 5 & N & 33.50 & 29.54 & 34.89 & 29.75 & 27.59 & 33.94 \\
\hline
\multirow{2}{10em}{PaDIR} & 5 & 4 & 33.49 & 28.61 & 33.98 & 28.98 & 26.47 & 32.59 \\
& 5 & 10 & 33.63 & 28.58 & 33.99 & 28.97 & 26.54 & 32.68 \\
\bottomrule
\end{tabular}
\end{small}
\end{center}
\end{table*}

\begin{table*}[h]
\caption{Test BLEU scores on three WMT datasets for baselines. 
Note that they use different BLEU implementations and sometimes additional preprocessing than the results reported for our approach.
We include results for our PaDIR under the T5X default BLEU score (SacreBLEU tok\_intl).
As we remarked in \Cref{sec:appendix:experiments:metrics}, these different BLEU implementations may not be directly comparable.
}
\label{table:baseline_bleu}
\vskip 0.15in
\begin{center}
\begin{small}
\begin{tabular}{ lcccccccc }
\toprule
& & & \multicolumn{2}{c}{WMT14} & \multicolumn{2}{c}{WMT16} & \multicolumn{2}{c}{WMT17} \\ 
Model & \# Hyp. & Steps & DE$\rightarrow$EN & EN$\rightarrow$DE & RO$\rightarrow$EN & EN$\rightarrow$RO & ZH$\rightarrow$EN & EN$\rightarrow$ZH \\
\hline
DisCo AR Baselines & 5 & N & 31.71 & 28.60 & 34.46 & 34.16 & 24.65 & 35.01 \\
\hline
\multirow{2}{10em}{CMLM} & 5 & 4 & 30.75 & 26.73 & 33.02 & 33.67 & 22.57 & 33.58 \\
& 5 & 10 & 31.24 & 27.39 & 33.67 & 33.33 & 23.76 & 34.24 \\
DisCo Easy-First & 5 & 3-6 & 31.31 & 27.34 & 33.25 & 33.22 & 23.83 & 34.63 \\
\multirow{2}{10em}{SUNDAE Stochastic} & 16 & 4 & 32.10 & 27.94 & - & - & - & - \\
& 16 & 10 & 32.29 & 28.33 & - & - & - & - \\
\hline
\multirow{2}{10em}{ PaDIR} &  5 &  4 & 34.17 & 29.49 & 34.55 & 29.57 & 27.18 & 32.59 \\
& 5 & 10 & 34.33 & 29.48 & 34.57 & 29.56 & 27.25 & 32.60 \\
\bottomrule
\end{tabular}
\end{small}
\end{center}
\end{table*}

\begin{table*}[h]
\caption{Test BLEURT scores on three WMT datasets for our models.}
\label{table:experiments_bleurt}
\vskip 0.15in
\begin{center}
\begin{small}
\begin{tabular}{ lcccccccc }
\toprule
& & & \multicolumn{2}{c}{WMT14} & \multicolumn{2}{c}{WMT16} & \multicolumn{2}{c}{WMT17} \\ 
Model & \# Hyp. & Steps & DE$\rightarrow$EN & EN$\rightarrow$DE & RO$\rightarrow$EN & EN$\rightarrow$RO & ZH$\rightarrow$EN & EN$\rightarrow$ZH \\
\hline
AR Baselines & 5 & N & 73.55 & 74.97 & 67.23 & 71.76 & 68.14 & 65.71 \\
\hline
\multirow{2}{10em}{PaDIR} & 5 & 4 & 71.26 & 72.08 & 65.90 & 70.23 & 65.16 & 63.95 \\
& 5 & 10 & 71.82 & 73.28 & 66.09 & 70.49 & 66.19 & 64.30 \\
\bottomrule
\end{tabular}
\end{small}
\end{center}
\end{table*}

\clearpage
\subsection{Quantifying dependency via attention scores}
\label{sec:appendix:experiments:attention}

We report quantitative results of the investigations introduced in \Cref{sec:experiments:attention}.

\begin{table*}[h]
\caption{Stuttering positions have comparable average last-layer self-attentions compared with non-stuttering adjacent positions. 
For each pair of adjacent positions in the generated sequence:
(1) the `self-attention scores' include both directions ;
(2) The column `min' denotes only including the minimum among such score over all attention heads,
and likewise for `avg' and `max';
(3) the entries are mean $\pm$ standard deviation;
(4) $\probp{\text{top-$k$ overlap}}$ denotes the chances that the self-attention distribution at one position includes the other position among its top-$k$ ``most attended to" positions.
}
\label{table:self_attn_main}
\vskip 0.15in
\begin{center}
\begin{small}
\begin{tabular}{ lcccccc }
\toprule
& \multicolumn{3}{c}{self-attention scores} &  \multicolumn{2}{c}{$\probp{\text{top-$k$ overlap}}$}  \\ 
stutter & min & avg & max & $k=1$ & $k=2$ \\
\hline
yes & 0.0004 $\pm$ 0.0007 & 0.032 $\pm$ 0.023  & 0.16 $\pm$ 0.11 & 0.20 & 0.39 \\
\hline
no & 0.0005 $\pm$ 0.0007 & 0.033 $\pm$ 0.025  & 0.17 $\pm$ 0.12 & 0.17 & 0.37 \\
\bottomrule
\end{tabular}
\end{small}
\end{center}
\end{table*}

\begin{table*}[h]
\caption{Stuttering positions on average have more similar last-layer cross-attentions than non-stuttering adjacent positions. 
For each pair of adjacent positions in the generated sequence:
(1) the `total variation distance' and `cosine distance' (both have range $[0, 1]$) are taken for the two corresponding cross-attention distributions;
(2) The column `min' denotes only including the minimum among such distance over all attention heads,
and likewise for `avg' and `max';
(3) the entries are mean $\pm$ standard deviation;
(4) $\probp{\text{top-$k$ overlap}}$ denotes the chances that the two cross-attention distributions overlap in terms of their top-$k$ ``most attended to" source positions.
}
\label{table:cross_attn_main}
\vskip 0.15in
\begin{center}
\begin{small}
\begin{tabular}{ lccccccccc }
\toprule
& \multicolumn{3}{c}{total variation distance} & \multicolumn{3}{c}{cosine distance} & \multicolumn{2}{c}{$\probp{\text{top-$k$ overlap}}$}  \\ 
stutter & min & avg & max & min & avg & max & $k=1$ & $k=2$ \\
\hline
yes & 0.06 $\pm$ 0.05 & 0.13 $\pm$ 0.09 & 0.23 $\pm$ 0.15 & 0.01 $\pm$ 0.01 & 0.10 $\pm$ 0.06 & 0.25 $\pm$ 0.11 & 0.57 & 0.89 \\
\hline
no  & 0.11 $\pm$ 0.10 & 0.23 $\pm$ 0.14 & 0.35 $\pm$ 0.18 & 0.04 $\pm$ 0.08 & 0.20 $\pm$ 0.11 & 0.38 $\pm$ 0.12 & 0.40 & 0.81 \\
\bottomrule
\end{tabular}
\end{small}
\end{center}
\end{table*}

\clearpage
\subsection{Masking more is statistically better for learning synthetic Ising models}
\label{sec:appendix:experiments:synthetic}

\begin{figure}[!h]
  \centering
  \begin{minipage}[b]{0.5\textwidth}  
    \centering
    \includegraphics[width=1.0\textwidth]{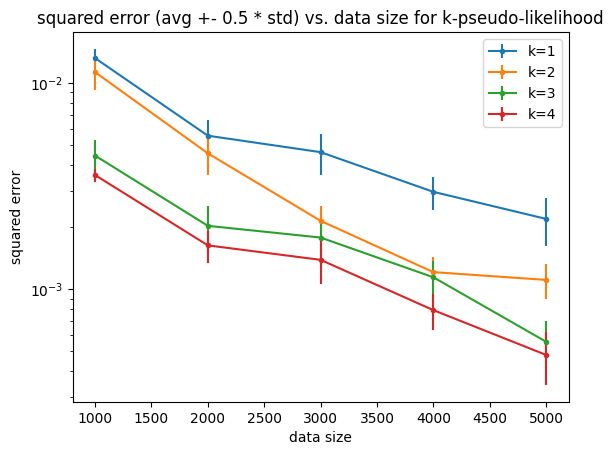}  
  \end{minipage}
  \caption{Average squared error in parameter estimation 
  for fitting an Ising model on data generated by a groundtruth Ising model
  ($N = \abs{C_G} = 4, J = 0.05, h_i = 0$ in \Cref{eqn:ising:construction})
  using the $k$-pseudolikelihood objective optimized by gradient descent.
  Error bars denote $\pm$ 0.5 * stdev for 10 repetitions of the experiment.
  }  
  \label{fig:ising_k_monotone_flat}
\end{figure}

\begin{figure}[!h]
  \centering
  \begin{minipage}[b]{0.5\textwidth}  
    \centering
    \includegraphics[width=1.0\textwidth]{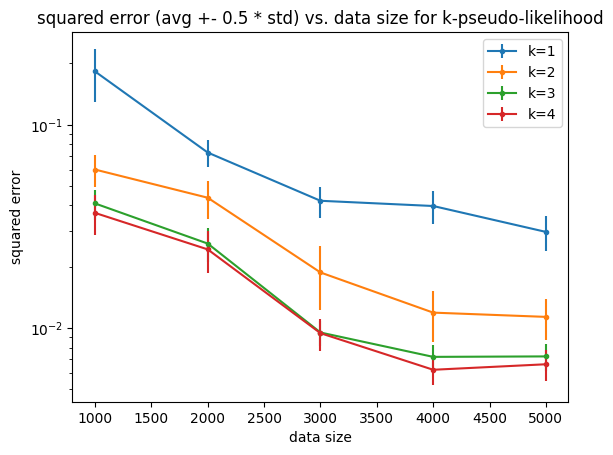}  
  \end{minipage}
  \caption{Average squared error in parameter estimation 
  for fitting an Ising model on data generated by a groundtruth Ising model
  ($N = \abs{C_G} = 4, J = 0.3, h_i = 0$ in \Cref{eqn:ising:construction})
  using the $k$-pseudolikelihood objective optimized by gradient descent.
  Error bars denote $\pm$ 0.5 * stdev for 10 repetitions of the experiment.
  }  
  \label{fig:ising_k_monotone_peaky}
\end{figure}

We show our observations in \Cref{sec:experiments:ising} and \Cref{fig:ising_k_monotone_flat} are robust to the shape of the groundtruth Ising model distribution:
under a much more peaky groundtruth distribution (with 2 modes),
it still holds that with the same training data size, larger $k$ leads to lower error. 
We plot the results in \Cref{fig:ising_k_monotone_peaky}.

\clearpage
\subsection{Markov Chains with dependent transitions can be (much) faster in sampling Ising models}
\label{sec:appendix:experiments:synthetic_sampling}

To verify our theory in \Cref{sec:theory:sampling:gibbs}, 
we run controlled experiments benchmarking various sampling algorithms for Ising models:
$k$-Gibbs sampler (\Cref{d:kgibbs}), and
the independent parallel sampler (\Cref{item:gibbs:parallel}).

The Ising model distribution that we sample from contains two modes, one larger and the other smaller, 
corresponding to $\mathcal{R}_1$ and $\mathcal{R}_{-1}$ defined in \Cref{eq:larger_mode} and \Cref{eq:smaller_mode}, respectively.

\begin{figure}[!h]
  \centering
  \begin{minipage}[b]{0.5\textwidth}  
    \centering
    \includegraphics[width=1.0\textwidth]{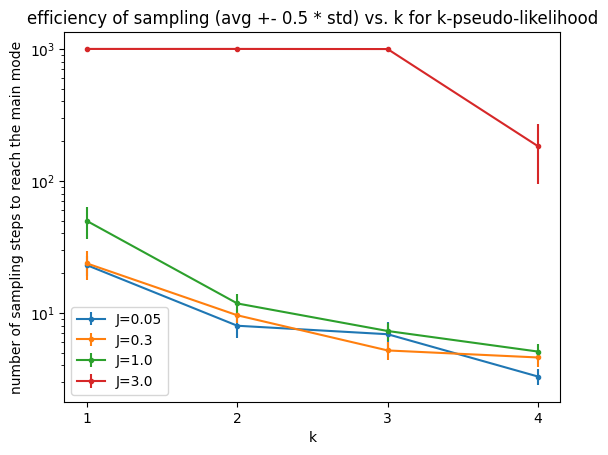}  
  \end{minipage}
  \caption{Number of steps for the $k$-Gibbs sampler (\Cref{d:kgibbs}) to reach the larger mode $\mathcal{R}_1$ (\Cref{eq:larger_mode}) of Ising models, starting from the smaller mode $\mathcal{R}_{-1}$ (\Cref{eq:smaller_mode}).
  The parameters of our Ising models are:
  $N = 10, \abs{C_G} = 4, h_i = 5.0$ in \Cref{eqn:ising:construction}.
  We vary the parameter $J$ (a larger $J$ corresponds to a more peaky distribution).
  Error bars denote $\pm$ 0.5 * stdev for 10 repetitions of the experiment.
  The compute budget is 1000 steps. Thus, a point with vertical coordinate $10^3$ means that the sampler did not reach $\mathcal{R}_1$ within compute budget.
  The $k$-Gibbs sampler can often reach the $\mathcal{R}_1$ (larger $k$ is faster).
  For context, the independent parallel sampler (not on the plot) can never reach $\mathcal{R}_1$ within the compute budget for any of the $J$'s we tried.
  }  
  \label{fig:sampling}
\end{figure}

We show in \Cref{fig:sampling} that if we initialize the sample in the smaller mode $\mathcal{R}_{-1}$, 
running the $k$-Gibbs sampler (\Cref{d:kgibbs}) can often reach the larger mode $\mathcal{R}_1$ within a relatively small number of steps
(though more peaky distributions i.e. those with larger $J$, are slower to sample).
Moreover, larger $k$ is faster than smaller $k$.
By contrast, running the independent parallel sampler (\Cref{item:gibbs:parallel}) cannot reach $\mathcal{R}_1$ within the compute budget we set.
The results verify our theory in \Cref{sec:theory:sampling:gibbs}
that Markov Chains with dependent transitions can be (much) faster in sampling Ising models (compared with the independent parallel sampler).

\newpage
\clearpage
\section{Additional related works}
\label{sec:appendix:related_works}

We expand on the discussion in \Cref{sec:related_works}.

\paragraph{Non-autoregressive text generation}
Previous works applied various generative models to text, such as
VAEs \citep{bowman2016generating, bosc2020sequence},
GANs \citep{che2017maximumlikelihood, yu2017seqgan, lin2017adversarial, guo2018long},
and normalizing flows \citep{ziegler2019latent, ma2019flowseq, hoogeboom2021argmax},
but without a strong autoregressive component, the quality of generated text is often suboptimal.
Later works achieve high-quality text generation through
diffusion models \citep{hoogeboom2021argmax, austin2021structured, li2022diffusionlm, gong2023diffuseq, zheng2023reparameterized} 
and energy-based models \citep{deng2020residual, goyal2022exposing, qin2022cold},
but their generation speeds tend to be much slower than autoregressive language models.
Inference latency can be mitigated by approaches like \citet{lee2020iterative}.
Unlike the above paradigms that adapt continuous-domain generative models to text,
our approach is closer to the following line of works that iteratively refine the generation process through parallel updates in the space of discrete token sequences, which tend to be at least twice faster than autoregressive approaches with a small drop in quality \citep{lee2018deterministic, ghazvininejad2019mask, stern2019insertion, guo2020jointly, ghazvininejad2020semiautoregressive, kasai2020nonautoregressive, savinov2022stepunrolled}
(though autoregressive models also have the potential for speedup by using a shallower decoder for certain tasks  \citep{kasai2021deep}).
The generation quality of non-autoregressive models can be further improved by incorporating some autoregressive components \citep{kong2020incorporating, reid2022diffuser}
or input-output alignment \citep{chan2020imputer, saharia2020non},
or adaptive training curriculum \citep{qian2021glancing}.
Insights such as the multimodality problem and components such as sequence-level knowledge distillation and input token fertility prediction were also proposed in \citep{gu2018nonautoregressive}.
The benefit of distillation was verified in \citet{kim2016sequence, gu2018nonautoregressive, zhou2020understanding, gu2021fully}.
Positional attention was tested in \citet{gu2018nonautoregressive, kreutzer2020inference}.
Relevant to our experiments in \Cref{sec:experiments:attention},
\citet{ren2020study} measure the target-side dependency as the proportion of attention paid to target tokens as opposed to the source tokens, in some modified attention architecture. 
Related to generation from MLMs, \citet{wang2019bert} use the learned conditionals inside a Gibbs sampler,
but when the conditionals are not \emph{consistent}, i.e. there is not a joint distribution that satisfies these conditionals, Gibbs sampler may amplify errors. 
In general, mathematical understanding about sampling from masked language models is still lagging substantially behind.
Additionally, related to MLMs,
\citet{meng2023representation} analyzes some representational limitations,
and \citet{liu2022masked} analyzes subtleties from a parameter identifiability view.
Related to parallel decoding,
recent work \citep{cai2024medusa} parallelizes the inference with multiple heads by finetuning autoregressive LLM backbones.

\paragraph{Theory about parallel sampling}
\citet{koehler2023statistical} proved a generalization bound for pseudolikelihood estimator via the classic ($k = 1$) approximate tensorization of entropy,
in the ``proper learning" setting.
Our generalization bound (\Cref{thm:generalization_bound_joint}) uses the generalized notion of the approximate tensorization of entropy (\Cref{d:block_catconst}),
also apply to ``improper learning" settings,
and the proof involves quite different techniques.
The classic approximate tensorization of entropy 
are discussed in 
\citet{marton2013inequality, marton2015logarithmic, caputo2015approximate},
which was more recently generalized to the ``$\alpha$-weighted block" version (\Cref{d:block_catconst}) in \citet{caputo2021block}.
\citet{lee2023parallelising} proves that $k$-Gibbs sampler mixes at least $k$ times faster than $1$-Gibbs sampler.
For future works, 
recent algorithmic advances in parallel sampling could potentially be incorporated into our framework to achieve finer-grained theoretical analysis or better empirical quality-efficiency trade-off \citep{anari2023parallel}.

\end{document}